\documentclass[runningheads]{llncs}
\usepackage{graphicx}
\usepackage{amsmath,amssymb} % define this before the line numbering.
\usepackage{color}
\usepackage[width=122mm,left=12mm,paperwidth=146mm,height=193mm,top=12mm,paperheight=217mm]{geometry}

\usepackage{eqnarray}
\usepackage{units}
\usepackage{caption}
\usepackage{subcaption}
\usepackage{url}
\usepackage{algorithm}
\usepackage{algpseudocode}
\usepackage{mathtools}
\usepackage{bbm}
\usepackage{colortbl}
\usepackage{multirow}

\usepackage{microtype}

\def\etal{\emph{et al}.}

\DeclarePairedDelimiter{\norm}{\lVert}{\rVert}

\definecolor{Yellow}{rgb}{1,1, 0.6}
\definecolor{Orange}{rgb}{1,0.8, 0.6}
\definecolor{Red}{rgb}{1, 0.6, 0.6}

\definecolor{Yellow}{rgb}{1,1, 0.6}
\definecolor{Orange}{rgb}{1,0.8, 0.6}
\definecolor{Red}{rgb}{1, 0.6, 0.6}

\definecolor{Teal}{rgb}{0.85, 0.85, 0.85}
\definecolor{Blue}{rgb}{0.85, 0.85, 0.85}
\definecolor{Pink}{rgb}{0.85, 0.85, 0.85}
\definecolor{Green}{rgb}{0.85, 0.85, 0.85}
\definecolor{WindowsColor}{rgb}{0.85, 0.85, 0.85}

\begin{document}
\pagestyle{headings}
\mainmatter

\title{The Fast Bilateral Solver}

\titlerunning{The Fast Bilateral Solver}

\authorrunning{Jonathan T. Barron \& Ben Poole}

\author{
Jonathan T. Barron \hspace{0.65in} Ben Poole \hspace{0.15in} \\
\hspace{0.45in} \email{barron@google.com}  \hspace{0.25in} \email{poole@cs.stanford.edu}
}

\institute{}

\maketitle

\begin{abstract}
We present the bilateral solver, a novel algorithm for edge-aware smoothing that combines the flexibility and speed of simple filtering approaches with the accuracy of domain-specific optimization algorithms.
Our technique is capable of matching or improving upon state-of-the-art results on several different computer vision tasks (stereo, depth superresolution, colorization, and semantic segmentation) while being $10$-$1000\times$ faster than baseline techniques with comparable accuracy, and producing lower-error output than techniques with comparable runtimes.
The bilateral solver is fast, robust, straightforward to generalize to new domains, and simple to integrate into deep learning pipelines.
\end{abstract}

\section{Introduction}

\newcommand{\figonewidth}{2.42in}

\begin{figure}[b!]
        \begin{tabular}{l}
        \begin{subfigure}[!]{\figonewidth}
            \includegraphics[width=2.3in]{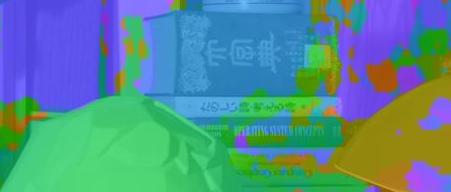}
            \caption{Input (MAE = $6.00$, RMSE = $38.8$) \label{fig:middleburyA}}
        \end{subfigure}
        \begin{subfigure}[!]{\figonewidth}
            \includegraphics[width=2.3in]{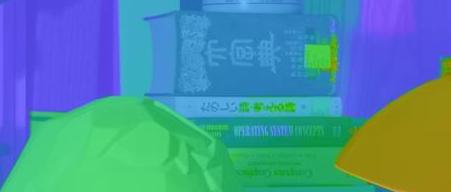}
            \caption{Output (MAE = $3.02$, RMSE = $17.9$) \label{fig:middleburyB}}
        \end{subfigure}
        \\ \\
        \begin{subfigure}[!]{\figonewidth}
        \includegraphics[width=2.3in]{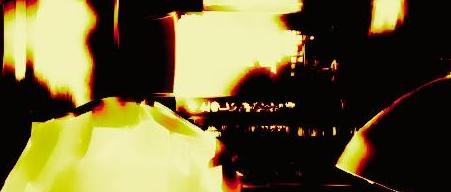}
        \caption{Input Confidence \label{fig:middlebury_conf}}
        \end{subfigure}
        \begin{subfigure}[!]{\figonewidth}
                \includegraphics[width=2.3in]{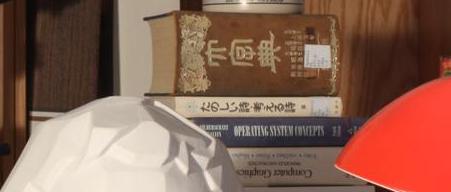}
                \caption{Input Reference \label{fig:middlebury_ref}}
        \end{subfigure}
        \end{tabular}
        \caption{
        The bilateral solver can be used to improve depth maps.
        A depth map (\subref{fig:middleburyA}) from a state-of-the-art stereo method \cite{Zbontar2015} is processed with our robust bilateral solver using a reference RGB image (\subref{fig:middlebury_ref}).
        Our output (\subref{fig:middleburyB}) is smooth with respect to the reference image, resulting in a $50\%$ reduction in error.
        \label{fig:middlebury}
        }
\end{figure}

Images of the natural world exhibit a useful prior -- many scene properties (depth, color, object category, etc.) are correlated within smooth regions of an image, while differing across discontinuities in the image.
Edge-aware smoothing techniques exploit this relationship to propagate signals of interest within, but not across edges present in an image.
Traditional approaches to edge-aware smoothing apply an image-dependent filter to a signal of interest.
Examples of this include joint bilateral filtering \cite{Smith1997,Tomasi1998} and upsampling \cite{Kopf2007}, adaptive manifolds \cite{GastalOliveira2012AdaptiveManifolds}, the domain transform \cite{GastalOliveira2011DomainTransform}, the guided filter \cite{He2010,He2015}, MST-based filtering \cite{Yang2015}, and weighted median filtering \cite{Ma2013,Zhang2014}.
These techniques are flexible and computationally efficient, but often insufficient for solving more challenging computer vision tasks.
Difficult tasks often necessitate complex iterative inference or optimization procedures that encourage smoothness while maintaining fidelity with respect to some observation.
Optimization algorithms of this nature have been used in global stereo \cite{Scharstein2002}, depth superresolution \cite{ferstl2013b,Kiechle2013,Kwon2015,Li2013,Lu2015,Park2011}, colorization \cite{Levin2004}, and semantic segmentation \cite{deeplab,KrahenbuhlK11,fcn,crfasrnn}.
These approaches are tailored to their specific task, and are generally computationally expensive.
In this work we present an optimization algorithm that is $10$-$1000\times$ faster than existing domain-specific approaches with comparable accuracy, and produces higher-quality output than lightweight filtering techniques with comparable runtimes.

Our algorithm is based on the work of Barron \etal \cite{Barron2015A}, who presented the idea of using fast bilateral filtering techniques to solve optimization problems in ``bilateral-space''.
This allows for some optimization problems with bilateral affinity terms to be solved quickly, and also guarantees that the solutions to those problems are ``bilateral-smooth'' --- smooth within objects, but not smooth across edges.
In this paper we present a new form of bilateral-space optimization which we call \emph{the bilateral solver}, which efficiently solves a regularized least-squares optimization problem to produce an output that is bilateral-smooth and close to the input.
This approach has a number of benefits:

{\bf General} The bilateral solver is a single intuitive abstraction that can be applied to many different problems, while matching or beating the specialized state-of-the-art algorithms for each of these problems.
It can be generalized to a variety of loss functions using standard techniques from M-estimation \cite{Hampel86a}.

{\bf Differentiable} Unlike other approaches for edge-aware smoothness which require a complicated and expensive ``unrolling'' to perform backpropagation \cite{crfasrnn}, the backward pass through our solver is as simple and fast as the forward pass, allowing it to be easily incorporated into deep learning architectures.

{\bf Fast} The bilateral solver is expressible as a linear least-squares optimization problem, unlike the non-linear optimization problem used in \cite{Barron2015A}.
This enables a number of optimization improvements including a hierarchical preconditioner and initialization technique that hasten convergence, as well as efficient methods for solving multiple problems at once.

\section{Problem Formulation}

We begin by presenting the objective and optimization techniques that make up our bilateral solver.
Let us assume that we have some per-pixel input quantities $\mathbf{t}$ (the ``target'' value, see Figure~\ref{fig:middleburyA}) and some per-pixel confidence of those quantities $\mathbf{c}$ (Figure \ref{fig:middlebury_conf}), both represented as vectorized images.
Let us also assume that we have some ``reference'' image (Figure~\ref{fig:middlebury_ref}), which is a normal RGB image.
Our goal is to recover an ``output'' vector $\mathbf{x}$ (Figure~\ref{fig:middleburyB}), which will resemble the input target where the confidence is large while being smooth and tightly aligned to edges in the reference image.
We will accomplish this by constructing an optimization problem consisting of an image-dependent smoothness term that encourages $\mathbf{x}$ to be bilateral-smooth, and a data-fidelity term that minimizes the squared residual between $\mathbf{x}$ and the target $\mathbf{t}$ weighted by our confidence $\mathbf{c}$:
\begin{align}
  \underset{\mathbf{x}}{\mathrm{minimize}} &\,\, \frac{\lambda}{2} \sum_{i, j} \hat W_{i,j} \left(x_i - x_j \right)^2  + \sum_i c_i (x_i - t_i)^2
\label{eq:pixel_loss}
\end{align}
The smoothness term in this optimization problem is built around an affinity matrix $\hat W$, which is a bistochastized version of a bilateral affinity matrix $W$.
Each element of the bilateral affinity matrix $W_{i, j}$ reflects the affinity between pixels $i$ and $j$ in the reference image in the YUV colorspace:
\begin{equation}
\resizebox{4.4in}{!}{
$
W_{i,j} = \exp\left( - { \norm{ [[p^x_i, p^y_i] - [p^x_j, p^y_j] }^2 \over 2 \sigma_{xy}^2 } -  { ( p^l_i - p^l_j )^2 \over 2 \sigma_{l}^2 } -  { \norm{ [p^u_i, p^v_i] - [p^u_j, p^v_j] }^2 \over 2 \sigma_{uv}^2 } \right)
$
}
\label{eq:bilateral_affinity}
\end{equation}
Where $p_i$ is a pixel in our reference image with a spatial position $(p^x_i, p^y_i)$ and color $(p^l_i, p^u_i, p^v_i)$\footnote{To reduce confusion between the Y's in ``YUV'' and ``XY'' we refer to luma as ``l''}.
The $\sigma_{xy}$, $\sigma_{l}$, and $\sigma_{uv}$ parameters control the extent of the spatial, luma, and chroma support of the filter, respectively.

This $W$ matrix is commonly used in the bilateral filter \cite{Tomasi1998}, an edge-preserving filter that blurs within regions but not across edges by locally adapting the filter to the image content.
There are techniques for speeding up bilateral filtering \cite{Adams2010,Chen2007} which treat the filter as a ``splat/blur/slice'' procedure:
pixel values are ``splatted'' onto a small set of vertices in a grid \cite{Barron2015A,Chen2007} or lattice \cite{Adams2010} (a soft histogramming operation), then those vertex values are blurred, and then the filtered pixel values are produced via a ``slice'' (an interpolation) of the blurred vertex values.
These splat/blur/slice filtering approaches all correspond to a compact and efficient factorization of $W$:
\begin{equation}
W = S^\mathrm{T} {\bar B} S \label{eq:factorization}
\end{equation}
Barron \etal \cite{Barron2015A} built on this idea to allow for optimization problems to be ``splatted'' and solved in bilateral-space.
They use a ``simplified'' bilateral grid and a technique for producing bistochastization matrices $D_\mathbf{n}$, $D_\mathbf{m}$ that together give the the following equivalences:
\begin{equation}
\hat W = S^\mathrm{T} D_\mathbf{m}^{-1} D_\mathbf{n} {\bar B} D_\mathbf{n} D_\mathbf{m}^{-1} S \quad\quad SS^\mathrm{T} = D_\mathbf{m} \label{eq:equiv}
\end{equation}
They also perform a variable substitution, which reformulates a high-dimensional pixel-space optimization problem in terms of the lower-dimensional bilateral-space vertices:
\begin{equation}
\mathbf{x} = S^\mathrm{T} \mathbf{y} \label{eq:substitution}
\end{equation}
Where $\mathbf{y}$ is a small vector of values for each bilateral-space vertex, while $\mathbf{x}$ is a large vector of values for each pixel.
With these tools we can not only reformulate our pixel-space loss function in Eq~\ref{eq:pixel_loss} in bilateral-space, but we can rewrite that bilateral-space loss function in a quadratic form:
\begin{align}
\underset{\mathbf{y}}{\mathrm{minimize}} \quad & {1 \over 2} \mathbf{y}^\mathrm{T} A \mathbf{y} - \mathbf{b}^\mathrm{T} \mathbf{y} + c \label{eq:quad_min} \\
A = \lambda (D_\mathbf{m} - D_\mathbf{n} \bar B D_\mathbf{n}) + \mathrm{diag}(S \mathbf{c}) & \quad\quad \mathbf{b} = S ( \mathbf{c} \circ \mathbf{t} ) \quad\quad c = {1 \over 2} (\mathbf{c} \circ \mathbf{t})^\mathrm{T} \mathbf{t} \nonumber
\end{align}
where $\circ$ is the Hadamard product.
A derivation of this reformulation can be found in the supplement.
While the optimization problem in Equation~\ref{eq:pixel_loss} is intractably expensive to solve naively, in this bilateral-space formulation optimization can be performed quickly.
Minimizing that quadratic form is equivalent to solving a sparse linear system:
\begin{equation}
A\mathbf{y} = \mathbf{b}
\label{eq:linear_system}
\end{equation}
We can produce a pixel-space solution $\mathbf{\hat x}$ by simply slicing the solution to that linear system:
\begin{equation}
\mathbf{\hat x} = S^\mathrm{T}(A^{-1} \mathbf{b})
\label{eq:solution}
\end{equation}

With this we can describe our algorithm, which we will refer to as the ``bilateral solver.''
The input to the solver is a reference RGB image, a target image that contains noisy observed quantities which we wish to improve, and a confidence image.
We construct a simplified bilateral grid from the reference image, which is bistochastized as in \cite{Barron2015A} (see the supplement for details), and with that we construct the $A$ matrix and $\mathbf{b}$ vector described in Equation~\ref{eq:quad_min} which are used to solve the linear system in Equation~\ref{eq:solution} to produce an output image.
If we have multiple target images (with the same reference and confidence images) then we can construct a larger linear system in which $\mathbf{b}$ has many columns, and solve for each channel simultaneously using the same $A$ matrix. In this many-target case, if $\mathbf{b}$ is low rank then that property can be exploited to accelerate optimization, as we show in the supplement.

Our pixel-space loss (Eq~\ref{eq:pixel_loss}) resembles that of weighted least squares filtering \cite{Elad2002,FFLS2008,Min2014}, with one critical difference being our use of bilateral-space optimization which allows for efficient optimization even when using a large spatial support in the bilateral affinity, thereby improving the quality of our output and the speed of our algorithm.
Our algorithm is similar to the optimization problem that underlies the stereo technique of \cite{Barron2015A}, but with several advantages:
Our approach reduces to a simple least-squares problem, which allows us to optimize using standard techniques (we use the preconditioned conjugate gradient algorithm of \cite{Shewchuk1994}, see the supplement for details).
This simple least-squares formulation also allows us to efficiently backpropagate through the solver (Section~\ref{sec:backprop}), allowing it to be integrated into deep learning pipelines.
This formulation also improves the rate of convergence during optimization, provides guarantees on correctness, allows us to use advanced techniques for preconditioning and initialization (Section~\ref{sec:precond}), and enables robust and multivariate generalizations of our solver (see the supplement).

\section{Backpropagation}
\label{sec:backprop}

Integrating any operation into a deep learning framework requires that it is possible to backpropagate through that operation.
Backpropagating through global operators such as our bilateral solver is generally understood to be difficult, and is an active research area \cite{Ionescu_2015_ICCV}.
Unlike most global smoothing operators, our model is easy to backpropagate through by construction.
Note that we do not mean backpropagating \emph{through} a multiplication of a matrix inverse $A^{-1}$, which would simply be another multiplication by $A^{-1}$.
Instead, we will backpropagate \emph{onto} the $A$ matrix used in the least-squares solve that underpins the bilateral solver, thereby allowing us to backpropagate through the bilateral solver itself.

Consider the general problem of solving a linear system:
\begin{eqnarray}
A \mathbf{y} = \mathbf{b}
\end{eqnarray}
Where $A$ is an invertible square matrix, and $\mathbf{y}$ and $\mathbf{b}$ are vectors. We can solve for $\mathbf{\hat y}$ as a simple least squares problem:
\begin{eqnarray}
\mathbf{\hat y} = A^{-1} \mathbf{b}
\end{eqnarray}
Let us assume that $A$ is symmetric in addition to being positive definite, which is true in our case.
Now let us compute some loss with respect to our estimated vector $g(\mathbf{\hat y})$, whose gradient will be $\partial g / \partial \mathbf{\hat y}$.
We would like to backpropagate that quantity onto $A$ and $\mathbf{b}$:
\begin{equation}
{\partial_g \over \partial_\mathbf{b}} =  A^{-1} {\partial_g \over \partial_\mathbf{\hat y}}  \quad\quad\quad
{\partial_g \over \partial_A} = \left( - A^{-1}  {\partial_g \over \partial_\mathbf{\hat y}} \right) \mathbf{\hat y}^\mathrm{T} = -{\partial_g \over \partial_\mathbf{b}} \mathbf{\hat y}^\mathrm{T} \label{eq:linear_backprop}
\end{equation}
This can be derived using the implicit function theorem. We see that backpropagating a gradient through a linear system only requires a single least-squares solve. The gradient of the loss with respect to the diagonal of $A$ can be computed more efficiently:
\begin{align}
{\partial_g \over \partial_{\mathrm{diag}(A)} } &= -{\partial_g \over \partial_\mathbf{b}} \circ \mathbf{\hat y}
\end{align}

We will use these observations to backpropagate through the bilateral solver.
The bilateral solver takes some input target $\mathbf{t}$ and some input confidence $\mathbf{c}$, and then constructs a linear system that gives us a bilateral-space solution $\mathbf{\hat y}$, from which we can ``slice'' out a pixel-space solution $\mathbf{\hat x}$.
\begin{equation}
\mathbf{\hat y} = A^{-1} \mathbf{b} \quad\quad\quad \mathbf{\hat x} = S^\mathrm{T} \mathbf{\hat y}
\label{eq:forward}
\end{equation}
Note that $A$ and $\mathbf{b}$ are both functions of $\mathbf{t}$ and $\mathbf{c}$, though they are not written as such.
Let us assume that we have computed some loss $f(\hat x)$ and its gradient $\partial_f / \partial_\mathbf{\hat x}$.
Remember that the $A$ matrix and $\mathbf{b}$ vector in our linear system are functions of some input signal $\mathbf{t}$ and some input confidence $\mathbf{c}$.
Using (\ref{eq:linear_backprop}) we can compute the gradient of the loss with respect to the parameters of the linear system within the bilateral solver:
\begin{equation}
{\partial_f \over \partial_\mathbf{b}} = A^{-1} \left( S { \partial_f \over \partial_\mathbf{\hat x}}\right) \quad\quad\quad
{\partial_f \over \partial_{\mathrm{diag}(A)} } = -{\partial_f \over \partial_\mathbf{b}} \circ \mathbf{\hat y}
\end{equation}
We need only compute the gradient of the loss with respect to the diagonal of $A$ as opposed to the entirety of $A$, because the off-diagonal elements of $A$ do not depend on the input signal or confidence.
We can now backpropagate the gradient of the loss $f( \mathbf{\hat x} )$ onto the inputs of the bilateral solver:
\begin{equation}
{\partial_f \over \partial_{\mathbf{t}}} = \mathbf{c} \circ \left( S^\mathrm{T} {\partial_f \over \partial_\mathbf{b}} \right) \quad\quad\quad
{\partial_f \over \partial_{\mathbf{c}}} = \left(S^\mathrm{T} {\partial_f \over \partial_{\mathrm{diag}(A)} } \right) + \left(S^\mathrm{T} {\partial_f \over \partial_\mathbf{b}} \right) \circ \mathbf{t}
\end{equation}

To review, the bilateral solver can be viewed as a function which takes in a reference image, some input signal and a per-pixel confidence in that input signal, and produces some smoothed output:
\begin{equation}
\mathrm{output} \leftarrow \mathrm{solver}_\mathrm{reference}(\mathrm{target}, \mathrm{confidence} )
\end{equation}
And we have shown how to backpropagate through the solver:
\begin{equation}
(\nabla\mathrm{target}, \nabla\mathrm{confidence}) \leftarrow \mathrm{backprop}_\mathrm{reference}( \nabla\mathrm{output} )
\end{equation}
Because the computational cost of the backwards pass is dominated by the least squares solve necessary to compute ${\partial_f / \partial_\mathbf{b}}$, computing the backward pass through the solver is no more costly than computing the forward pass.
Contrast this with past approaches for using iterative optimization algorithms in deep learning architectures, which create a sequence of layers, one for each iteration in optimization \cite{crfasrnn}.
The backward pass in these networks is a fixed function of the forward pass and so cannot adapt like the bilateral solver to the structure of the error gradient at the output.
Furthermore, in these ``unrolled'' architectures, the output at each iteration (layer) must be stored during training, causing the memory requirement to grow linearly with the number of iterations.
In the bilateral solver, the memory requirements are small and independent of the number of iterations, as we only need to store the bilateral-space output of the solver $\mathbf{\hat y}$ during training.
These properties make the bilateral solver an attractive option for deep learning architectures where speed and memory usage are important.

\section{Preconditioning \& Initialization}
\label{sec:precond}

Optimization of the quadratic objective of the bilateral solver can be sped up with improved initialization and preconditioning.
In the previous work of \cite{Barron2015A}, the non-linear optimization used a hierarchical technique which lifted optimization into a pyramid space, using a bilateral variant of the image pyramid optimization approach of \cite{BarronTPAMI2015}.
This approach cannot be used by our solver, as most linear solvers require a preconditioner where the input is of the same dimensionality as the output.
Regardless, the approach of \cite{Barron2015A} is also suboptimal for our use case, as the simple linear structure of our system allows us to construct more accurate and effective preconditioning and initialization techniques.

\begin{figure}[!t]
\centering
  \includegraphics[width=4.2in]{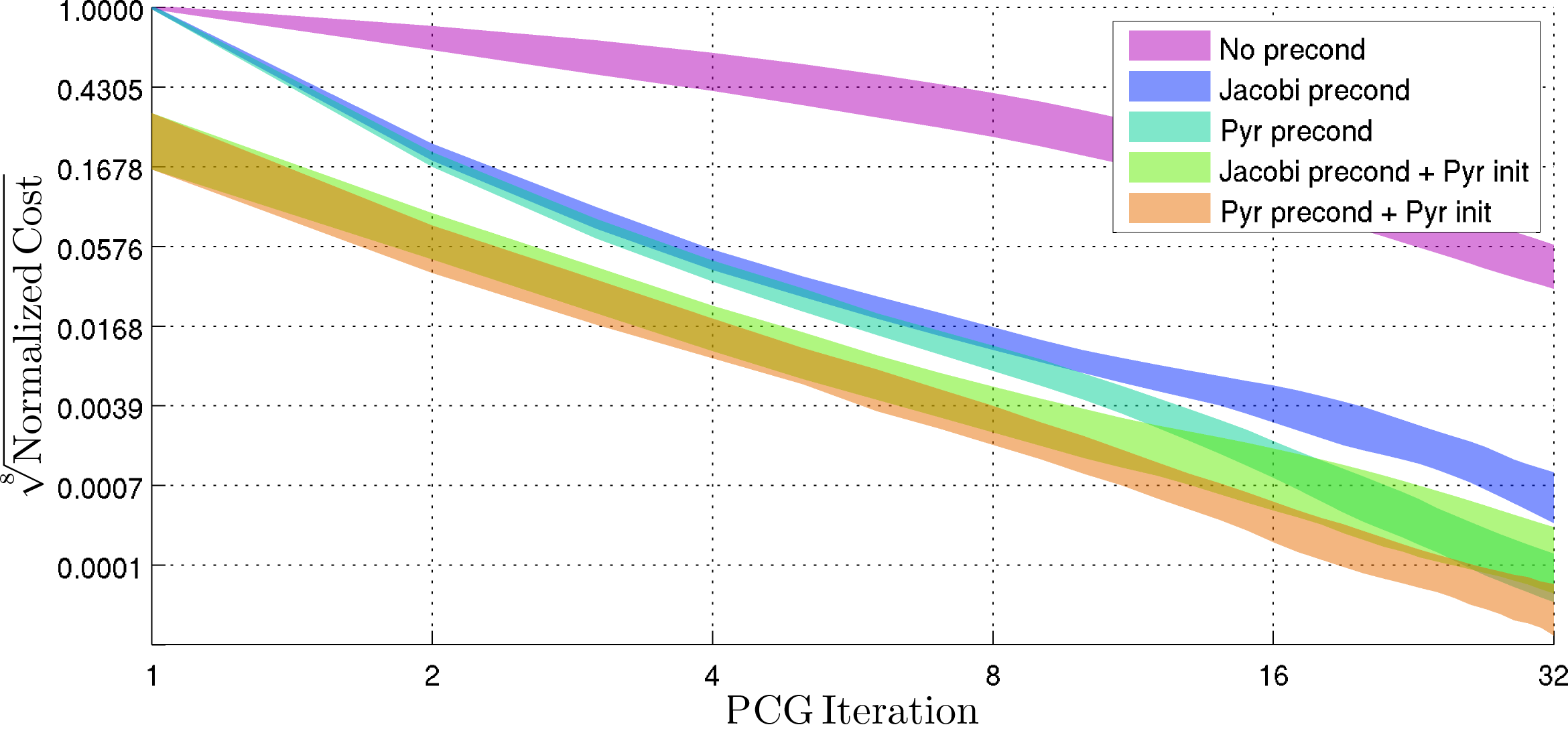}
  \caption{
    Our loss during PCG for $20$ $4$-megapixel images, with the loss for each image normalized to $[0, 1]$ and with the $25$th-$75$th percentiles plotted.
    We see that preconditioning is critical, and that our hierarchical (``Pyr'') preconditioning and initialization techniques significantly improve performance over the naive Jacobi preconditioner and ``flat'' initialization.
    Note the non-linear $y$-axis and logarithmic $x$-axis.
    \label{fig:precon_init}
  }
\end{figure}

To best explain our preconditioning and initialization techniques we must first present baselines techniques for both.
We can extract the diagonal of our $A$ matrix to construct a Jacobi preconditioner:
\begin{equation}
\resizebox{3.2in}{!}{$
\mathrm{diag}(A) = \lambda \left( \mathrm{diag}\left(D_\mathbf{m}) - \mathrm{diag}(D_\mathbf{n} \right) \bar B_{\mathrm{diag}} \mathrm{diag}(D_\mathbf{n}) \right) + S \mathbf{c} \nonumber
$}
\end{equation}
This is straightforward to compute, as $D_\mathbf{m}$ and $D_\mathbf{n}$ are diagonal matrices and $\bar B$ has a constant value along the diagonal denoted here as $\bar B_{\mathrm{diag}}$.
The Jacobi preconditioner is simply the inverse of the diagonal of $A$:
\begin{equation}
M_\mathit{jacobi}^{-1}(\mathbf{y}) = \mathrm{diag}(A)^{-1} \mathbf{y}
\end{equation}
We can also initialize the state vector $\mathbf{y}$ in our optimization to the value which minimizes the data term in our loss, which has a closed form:
\begin{equation}
\mathbf{y}_\mathit{flat} = S ( \mathbf{c} \circ \mathbf{t} ) / S ( \mathbf{c} ) \label{eq:flat_init}
\end{equation}
This preconditioner and initialization technique perform well, as can be seen in Figure~\ref{fig:precon_init}.
But we can improve upon these baseline techniques by constructing hierarchical generalizations of each.

Hierarchical preconditioners have been studied extensively for image interpolation and optimization tasks.
Unfortunately, techniques based on image pyramids \cite{Szeliski1990} are not applicable to our task as our optimization occurs in a sparse $5$-dimensional bilateral-space.
More sophisticated image-dependent or graph based techniques \cite{koutis2011combinatorial,Krishnan2013,Szeliski2006} are effective preconditioners, but in our experiments the cost of constructing the preconditioner greatly outweighs the savings provided by the improved conditioning.
We will present a novel preconditioner which is similar in spirit to hierarchical basis functions \cite{Szeliski1990} or push-pull interpolation \cite{Gortler1996}, but adapted to our task using the bilateral pyramid techniques presented in \cite{Barron2015A}.
Because of its bilateral nature, our preconditioner is inherently locally adapted and so resembles image-adapted preconditioners \cite{Krishnan2013,Szeliski2006}.

We will use the multiscale representation of bilateral-space presented in \cite{Barron2015A} to implement our hierarchical preconditioner.
This gives us
$P(\mathbf{y})$ and $P^\mathrm{T}(\mathbf{z})$,
which construct a pyramid-space vector $\mathbf{z}$ from a bilateral-space vector $\mathbf{y}$, and collapse $\mathbf{z}$ down to $\mathbf{y}$ respectively (see the supplement for details).
To evaluate our preconditioner, we lift our bilateral-space vector into pyramid-space, apply an element-wise scaling of each pyramid coefficient, and then project back onto bilateral-space:
\begin{equation}
M_\mathit{hier}^{-1}(\mathrm{y}) = P^\mathrm{T}\left( { \mathbf{z}_\mathit{weight} \circ P(\mathbf{1}) \circ P(\mathbf{y}) \over P(\mathrm{diag}(A)) } \right)
\label{eq:hier_precond}
\end{equation}
where the division is element-wise. $M_\mathit{hier}^{-1}(\cdot)$ includes an ad-hoc element-wise scaling:
\begin{align}
\mathbf{z}_\mathit{weight} &=
\begin{cases}
  1                     & \mbox{if } k = 0 \\
  \alpha^{-(\beta + k)} & \mbox{otherwise}
\end{cases}
\end{align}
The pyramid-space scaling we use in Equation~\ref{eq:hier_precond} is proportional to:
1) the number of bilateral-space vertices assigned to each pyramid-space coefficient (computed by lifting a vector of ones),
2) the inverse of the diagonal of the $A$ matrix, computed by lifting and inverting the diagonal of the $A$ matrix, and
3) an exponential weighting of each pyramid-space coefficient according to its level in the pyramid.
This per-level scaling $\mathbf{z}_\mathit{weight}$ is computed as a function of the level $k$ of each coefficient, which allows us to prescribe the influence that each scale of the pyramid should have in the preconditioner.
Note that as the coarser levels are weighed less (ie, as $\alpha$ or $\beta$ increases) our preconditioner degenerates naturally to the Jacobi preconditioner.
In all experiments we use ($\alpha = 2$, $\beta = 5$) for the preconditioner.

This same bilateral pyramid approach can be used to effectively initialize the state before optimization.
Rather than simply taking the input target and using it as our initial state as was done in Equation~\ref{eq:flat_init}, we perform a push-pull filter of that initial state with the pyramid according to the input confidence:
\begin{equation}
\mathbf{y}_\mathit{hier} = P^\mathrm{T}\left( { \mathbf{z}_\mathit{weight} \circ P(S ( \mathbf{c} \circ \mathbf{t} )) \over P(\mathbf{1}) } \right) / P^\mathrm{T}\left( { \mathbf{z}_\mathit{weight} \circ P(S ( \mathbf{c} )) \over P(\mathbf{1}) } \right)
\end{equation}
Like our hierarchical preconditioner, this initialization degrades naturally to our non-hierarchical initialization in Eq.~\ref{eq:flat_init} as $\alpha$ and $\beta$ increase.
In all experiments we use ($\alpha = 4$, $\beta = 0$) for initialization.

See Figure~\ref{fig:precon_init} for a visualization of how our hierarchical preconditioning and initialization improve convergence during optimization, compared to the ``flat'' baseline algorithms.
See Table~\ref{table:stereo_speed} for a comparison of our runtime compared to \cite{Barron2015A}, where we observe a substantial speedup with respect to the solver of \cite{Barron2015A}.
Though the techniques presented here for efficient optimization and initialization are framed in terms of the forward pass through the solver, they all apply directly to the backward pass through the solver described in Section~\ref{sec:backprop}, and produce equivalent improvements in speed.

\begin{table}[t!]
\begin{center}
\caption{
  Our approach's runtime has a lower mean and variance than that of \cite{Barron2015A}.
  Runtimes are from the same workstation, averaged over the $20$ $4$-megapixel images used in \cite{Barron2015A} for profiling.
  \label{table:stereo_speed}
}
\resizebox{3.4in}{!}{
\small
\begin{tabular}{l | c c }
& \multicolumn{2}{c}{Time (ms)} \\
$\mathrm{Algorithm}$ $\mathrm{Component}$  & $\mathrm{Barron}$ \etal \cite{Barron2015A} \hspace{0.05in} & \hspace{0.05in}  $\mathrm{This}$ $\mathrm{Work}$ \\
\hline
$\mathrm{Problem}$ $\mathrm{Construction}$ & $ 190 \pm 167 $ & $ 35 \pm 7 $  \\
$\mathrm{Optimization}$ & $ 460 \pm 207 $ & $ 152 \pm 36 $  \\
\hline
$\mathrm{Total}$ & $ 650 \pm 266 $ & $ 187 \pm 37 $
\end{tabular}
}
\end{center}
\end{table}

\section{Applications}

We evaluate our solver on a variety of applications: stereo, depth superresolution, image colorization, and semantic segmentation.
Each of these tasks has been the focus of significant research, with specialized techniques having been developed for each problem.
For some of these applications (semantic segmentation and stereo) our solver serves as a building block in a larger algorithm, while for others (colorization and depth superresolution) our solver is a complete algorithm.
We will demonstrate that our bilateral solver produces results that are comparable to or better than the state-of-the-art for each problem, while being either 1-3 orders of magnitude faster.
For those techniques with comparable runtimes, we will demonstrate that the bilateral solver produces higher quality output.
Unless otherwise noted, all runtimes were benchmarked on a 2012 HP Z420 workstation (Intel Xeon CPU E5-1650, 3.20GHz, 32 GB RAM), and our algorithm is implemented in standard, single-threaded C++.
As was done in \cite{Barron2015A}, the output of our bilateral solver is post-processed by the domain transform \cite{GastalOliveira2011DomainTransform} to smooth out the blocky artifacts introduced by the simplified bilateral grid, and the domain transform is included in all runtimes.
For all results of each application we use the same implementation of the same algorithm with different parameters, which are noted in each sub-section. Parameters are:
the spatial bandwidths of the bilateral grid ($\sigma_{\mathit{xy}}$, $\sigma_{\mathit{l}}$, $\sigma_{\mathit{uv}}$),
the smoothness multiplier ($\lambda$),
the spatial and range bandwidths of the domain transform ($\sigma'_{\mathit{xy}}$, $\sigma'_{\mathit{rgb}}$).
Unless otherwise stated, the bilateral solver is run for $25$ iterations of PCG.

\subsection{Stereo}

We first demonstrate the utility of the bilateral solver as a post-processing procedure for stereo algorithms.
Because depth maps produced by stereo algorithms tend to have heavy-tailed noise distributions, we use a variant of our technique called the robust bilateral solver (RBS) with the Geman-McClure loss (described in the supplement).
We applied the RBS to the output of the top-performing MC-CNN \cite{Zbontar2015} algorithm on the Middlebury Stereo Benchmark V3 \cite{Zbontar2015}.
For comparison, we also evaluated against four other techniques which can or have been used to post-process the output of stereo algorithms.
In Table~\ref{table:middlebury} we see that the RBS cuts test- and training-set absolute and RMS errors in half while having little negative effect on the ``bad $1$\%'' error metric (the percent of pixels which whose disparities are wrong by more than $1$).
This improvement is smaller when we only consider non-occluded (NoOcc) as most state-of-the-art stereo algorithms already perform well in the absence of occlusions.
The improvement provided by the RBS is more dramatic when the depth maps are visualized, as can be seen in Figure~\ref{fig:middlebury} and in the supplement.
At submission time our technique achieved a lower test-set MAE and RMSE on the Middlebury benchmark than any published technique\footnote{\url{http://vision.middlebury.edu/stereo/eval3/}}.

See the supplement for a discussion of how our baseline comparison results were produced, an evaluation of our RBS and our baseline techniques on three additional contemporary stereo algorithms, the parameters settings used in this experiment, how we compute the initial confidence $\mathbf{c}$ for the RBS, and many visualizations.

\begin{table}[b!]
\begin{center}
\caption{
Our robust bilateral solver significantly improves depth map quality the state-of-the-art MC-CNN\cite{Zbontar2015} stereo algorithm on the Middlebury dataset V3 \cite{Scharstein2014}.
\label{table:middlebury}
}
\resizebox{3.5in}{!}{
\begin{tabular}{ l | c c c | c c c }
$\mathrm{Method}$ & \multicolumn{3}{c|}{$\mathrm{All}$} & \multicolumn{3}{c}{$\mathrm{NoOcc}$}  \\
 & $\mathrm{bad}\,1\%$ & $\mathrm{MAE}$ & $\mathrm{RMSE}$ & $\mathrm{bad}\,1\%$ & $\mathrm{MAE}$ & $\mathrm{RMSE}$  \\
\hline
\multicolumn{7}{l}{\quad Test Set} \\
\hline
$\mathrm{                   MC-CNN}$\cite{Zbontar2015} & $ 28.1 $ \cellcolor{Yellow}  & $ 17.9 $                     & $ 55.0 $                     & $ 18.0 $ \cellcolor{Yellow} & $ 3.82 $                     & $ 21.3 $                    \\
$\mathrm{             MC-CNN}$\cite{Zbontar2015}$\mathrm{ + RBS}$ & $ 28.2 $                     & $ 8.19 $ \cellcolor{Yellow}  & $ 29.9 $ \cellcolor{Yellow}  & $ 18.9 $                    & $ 2.67 $ \cellcolor{Yellow}  & $ 15.0 $ \cellcolor{Yellow} \\
\multicolumn{7}{c}{} \\
\hline
\multicolumn{7}{l}{\quad Training Set} \\
\hline
    $\mathrm{MC}$-$\mathrm{CNN}$\cite{Zbontar2015}                        & $ 20.07 $  & $ 5.93 $  & $ 18.36 $  & $ 10.42 $ \cellcolor{Yellow}  & $ 1.94 $  & $ 9.07 $  \\
    $\mathrm{MC}$-$\mathrm{CNN}$\cite{Zbontar2015} + $\mathrm{TF}$\cite{Yang2015} & $ 29.15 $  & $ 5.67 $  & $ 16.18 $  & $ 20.15 $  & $ 2.17 $  & $ 7.71 $  \\
    $\mathrm{MC}$-$\mathrm{CNN}$\cite{Zbontar2015} + $\mathrm{FGF}$\cite{He2015} & $ 32.29 $  & $ 5.91 $  & $ 16.32 $  & $ 23.62 $  & $ 2.42 $  & $ 7.98 $  \\
    $\mathrm{MC}$-$\mathrm{CNN}$\cite{Zbontar2015} + $\mathrm{WMF}$\cite{Ma2013} & $ 33.37 $  & $ 5.30 $  & $ 15.62 $  & $ 26.29 $  & $ 2.32 $  & $ 8.22 $  \\
    $\mathrm{MC}$-$\mathrm{CNN}$\cite{Zbontar2015} + $\mathrm{DT}$\cite{GastalOliveira2011DomainTransform} & $ 25.17 $  & $ 5.69 $  & $ 16.53 $  & $ 15.53 $  & $ 2.01 $  & $ 7.72 $  \\
    $\mathrm{MC}$-$\mathrm{CNN}$\cite{Zbontar2015} + $\mathrm{RBS\,(Ours)}$ & $ 19.49 $ \cellcolor{Yellow}  & $ 2.81 $ \cellcolor{Yellow}  & $ 8.44 $ \cellcolor{Yellow}  & $ 11.33 $  & $ 1.40 $ \cellcolor{Yellow}  & $ 5.23 $ \cellcolor{Yellow}
\end{tabular}
}
\end{center}
\end{table}

\subsection{Depth Superresolution}

With the advent of consumer depth sensors, techniques have been proposed for upsampling noisy depth maps produced by these sensors using a high-resolution RGB reference image \cite{ferstl2013b,Kiechle2013,Kwon2015,Li2013,Lu2015,Park2011,chan2008,Liu2013,Min2014}.
Other techniques have been developed for post-processing depth maps in other contexts \cite{Yang2015,Ma2013,Shen2015}, and many general edge-aware upsampling or filtering techniques can be used for this task\cite{Kopf2007,GastalOliveira2011DomainTransform,He2010,He2015,Zhang2014}.
We present an extensive evaluation of the bilateral solver against these approaches for the depth superresolution task.
Given a noisy input depth map and an RGB reference image, we resize the depth map to be the size of the reference image with bicubic interpolation and then apply the bilateral solver or one of our baseline techniques.
\begin{table}[b!]
\begin{center}
\caption{
Performance on the depth superresolution task of \cite{ferstl2013b}.
Runtimes in gray were not computed on our reference hardware, and algorithms which use external training data are indicated with a dagger.
\label{table:super}
}
\resizebox{4.0in}{!}{
\begin{tabular}{c c c}
\begin{tabular}{@{}r@{}l | c | c }
 & $\mathrm{Method}$ & $\mathrm{Err}$ & $\mathrm{Time\,(sec)}$ \\
\hline
&                               $\mathrm{Nearest\,Neighbor}$   &  $ 7.26 $  &  $ 0.003 $  \\
&                                         $\mathrm{Bicubic}$   &  $ 5.91 $  &  $ 0.007 $  \\
$\dagger$ &               $\mathrm{Kiechle\,\etal }$\cite{Kiechle2013}   &  $ 5.86 $  &  \cellcolor{Blue} $ 450 $  \\
&                                        $\mathrm{Bilinear}$   &  $ 5.16 $  &  $ 0.004 $  \\
&                      $\mathrm{Liu\,\etal\,}$\cite{Liu2013}   &  $ 5.10 $  &  $ 16.60 $  \\
&                    $\mathrm{Shen\,\etal\,}$\cite{Shen2015}   &  $ 4.24 $  &   $ 31.48 $  \\
&             $\mathrm{Diebel\,\&\,Thrun\,}$\cite{Diebel05b}   &  $ 3.98 $  &  $ - $  \\
&                     $\mathrm{Chan\,\etal }$\cite{chan2008}   &  $ 3.83 $  &  \cellcolor{Green}$ 3.02 $   \\
&         $\mathrm{Guided Filter }$\cite{He2010,ferstl2013b}   &  $ 3.76 $  &  \cellcolor{Green} $ 23.89 $   \\
&                $\mathrm{Min\,\etal\,}$\cite{Min2014}   &  $ 3.74 $  &  $ 0.383 $  \\
$\dagger$ &                   $\mathrm{Lu\,\&\,Forsyth }$\cite{Lu2015}   &  $ 3.69 $  &  \cellcolor{Pink} $ 20 $  \\
&                     $\mathrm{Park\,\etal }$\cite{Park2011}   &  $ 3.61 $  &  \cellcolor{Green}$ 24.05 $   \\
\multicolumn{3}{c}{ \vdots }
\end{tabular}
& \hspace{0.15in} &
\begin{tabular}{@{}r@{}l | c | c }
\multicolumn{3}{c}{ \vdots } \\
&$\mathrm{Domain\,Transform\,}$\cite{GastalOliveira2011DomainTransform}   &  $ 3.56 $  &  $ 0.021 $  \\
&                        $\mathrm{Ma\,\etal\,}$\cite{Ma2013}   &  $ 3.49 $  &  \cellcolor{Teal} $ 18 $  \\
&            $\mathrm{Guided Filter (Matlab) }$\cite{He2010}   &  $ 3.47 $  &  $ 0.434 $  \\
&                  $\mathrm{Zhang\,\etal\,}$\cite{Zhang2014}   &  $ 3.45 $  &  \cellcolor{WindowsColor} $ 1.346 $  \\
&                $\mathrm{Fast Guided Filter }$\cite{He2015}   &  $ 3.41 $  &  $ 0.225 $  \\
&                     $\mathrm{Yang\,2015\,}$\cite{Yang2015}   &  $ 3.41 $  &  \cellcolor{WindowsColor} $ 0.304 $  \\
&              $\mathrm{Yang\,\etal\,2007\,}$\cite{Yang2007}   &  $ 3.25 $  &  $ - $  \\
&                 $\mathrm{Farbman\,\etal\,}$\cite{FFLS2008}   &  $ 3.19 $  &   $ 6.11 $  \\
&                  $\mathrm{JBU\,}$\cite{Adams2010,Kopf2007}   &  $ 3.14 $  &  $ 1.98 $  \\
&                $\mathrm{Ferstl\,\etal }$\cite{ferstl2013b}   &  $ 2.93 $  &  \cellcolor{Green}$ 140 $  \\
$\dagger$ &                         $\mathrm{Li\,\etal }$\cite{Li2013}   & \cellcolor{Orange}  $ 2.56 $  &  \cellcolor{Blue} $ 700 $  \\
$\dagger$ &                     $\mathrm{Kwon\,\etal }$\cite{Kwon2015}   & \cellcolor{Red}  $ 1.21 $  &  \cellcolor{Blue}$ 300 $  \\
\hline
&                                      $\mathrm{BS\,(Ours)}$   & \cellcolor{Yellow}  $ 2.70 $  &  $ 0.234 $
\end{tabular}
\end{tabular}
}
\end{center}
\end{table}
The hyperparameters used by the solver for all experiments are: $\sigma_{\mathit{xy}} = 8$, $\sigma_l = 4$, $\sigma_{\mathit{uv}} = 3$, $\sigma'_{\mathit{xy}} = \sigma'_{\mathit{rgb}} = 16$, $\lambda = 4^{f-1/2}$ (where $f$ is the upsampling factor) and $15$ iterations of PCG.
Our confidence $\mathbf{c}$ is a Gaussian bump ($\sigma=f/4$) modeling the support of each low-resolution pixel in the upsampled image.
To evaluate our model, we use the depth superresolution benchmark of \cite{ferstl2013b} which is based on the Middlebury stereo dataset \cite{Scharstein2002}.
Our performance can be see in Table~\ref{table:super} and Figure~\ref{fig:super}, with more detailed results in the supplement.
The bilateral solver produces the third-lowest error rate for this task, though the two better-performing tasks \cite{Kwon2015,Li2013} use large amounts of external training data and so have an advantage over our technique, which uses no learning for this experiment.
Our approach is $600\times$, $1200\times$, and $3000\times$ faster than the three most accurate techniques.
The techniques with speeds comparable to or better than the bilateral solver \cite{GastalOliveira2011DomainTransform,Yang2015,He2015,Min2014} produce error rates that are 25-40$\%$ greater than our approach.
The bilateral solver represents a effective combination of speed and accuracy, while requiring no training or learning.
See the supplement for a more detailed table, a discussion of baselines and runtimes, and many visualizations.

\newcommand{\superwidth}{1.15in}

\begin{figure}[!]
\centering
        \begin{subfigure}[!]{\superwidth}
                \includegraphics[width=\superwidth]{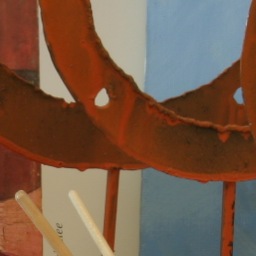}
                \caption{Input Image}
        \end{subfigure}
        \begin{subfigure}[!]{\superwidth}
                \includegraphics[width=\superwidth]{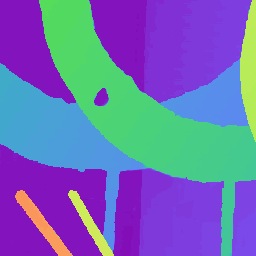}
                \caption{True Depth}
        \end{subfigure}
        \begin{subfigure}[!]{\superwidth}
                \includegraphics[width=\superwidth]{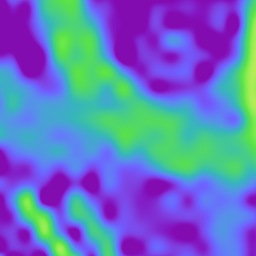}
                \caption{Input Depth}
        \end{subfigure}
        \begin{subfigure}[!]{\superwidth}
               \includegraphics[width=\superwidth]{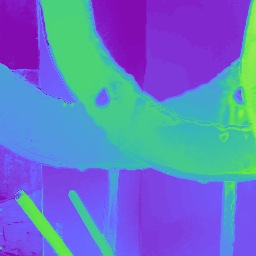}
               \caption{JBU \cite{Adams2010,Kopf2007}}
        \end{subfigure}
        \\
        \begin{subfigure}[!]{\superwidth}
               \includegraphics[width=\superwidth]{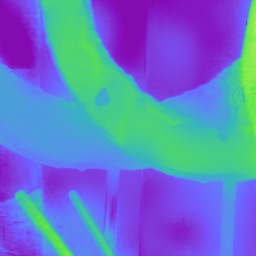}
               \caption{Guided Filter \cite{He2015}}
        \end{subfigure}
        \begin{subfigure}[!]{\superwidth}
                \includegraphics[width=\superwidth]{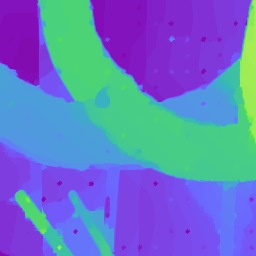}
               \caption{Ferstl \etal \cite{ferstl2013b}}
        \end{subfigure}
        \begin{subfigure}[!]{\superwidth}
                \includegraphics[width=\superwidth]{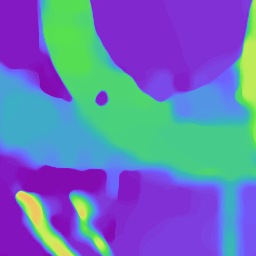}
               \caption{Li \etal \cite{Li2013}}
        \end{subfigure}
        \begin{subfigure}[!]{\superwidth}
                \includegraphics[width=\superwidth]{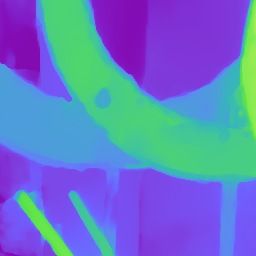}
                \caption{Our results}
        \end{subfigure}
        \caption{
        Partial results for the depth superresolution task of \cite{ferstl2013b}, see the supplement for exhaustive visualizations.
        \label{fig:super}
        }
\end{figure}

\subsection{Colorization}

\begin{figure}[b!]
\centering
    \begin{subfigure}[!]{1.55in}
        \includegraphics[width=1.55in]{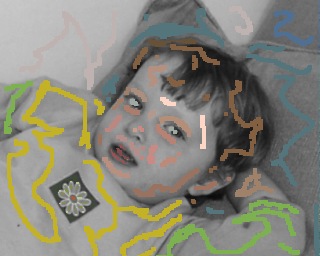}
        \caption{Input}
    \end{subfigure}
    \begin{subfigure}[!]{1.55in}
        \includegraphics[width=1.55in]{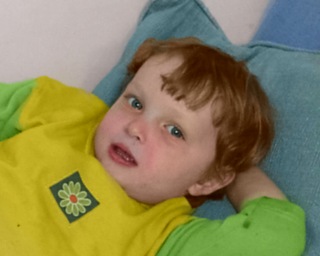}
        \caption{Levin \etal \cite{Levin2004}}
    \end{subfigure}
    \begin{subfigure}[!]{1.55in}
        \includegraphics[width=1.55in]{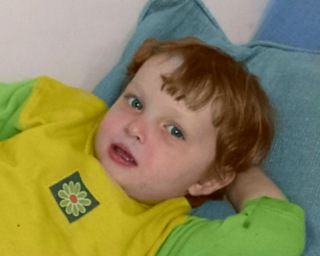}
        \caption{Our results}
    \end{subfigure}
    \caption{
    Results for the user-assisted colorization task.
    Our bilateral solver produces comparable results to the technique of Levin \etal \cite{Levin2004} while being $95\times$ faster.
    \label{fig:colorization}
    }
\end{figure}

Colorization is the problem of introducing color to a grayscale image with a small amount of user input or outside information, for the purpose of improving black-and-white films or photographs.
Levin \etal \cite{Levin2004} presented an effective technique for this task by formulating and solving a specialized optimization problem.
We can solve the same task using our bilateral solver: we use the grayscale image as the input reference image and the UV channels of the user-annotated scribbles as the input target images, with a confidence image that is $1$ where the user has scribbled and $0$ everywhere else.
We then construct our final output by combining the grayscale image with our output UV images, and converting from YUV to RGB.
Our results can be seen in Figure~\ref{fig:colorization}, where we see that our output is nearly indistinguishable from that of \cite{Levin2004}.
The important distinction here is speed, as the approach of \cite{Levin2004} take $80.99$ seconds per megapixel while our approach takes $0.854$ seconds per megapixel --- a $95\times$ speedup.
For all results our parameters are $\sigma_{\mathit{xy}} = \sigma_{\mathit{l}} = \sigma_{\mathit{uv}} = 4$, $\lambda = 0.5$, $\sigma'_{\mathit{xy}} = 4$, $\sigma'_{\mathit{rgb}} = 8$.
More results can be see in the supplement.

\subsection{Semantic Segmentation}

\begin{table}[b!]
\begin{center}
\caption{
  Semantic segmentation results and runtimes on Pascal VOC 2012 validation set.
  The bilateral solver improves performance over the CNN output while being substantially faster than the CRF-based approaches.
  ``Post'' is the time spent post-processing the CNN output, which is the dense CRF for DeepLab, and the generalized CRF-RNN component for CRF-RNN.
  FCN is the convolutional neural network component of the CRF-RNN model.
  $^*$DeepLab-LargeFOV model from \cite{deeplab} trained on Pascal VOC 2012 training data augmented with data from \cite{BharathICCV2011}.
  $^\dagger$CRF-RNN model from \cite{crfasrnn} trained with additional MSCOCO data and evaluated on reduced Pascal validation set of 346 images.
  \label{table:segmentation}
}
\resizebox{2.6in}{!}{
\begin{tabular}{ l | r |r|r|r}
  \multirow{2}{*}{$\mathrm{Method}$} & \multirow{2}{*}{$\mathrm{IOU (\%)}$} &
  \multicolumn{3}{c}{$\mathrm{Time\ (ms)}$} %#CNN time (ms) & Postprocess time (ms) & Total time (ms)\\
  \\
  &&$\mathrm{CNN}$ & $\mathrm{Post}$ & $\mathrm{Total}$\\
\hline
$\mathrm{DeepLab}$ & \cellcolor{Yellow} $62.25^*$ & $58$ & $0$ & \cellcolor{Red}$58$ \\
$\mathrm{DeepLab + CRF}$ & \cellcolor{Red} $67.64^*$ & $58$ & $918$ & \cellcolor{Yellow}$976$\\
$\mathrm{DeepLab + BS (Ours)}$& \cellcolor{Orange} $66.00^*$ & $58$ & $111$ & \cellcolor{Orange} $169$\\
\hline
$\mathrm{CNN}$ & \cellcolor{Yellow} $69.60^\dagger$ & $715$ & $0$ & \cellcolor{Red} $715$\\
$\mathrm{CRF-RNN}$ & \cellcolor{Red} $72.96^\dagger$ & $715$ & $2214$ & \cellcolor{Yellow}$2929$\\
$\mathrm{CNN + BS (Ours)}$&\cellcolor{Orange} $70.68^\dagger$ & $715$ & $217$ & \cellcolor{Orange} $913$
\end{tabular}
}
\end{center}
\end{table}

Semantic segmentation is the problem of assigning a category label to each pixel in an image.
State-of-the-art approaches to semantic segmentation use large convolutional neural networks (CNNs) to map from pixels to labels \cite{deeplab,fcn}.
The output of these CNNs is often smoothed across image boundaries, so recent approaches refine their output with a CRF (\cite{deeplab,crfasrnn}).
These CRF-based approaches improve per-pixel labeling accuracy, but this accuracy comes at a computational cost: inference in a fully connected CRF on a $500\times 500$ image can take up to a second (see Table~\ref{table:segmentation}).
To evaluate whether the bilateral solver could improve the efficiency of semantic segmentation pipelines, we use it instead of the CRF component in two state-of-the-art models: DeepLab-LargeFOV \cite{deeplab} and CRF-RNN \cite{crfasrnn}.
The DeepLab model consists of a CNN trained on Pascal VOC12 and then augmented with a fixed dense CRF.
The CRF-RNN model generalizes the CRF with a recurrent neural network, and trains this component jointly with the CNN on Pascal and MSCOCO.

As the bilateral solver operates on real-valued inputs, it is not immediately clear how to map it onto the discrete optimization problem of the dense CRF.
For each class, we compute the $21$-channel class probability image from the CNN outputs of either the DeepLab or CRF-RNN model.
As many class probability maps are zero across the entire image, the resulting $\mathbf{b}$ matrix in the bilateral solver is often low-rank, allowing us to solve a reduced linear system to recover the smoothed class probability maps (see the supplement).
This approach produces nearly identical output, with a $5\times$ speedup on average.

\begin{figure}[!t]
  \centering
  \begin{subfigure}[!]{1.15in}
        \includegraphics[width=1.15in]{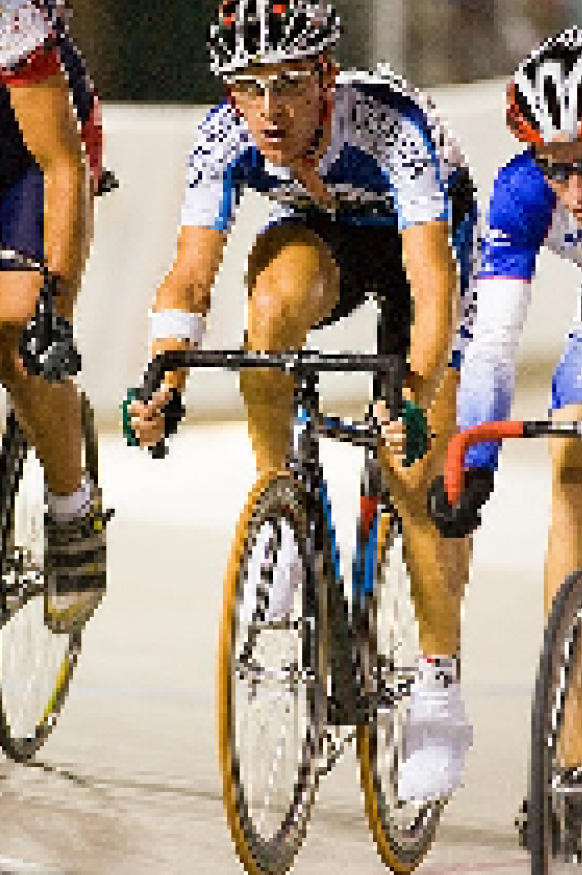}
        \caption{Image \label{fig:seg_exampleA}}
  \end{subfigure}
  \begin{subfigure}[!]{1.15in}
        \includegraphics[width=1.15in]{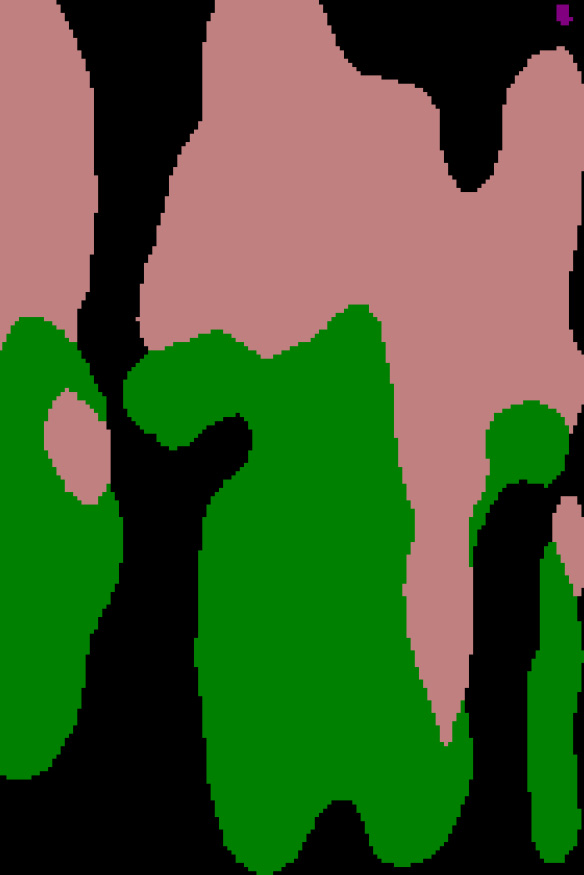}
        \caption{DeepLab \label{fig:seg_exampleB}}
  \end{subfigure}
  \begin{subfigure}[!]{1.15in}
        \includegraphics[width=1.15in]{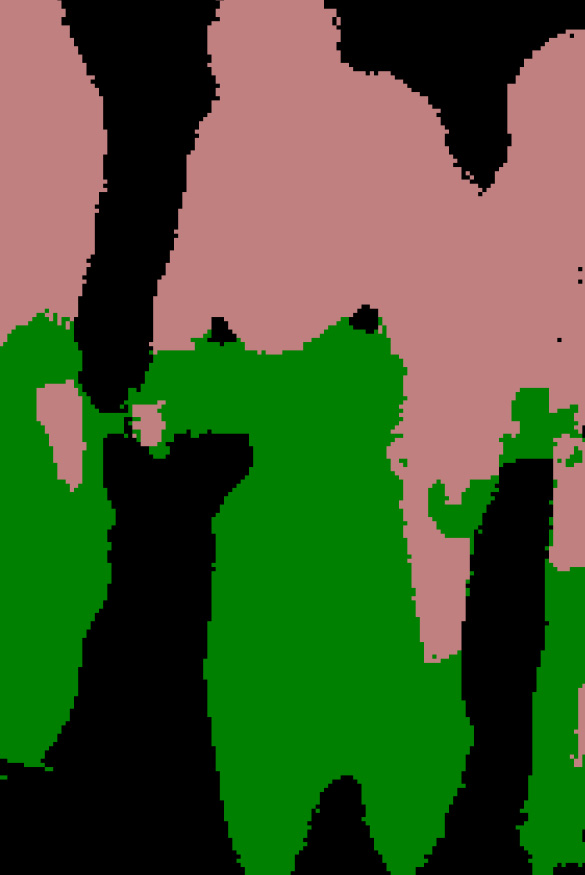}
        \caption{DenseCRF \label{fig:seg_exampleC}}
  \end{subfigure}
  \begin{subfigure}[!]{1.15in}
        \includegraphics[width=1.15in]{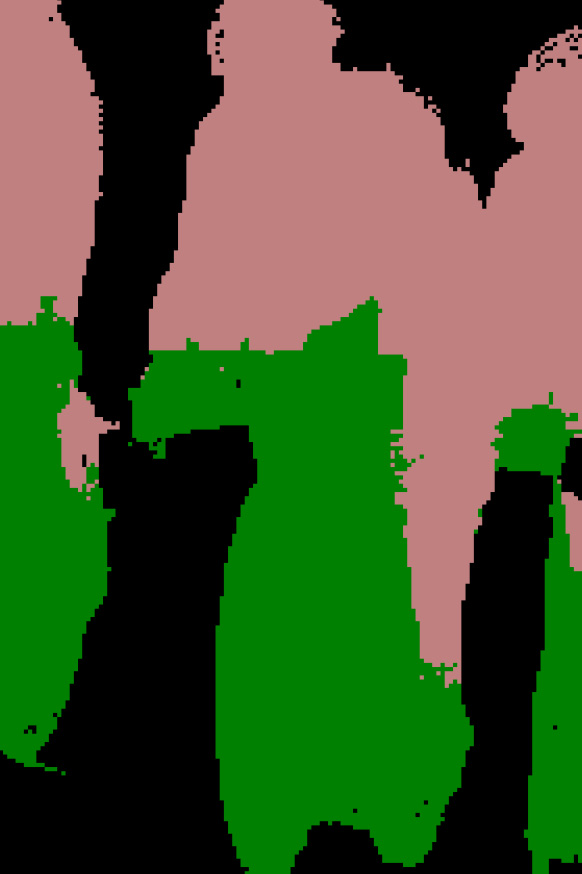}
        \caption{BS (Ours) \label{fig:seg_exampleD}}
  \end{subfigure}
  \caption{
  Using the DeepLab CNN-based semantic segmentation algorithm \cite{deeplab} (\ref{fig:seg_exampleB}) as input our bilateral solver can produce comparable edge-aware output (\ref{fig:seg_exampleD}) to the DenseCRF \cite{KrahenbuhlK11} used in \cite{deeplab} (\ref{fig:seg_exampleC}), while being $8\!\times$ faster.
  \label{fig:seg_example}
  }
\end{figure}

We applied our bilateral solver to the class probability maps using uniform confidence.
The resulting discrete segmentations are more accurate and qualitatively smoother than the CNN outputs, despite our per-channel smoothing providing no explicit smoothness guarantees on the argmax of the filtered per-class probabilities (Table.~\ref{table:segmentation}, Figure \ref{fig:seg_example}).
The bilateral solver is $8-10\times$ faster than the CRF and CRF-RNN approaches when applied to the same inputs (Table~\ref{table:segmentation}).
Although the bilateral solver performs slightly worse than the CRF-based approaches, its speed suggests that it may be a useful tool in contexts such as robotics and autonomous driving, where low latency is necessary.

\section{Conclusion}

We have presented the bilateral solver, a flexible and fast technique for inducing edge-aware smoothness.
We have demonstrated that the solver can produce or improve state-of-the-art results on a variety of different computer vision tasks, while being faster or more accurate than other approaches.
Its speed and generality suggests that the bilateral solver is a useful tool in the construction of computer vision algorithms and deep learning pipelines.

\clearpage

\bibliographystyle{splncs03}
\bibliography{bs}

\end{document}

\pagestyle{headings}
\mainmatter
\title{The Fast Bilateral Solver \\ Supplement}

\titlerunning{The Fast Bilateral Solver --- Supplement}

\authorrunning{Jonathan T. Barron \& Ben Poole}

\author{
Jonathan T. Barron \hspace{0.65in} Ben Poole \hspace{0.15in} \\
\hspace{0.45in} \email{barron@google.com}  \hspace{0.25in} \email{poole@cs.stanford.edu}
}

\institute{}

\maketitle
This supplement provides additional details on the bilateral solver and its extensions in Sections \ref{sec:derivation}--\ref{sec:qr}, and a plethora of experiments and comparisons to prior work in Section \ref{sec:results}.
In Section \ref{sec:derivation} we derive the bilateral-space quadratic objective at the core of the bilateral solver.
Section \ref{sec:pyramids} details the bilateral-space pyramid representation used for faster initialization and preconditioning.
Section \ref{sec:robustness} extends the bilateral solver to use a robust error function to cope with outliers.
Section \ref{sec:qr} improves the performance of the bilateral solver when applied to multiple output channels.
Finally, in Section \ref{sec:results} we provide additional evaluation of the bilateral solver and extensively compare it to numerous existing and novel baseline techniques.

\section{Derivation}
\label{sec:derivation}

In the main paper we presented an optimization problem where we solve for a per-pixel quantity $\mathbf{x}$ subject to a smoothness term that encourages $\mathbf{x}$ to be smooth and a data term that encourages $\mathbf{x}$ to resemble some observed input ``target'' quantities $\mathbf{t}$ proportionally to a per-pixel ``confidence'' $\mathbf{c}$:
\begin{align}
\underset{\mathbf{x}}{\mathrm{minimize}} &\,\, \frac{\lambda}{2} \sum_{i, j} \hat W_{i,j} \left(x_i - x_j \right)^2  + \sum_i c_i (x_i - t_i)^2
\label{eq:pixel_loss}
\end{align}
We can use the procedure detailed in Barron~\etal \cite{Barron2015A} to reformulate the smoothness term into matrix/vector notation.
The data term can be similarly reformulated:
\begin{align}
\sum_i c_i (x_i - t_i)^2 =& (\mathbf{x} - \mathbf{t})^\mathrm{T} \mathrm{diag}(\mathbf{c}) (\mathbf{x} - \mathbf{t}) \\
=& \mathbf{x}^\mathrm{T} \mathrm{diag}(\mathbf{c}) \mathbf{x} - 2\mathbf{x}^\mathrm{T} \mathrm{diag}(\mathbf{c}) \mathbf{t} + \mathbf{t}^\mathrm{T} \mathrm{diag}(\mathbf{c}) \mathbf{t} \\
=& \mathbf{x}^\mathrm{T} \mathrm{diag}(\mathbf{c}) \mathbf{x} - 2 \left(\mathbf{c} \circ \mathbf{t} \right)^\mathrm{T} \mathbf{x} + \left(\mathbf{c} \circ \mathbf{t} \right)^\mathrm{T} \mathbf{t}
\end{align}
Combining this reformulated data term with the reformulated smoothness term from \cite{Barron2015A} we get:
\begin{equation}
\underset{\mathbf{x}}{\mathrm{minimize}} \quad \mathbf{x}^\mathrm{T} \left( \lambda  \left( I - \hat W \right) + \mathrm{diag}(\mathbf{c}) \right) \mathbf{x} - 2 (\mathbf{c} \circ \mathbf{t})^\mathrm{T} \mathbf{x} + (\mathbf{c} \circ \mathbf{t})^\mathrm{T} \mathbf{t}
\label{eq:pixel_prob_mat}
\end{equation}

Using the bistochastization algorithm from \cite{Barron2015A}, we can decompose $\hat W$:
\begin{equation}
\hat W = S^\mathrm{T} D_\mathbf{m}^{-1} D_\mathbf{n} {\bar B} D_\mathbf{n} D_\mathbf{m}^{-1} S
\label{eq:What}
\end{equation}
The bistochastization algorithm from \cite{Barron2015A} is reproduced in Algorithm~\ref{alg:bistoch}.

\begin{algorithm}[!]
\caption{Bilateral-space bistochastization, reproduced from \cite{Barron2015A} \label{alg:bistoch} \\
{\bf Input:} \\
$W = S^\mathrm{T} \bar B S$ \quad // A splat-blur-splice decomposition of $W$ \\
{\bf Output:} \\
$D_\mathbf{n}, D_\mathbf{m}$ \quad // Matrices to bistochastize $W$
}
\begin{algorithmic}[1]
\State $\mathbf{m} \gets S\mathbf{1}$
\State $\mathbf{n} \gets \mathbf{1}$
\While{ not converged}
\State $\mathbf{n} \gets \sqrt{ ( \mathbf{n} \circ \mathbf{m} ) / (\bar B  \mathbf{n} ) }$
\EndWhile
\State $D_\mathbf{n} \gets \mathrm{diag}(\mathbf{n})$
\State $D_\mathbf{m} \gets \mathrm{diag}(\mathbf{m})$
%\State $D_r \gets D_\mathbf{n} D_\mathbf{m}^{-1} = \mathrm{diag}(\mathbf{n} / \mathbf{m})$
%\State $D_b \gets D_\mathbf{m}^{1 \over 2} D_r = \mathrm{diag}( \mathbf{n} / \sqrt{\mathbf{m}} )$
%\State $D_s \gets D_\mathbf{m}^{-{1 \over 2}} = \mathrm{diag}(1 / \sqrt{\mathbf{m}})$
\end{algorithmic}
\end{algorithm}
Using the simplified bilateral grid of \cite{Barron2015A} gives us the following equivalence:
\begin{equation}
SS^\mathrm{T} = D_\mathbf{m} \label{eq:SS}
\end{equation}
We will use the same bilateral-space variable substitution as \cite{Barron2015A}, by rewriting our optimization problem framed in terms of pixels $\mathbf{x}$ as an optimization in terms of bilateral-space vertices $\mathbf{y}$:
\begin{equation}
\mathbf{x} = S^\mathrm{T} \mathbf{y} \label{eq:substitution}
\end{equation}
Let us perform this variable substitution and simplify the resulting expression using our known equivalences, first for the parts of Equation~\ref{eq:pixel_prob_mat} which correspond to the smoothness term:
\begin{align}
\mathbf{x}^\mathrm{T} \left( I - \hat W \right) \mathbf{x} =& \mathbf{x}^\mathrm{T} \left( I - S^\mathrm{T} D_\mathbf{m}^{-1} D_\mathbf{n} {\bar B} D_\mathbf{n} D_\mathbf{m}^{-1} S \right) \mathbf{x} \tag*{by Eq \ref{eq:What}} \\
=& (S^\mathrm{T} \mathbf{y})^\mathrm{T} \left( I - S^\mathrm{T} D_\mathbf{m}^{-1} D_\mathbf{n} {\bar B} D_\mathbf{n} D_\mathbf{m}^{-1} S \right) (S^\mathrm{T} \mathbf{y}) \tag*{by Eq \ref{eq:substitution}} \\
=& \mathbf{y}^\mathrm{T} \left( S I S^\mathrm{T} - S S^\mathrm{T} D_\mathbf{m}^{-1} D_\mathbf{n} {\bar B} D_\mathbf{n} D_\mathbf{m}^{-1} S S^\mathrm{T} \right)  \mathbf{y} \\
=& \mathbf{y}^\mathrm{T} \left( D_\mathbf{m} - D_\mathbf{m} D_\mathbf{m}^{-1} D_\mathbf{n} {\bar B} D_\mathbf{n} D_\mathbf{m}^{-1} D_\mathbf{m} \right) \mathbf{y} \tag*{by Eq \ref{eq:SS}} \\
=& \mathbf{y}^\mathrm{T} \left( D_\mathbf{m} - D_\mathbf{n} {\bar B} D_\mathbf{n} \right) \mathbf{y}
\end{align}
And now, we will perform the variable substitution on the parts of Equation~\ref{eq:pixel_prob_mat} which correspond to the data term:
\begin{align}
 & \mathbf{x}^\mathrm{T} \mathrm{diag}(\mathbf{c}) \mathbf{x} - 2 (\mathbf{c} \circ \mathbf{t})^\mathrm{T} \mathbf{x} + (\mathbf{c} \circ \mathbf{t})^\mathrm{T} \mathbf{t} \\
=& (S^\mathrm{T} \mathbf{y})^\mathrm{T} \mathrm{diag}(\mathbf{c}) (S^\mathrm{T} \mathbf{y}) - 2 (\mathbf{c} \circ \mathbf{t})^\mathrm{T} (S^\mathrm{T} \mathbf{y}) + (\mathbf{c} \circ \mathbf{t})^\mathrm{T} \mathbf{t} \tag*{by Eq \ref{eq:substitution}} \\
=& \mathbf{y}^\mathrm{T} (S \mathrm{diag}(\mathbf{c}) S^\mathrm{T}) \mathbf{y} - 2 (S (\mathbf{c} \circ \mathbf{t}))^\mathrm{T} \mathbf{y} + (\mathbf{c} \circ \mathbf{t})^\mathrm{T} \mathbf{t} \\
=& \mathbf{y}^\mathrm{T} \mathrm{diag}(S \mathbf{c}) \mathbf{y} - 2 (S (\mathbf{c} \circ \mathbf{t}))^\mathrm{T} \mathbf{y} + (\mathbf{c} \circ \mathbf{t})^\mathrm{T} \mathbf{t}
\end{align}
Combining all of this we can rewrite Equation~\ref{eq:pixel_prob_mat} in bilateral-space as follows:
\begin{equation}
\underset{\mathbf{y}}{\mathrm{minimize}} \quad {1 \over 2} \mathbf{y}^\mathrm{T} \left( \lambda \left( D_\mathbf{m} - D_\mathbf{n} {\bar B} D_\mathbf{n} \right) + \mathrm{diag}(S \mathbf{c}) \right) \mathbf{y} - (S (\mathbf{c} \circ \mathbf{t}))^\mathrm{T} \mathbf{y} + {1 \over 2} (\mathbf{c} \circ \mathbf{t})^\mathrm{T} \mathbf{t} \label{eq:pixel_prob_after}
\end{equation}
Unlike the non-linear optimization problem of \cite{Barron2015A}, because we have a simple quadratic optimization problem we can rewrite it in standard form:
\begin{align}
\underset{\mathbf{y}}{\mathrm{minimize}} \quad & {1 \over 2} \mathbf{y}^\mathrm{T} A \mathbf{y} - \mathbf{b}^\mathrm{T} \mathbf{y} + c \label{eq:quad_min} \\
A = \lambda (D_\mathbf{m} - D_\mathbf{n} \bar B D_\mathbf{n}) + \mathrm{diag}(S \mathbf{c}) & \quad\quad \mathbf{b} = S ( \mathbf{c} \circ \mathbf{t} ) \quad\quad c = {1 \over 2} (\mathbf{c} \circ \mathbf{t})^\mathrm{T} \mathbf{t} \nonumber
\end{align}
By taking the derivative of this loss function and setting it to zero we see that minimizing that quadratic form is equivalent to solving the sparse linear system
\begin{equation}
A\mathbf{y} = \mathbf{b}
\label{eq:linear_system}
\end{equation}
We will solve this optimization problem using the preconditioned conjugate gradient (PCG) algorithm, using the initialization and preconditioning tricks described in the paper.
We use the PCG implementation described in Section B3 of \cite{Shewchuk1994}, which is reproduced here in Algorithm~\ref{alg:PCG}.

\begin{algorithm}[!]
\caption{Preconditioned Conjugate Gradients, reproduced from \cite{Shewchuk1994}  \label{alg:PCG}\\
{\bf Input:} \\
$A(\cdot)$ // A function which implements $A\mathbf{x}$ \\
$\mathbf{b}$ // The $\mathbf{b}$ vector in the linear system\\
$\mathbf{x}$ // The initial value of the state $\mathbf{x}$ \\
$M^{-1}(\cdot)$ // A function which implements a preconditioner \\
$n$ // the number of iterations \\
{\bf Output:} \\
$\mathbf{x}$ // $\mathbf{x}$ such that $A(\mathbf{x}) \approx \mathbf{b}$
}
\begin{algorithmic}[1]
\State $i \gets 0$
\State $\mathbf{r} \gets \mathbf{b} - A(\mathbf{x})$
\State $\mathbf{d} \gets M^{-1}(\mathbf{r})$
\State $\lambda_{new} \gets \mathbf{r}^\mathrm{T} \mathbf{d}$
\While{$i < n$}
  \State $\mathbf{q} \gets A(\mathbf{d})$
  \State $\alpha \gets {\lambda_{new} \over \mathbf{d}^\mathrm{T}\mathbf{q}}$
  \State $\mathbf{x} \gets \mathbf{x} + \alpha \mathbf{d}$
  \State $\mathbf{r} \gets \mathbf{r} - \alpha \mathbf{q}$
  \State $\mathbf{s} \gets M^{-1}(\mathbf{r})$
  \State $\lambda_{old} \gets \lambda_{new}$
  \State $\lambda_{new} \gets \mathbf{r}^\mathrm{T} \mathbf{s}$
  \State $\beta \gets {\lambda_{new} \over \lambda_{old}}$
  \State $\mathbf{d} \gets \mathbf{s} + \beta \mathbf{d}$
  \State $i \gets i + 1$
\EndWhile
\end{algorithmic}
\end{algorithm}

\section{Bilateral-Space Pyramids}
\label{sec:pyramids}

Applying the hierarchical initialization and preconditioning techniques in the main paper requires that we have a multiscale representation of our simplified bilateral grid.
As was done in \cite{Barron2015A}, will use the same bilateral grid which was applied to image pixels to instead construct a bilateral grid on top of the vertex coordinates $V$, where $V$ is an $m$ by $5$ matrix produced when constructing a bilateral grid from the input image ($m$ is the number of vertices in the simplified bilateral grid, and $5$ is the dimensionality of our XYLUV bilateral-space).
From $V$ we can construct a more coarse simplified bilateral grid by dividing the elements of $V$ by $2$ (an arbitrary scale factor), and then repeat that procedure to form a pyramid:
\begin{algorithmic}
\For{$k = [0 : K-1]$}
\State $V \gets V/2$
\State $(S_k, V) \gets \mathrm{simplified\_bilateral\_grid}(V)$
\EndFor
\end{algorithmic}
Where $K$ (the number of levels of the pyramid) is set such that the top of the pyramid contains just a single vertex.
As was done in \cite{Barron2015A}, we can use these $K$ splat matrices to lift from some bilateral-space vector $\mathbf{y}$ into a pyramid-space:
\begin{equation}
P(\mathbf{y}) = [S_{K-1} \ldots S_1 S_0 \mathbf{y}, \ldots, S_1 S_0 \mathbf{y}, S_0 \mathbf{y}, \mathbf{y}]
\end{equation}
We can transpose this pyramid operation, collapsing back down to bilateral-space:
\begin{equation}
P^\mathrm{T}(\mathbf{z}) = [S_0^\mathrm{T} S_1^\mathrm{T} \ldots S_{K-1}^\mathrm{T}  \mathbf{z}, \ldots, S_0^\mathrm{T} S_1^\mathrm{T} \mathbf{z}, S_0^\mathrm{T} \mathbf{z}, \mathbf{z}]
\end{equation}
We compute $P(\mathbf{y})$ and $P^\mathrm{T}(\mathbf{z})$ efficiently from the bottom up and the top down, respectively, by reusing the information from the previous scale.

\section{Robustness}
\label{sec:robustness}

Though the quadratic data term of Equation~\ref{eq:pixel_loss} enables our fast least-squares formulation, it has the natural consequence that the bilateral solver is sensitive to outliers in the input target (unless those outliers have been assigned a low confidence).
For some applications where the input target has non-Gaussian noise we may wish to be robust to outliers.
We therefore present a robustified variant of our bilateral solver, that we will call the ``robust bilateral solver'' (RBS).
The RBS minimizes the following loss:
\begin{align}
\underset{\mathbf{x}}{\mathrm{minimize}} &\,\, \frac{\lambda}{2} \sum_{i, j} \hat W_{i,j} \left(x_i - x_j \right)^2  + \sum_i \rho(x_i - t_i)
\label{eq:robust_loss}
\end{align}
Where $\rho(\cdot)$ is some robust error function.
Unless $\rho(\cdot)$ is defined as (weighted) squared-error like in Equation~\ref{eq:pixel_loss}, the optimization problem in Equation~\ref{eq:robust_loss} does not have a closed-form solution.
However, Equation~\ref{eq:robust_loss} can be solved using iteratively reweighted least squares (IRLS) \cite{beaton1974} by repeatedly linearizing the loss function around the current estimate of $\mathbf{x}$ and then solving a least-squares problem corresponding to that linearization.
The linearized version of Equation~\ref{eq:robust_loss} in IRLS takes on exactly the same form as Equation~\ref{eq:pixel_loss}, where the ``confidence'' $\mathbf{c}$ is replaced by the ``weight'' generated during IRLS.
Thus we can produce a robust bilateral solver by wrapping the standard bilateral solver in a loop, recomputing the weight at each iteration.

The RBS can be used with any standard robust loss function that would be used for M-estimation.
We use the Geman-McClure loss function \cite{geman1987}, a smooth approximation to the $\ell_0$-norm, whose estimator $\rho(\cdot)$ and corresponding weight in IRLS are:
\begin{equation}
\rho(e_i) = { e_i^2 \over \sigma_{\mathit{gm}}^2 + e_i^2 } \quad\quad w(e_i) = { 2 \sigma_{\mathit{gm}}^2 \over (\sigma_{\mathit{gm}}^2 + e_i^2)^2 }
\end{equation}
where $\sigma_{\mathit{gm}}$ is a scale parameter.
Pseudocode for the RBS with this Geman-McClure loss can be found in Algorithm~\ref{alg:RBS}.
\begin{algorithm}[!]
\caption{The Geman-McClure Robust Bilateral Solver \label{alg:RBS} \\
{\bf Input:} \\
$\mathrm{solve}(\cdot)$ // The bilateral solver \\
$\mathbf{t}$ // The input target vector \\
$\mathbf{c}_\mathit{init}$ // The input confidence vector \\
$\sigma_{\mathit{gm}}$ // The scale parameter of our Geman-McClure function \\
$n$ // The number of IRLS iterations \\
{\bf Output:} \\
$\mathbf{\hat x}$ // $\mathbf{x}$ such that Equation~\ref{eq:robust_loss} is minimized
}
\begin{algorithmic}[1]
\State $\mathbf{c} \gets \mathbf{c}_\mathit{init}$
\While{$i < n$}
  \State $\mathbf{\hat x} \gets \mathrm{solve}(\mathbf{t}, \mathbf{c} )$
  \State $\mathbf{e} \gets (\mathbf{\hat x} - \mathbf{t} )$
  \State $\mathbf{c} \gets { 2 \sigma_{\mathit{gm}}^2 \over \left(\sigma_{\mathit{gm}}^2 + \left(\mathbf{e} \circ \mathbf{e} \right) \right)^2 }$
  \State $i \gets i + 1$
\EndWhile
\end{algorithmic}
\end{algorithm}

\begin{figure}[!]
\centering
  \begin{subfigure}[!]{2.3in}
    \includegraphics[width=2.3in]{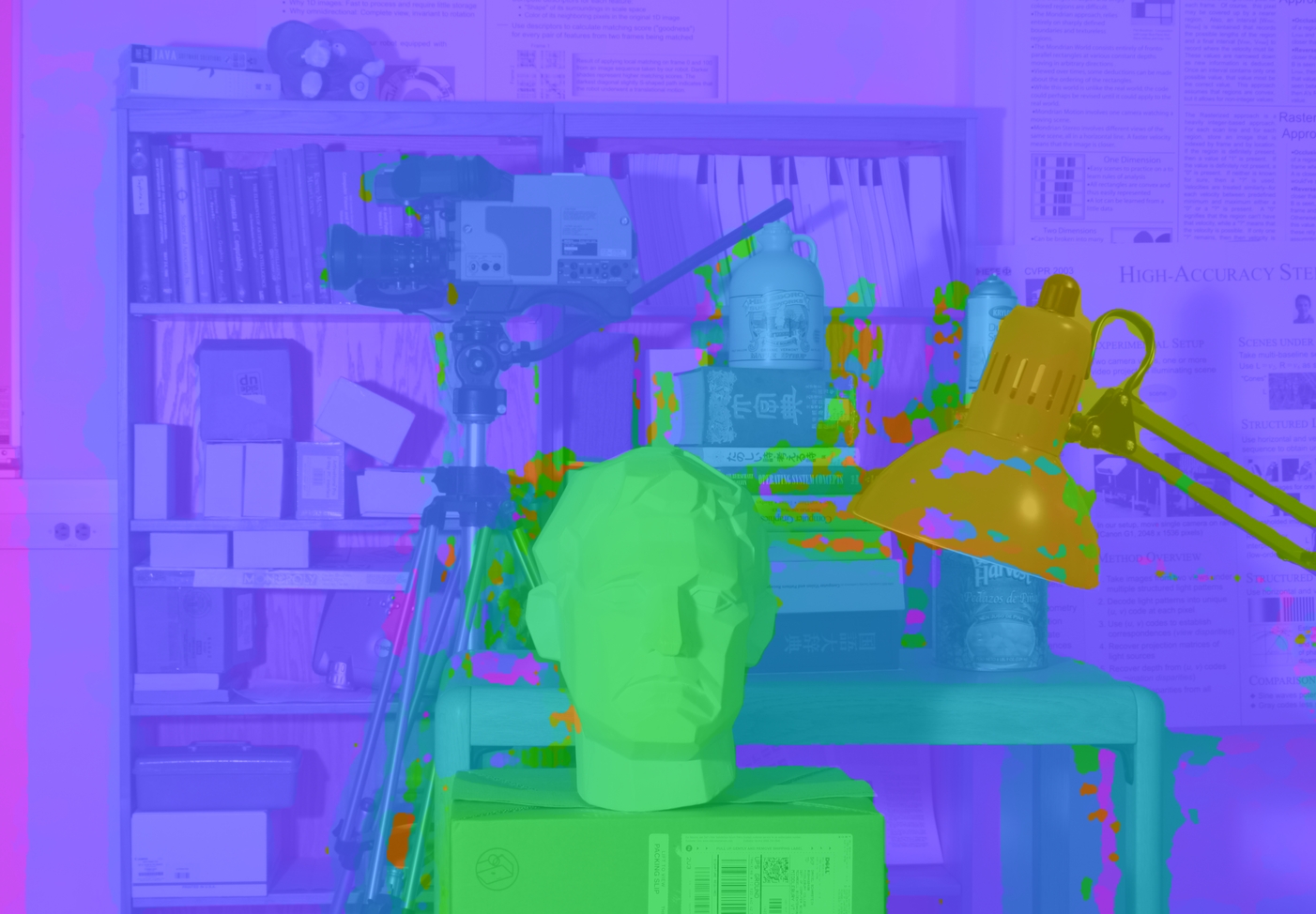}
    \caption{Input depth map $Z$ from MC-CNN\cite{Zbontar2015}
    }
  \end{subfigure}
  \begin{subfigure}[!]{2.3in}
    \includegraphics[width=2.3in]{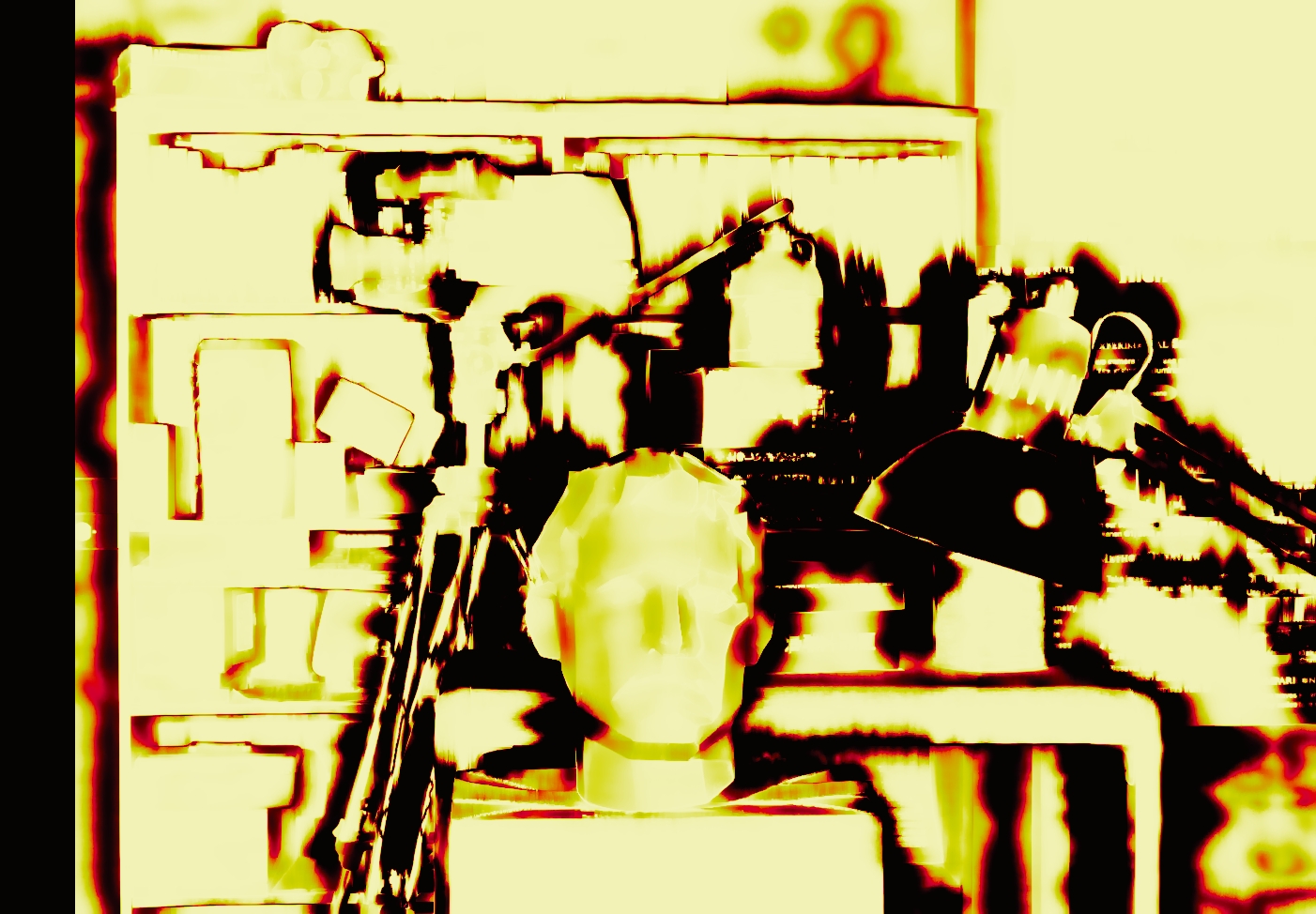}
    \caption{Estimated input confidence $\mathbf{c}_\mathit{init}$
    }
  \end{subfigure}
  \caption{
  When processing the depth maps produced by other stereo algorithms, we first use an edge-aware variance estimation technique to produce initial confidence measures used by our robust bilateral solver.
  This procedure downweights contiguous image regions with inconsistent depths.
  \label{fig:variance}
  }
\end{figure}

Because our loss function is non-convex and therefore sensitive to initialization, our RBS interface takes as input some initial confidence $\mathbf{c}_\mathit{init}$ that is used in the first least-squares solve, and then overwritten by the IRLS weights in subsequent iterations.
Because the IRLS / robust M-estimation loop of the RBS is sensitive to initialization, we can improve performance by setting the initial weights used in the IRLS loop $c_{\mathit{init}}$ to reflect some noise model computed from the input to the solver.
In the case of our Middlebury stereo experiment, we construct this initial confidence according to the heuristic observation that accurate depth maps tend to have contiguous image regions with low-variance depths.
To identify such regions we compute an edge-aware measure of depth variance on the input depth map $Z$ using the recursive formulation of the domain transform \cite{GastalOliveira2011DomainTransform}, which is a simple and fast edge-aware filter.
To see how we do this, let us first review the definition of variance:
\begin{equation}
\operatorname{Var}(x) = \operatorname{E}[x^2] - \left(\operatorname{E}[x]\right)^2
\end{equation}
Where $\operatorname{E}$ is the expectation operator.
Any (normalized) linear image filtering operation with non-negative weights can be thought of as computing some local expectation of the input image for every pixel.
When coupled with an edge-aware image filtering operation such as the domain transform (DT), we can compute a local edge-aware variance $V$ of our input depth map $Z$ as follows:
\begin{equation}
V = \mathrm{DT}( Z^2 ) - \left(\mathrm{DT}( Z )\right)^2 \\
\end{equation}
We initialize our input confidence by exponentiating this scaled, negated variance:
\begin{equation}
\mathbf{c}_\mathit{init} = \exp\left( - {V \over 2 \sigma_{dt}^2 }   \right)
\end{equation}
Our parameters are tuned to maximize performance on the Middlebury training set: $\sigma_{xy} = \sigma_{rgb} = 32$, $\sigma_{dt} = 2$.
We additionally modify this initial confidence using the observation that the depth map at one side of an image of each stereo pair generally has a poorly estimated depth, as the true match for those pixels are often not present in the other image of the stereo pair.
We address this by setting the leftmost $80$ columns of $\mathbf{c}_\mathit{init}$ to $0$, thereby causing them to be initially ignored.
See Figure~\ref{fig:variance} for a visualization of the initial confidence estimated using this procedure on one of the test-set depth maps from the Middlebury stereo benchmark v3.

\section{Multiple Output Channels}
\label{sec:qr}

The tasks which we apply our bilateral solver to consist of inputs with a single channel (depth, stereo, etc) or many channels ($21$, for our semantic segmentation benchmark).
As mentioned previously, our model generalizes straightforwardly to problems with $n$-channel ``target'' inputs by simply decomposing into $n$ independent optimization problems with the same $A$ matrix and different $\mathbf{b}$ vectors. % which can be solved in parallel.
By concatenating these $\mathbf{y}$ and $\mathbf{b}$ vectors into matrices, this can be rewritten as one large linear system:
\begin{equation}
AY = B
\label{eq:matrix_linear_system}
\end{equation}
For applications such as semantic segmentation this $B$ matrix is often very wide, but also very low-rank --- many object categories may have extremely low probabilities in the input, and some object categories may be strongly correlated.
In these cases we can produce a reduced linear system with fewer right-hand-sides by producing a low-rank approximation to $B$, solving the resulting linear system, and then expanding our low-rank solution back to input space.
This can dramatically speed up convergence, often by an order of magnitude, as many semantic segmentation algorithms often only assign a non-trivial probability to a small fraction of object categories for any given scene.
Our approach is similar to a simplified and approximate version of ``block'' conjugate gradient \cite{Oleary1980}.

We use a rank-revealing QR factorization \cite{Chan1987} to reduce our $B$ matrix, which is often used for similar tasks due to its stability and speed.
\begin{equation}
BP=QR
\end{equation}
Where $P$ is a permutation matrix, $Q$ is an orthonormal basis for the columns of $B$, and $R$ is an upper triangular matrix.
With this, let us construct a modified factorization of $B$:
\begin{equation}
B = Q' R'
\end{equation}
Where we construct $R'$ by taking $R P^\mathrm{T}$ and then shuffling the rows of the resulting matrix such the rows of $R'$ have non-increasing Euclidean norms.
$Q'$ is $Q$ where the columns of $Q$ have also been shuffled.
Let us define a vector $\mathbf{m}$ which is the ``mass'' (squared Euclidean norm) of each row of $R'$:
\begin{equation}
m_i = \sum_j {R'}_{i,j}^2
\end{equation}
Let us define some tolerance $\epsilon$, which is an upper bound on the residual tolerance fraction of $B$ that we are willing to tolerate.
We find the largest value of $t$ such that:
\begin{equation}
\sum_{i=0}^{i < t} m_i \leq (1 - \epsilon) \sum_i m_i
\end{equation}
With this we can drop the least important rows of R and columns of Q:
\begin{equation}
\tilde{Q} \approx Q'[:,1\!:\!t] \quad\quad\quad \tilde{R} \approx (R' P^\mathrm{T})[1\!:\!t, :]
\end{equation}
We can use $\tilde{Q}$ to solve the reduced linear system, and then multiply the solution to that linear system by $\tilde{R}$ to approximate the solution to the original system:
\begin{equation}
Y = (A^{-1} \tilde{Q}) \tilde{R}
\end{equation}
In our semantic segmentation experiment, For small values of $\epsilon = 0.01$ the output from this approximate is often indistinguishable from exact solution, despite being $\approx 5\times$ faster on average.

\section{Results}
\label{sec:results}

Here we present many additional results for the four tasks explored in the main paper, in the form of additional figures and expanded tables of results, as well as details regarding the evaluation of baseline techniques.

\subsection{Stereo}

\begin{figure*}[p]
\centering
  \begin{subfigure}[!]{4.8in}
    \includegraphics[width=3.72in]{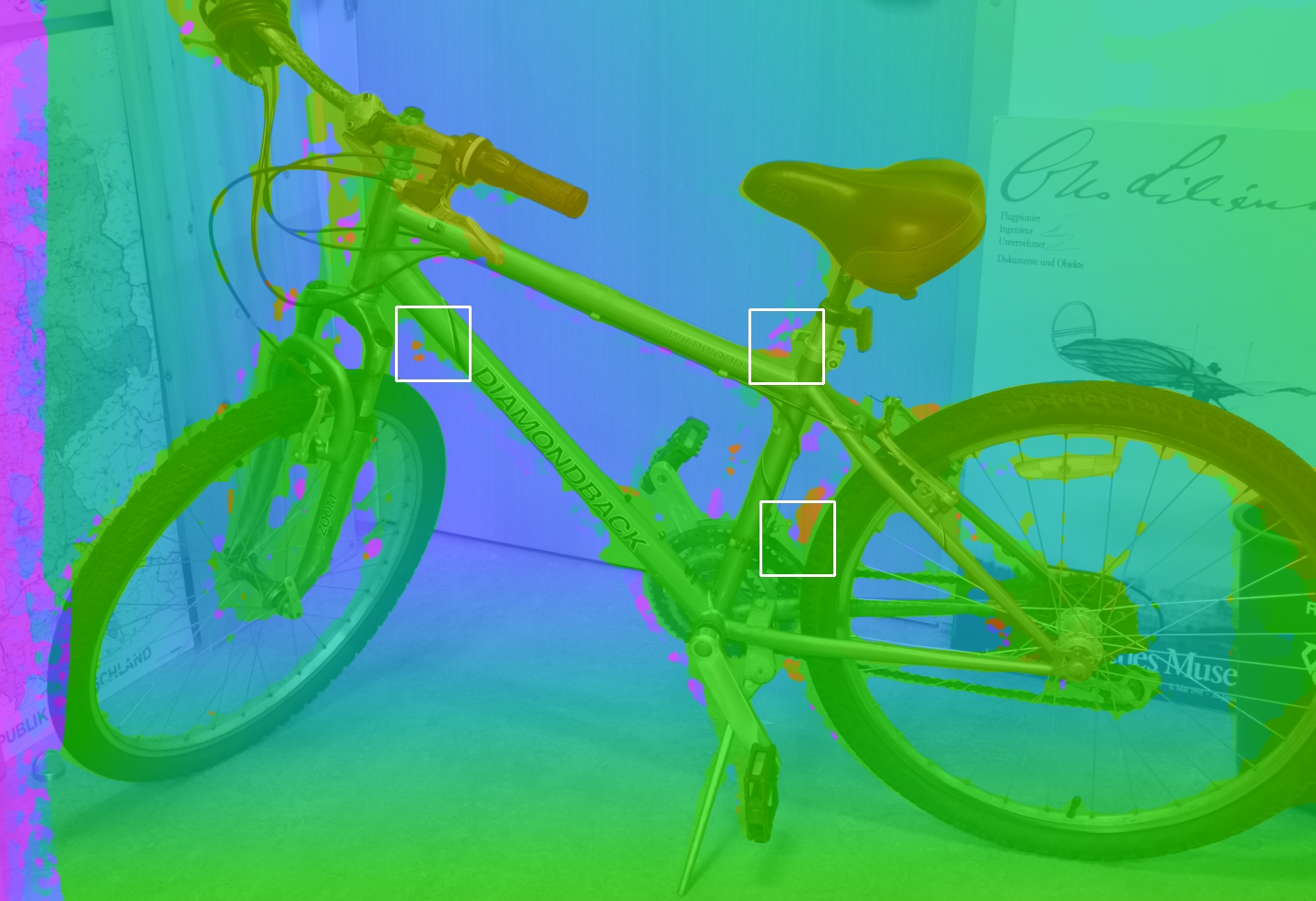}
    \includegraphics[width=0.85in]{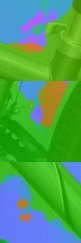}
    \caption{
    MC-CNN\cite{Zbontar2015} \\
    bad 1\% = $10.1$ \quad
    MAE = $1.69$ \quad
    RMSE = $9.60$
    }
  \end{subfigure}
  \\
  \begin{subfigure}[!]{4.8in}
    \includegraphics[width=3.72in]{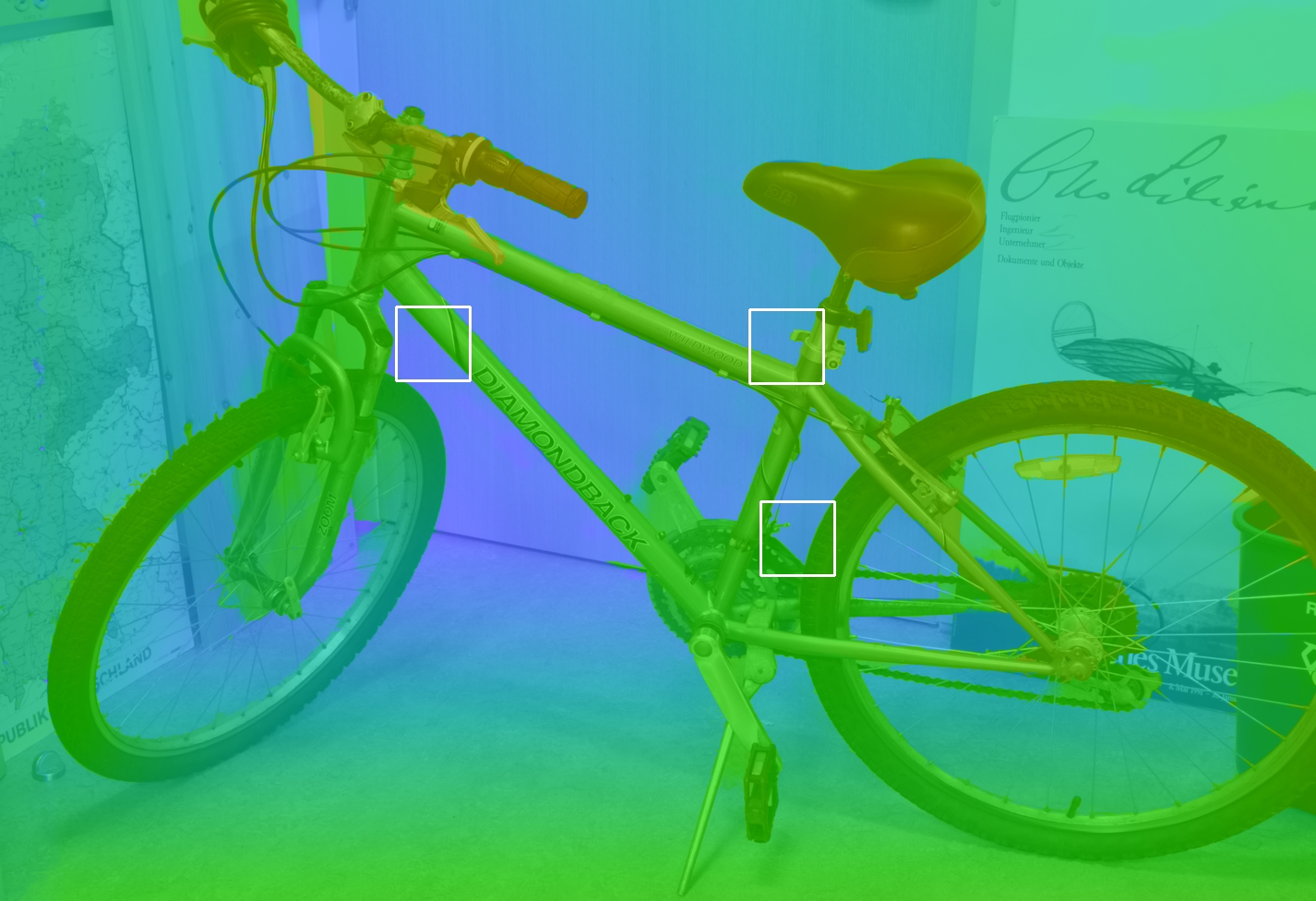}
    \includegraphics[width=0.85in]{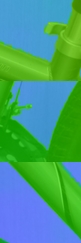}
    \caption{
    MC-CNN\cite{Zbontar2015} + RBS \\
    bad 1\% = $10.7$ \quad
    MAE = $1.63$ \quad
    RMSE = $8.72$
    }
  \end{subfigure}
  \caption{
  Results on the test set of the Middlebury Stereo Dataset V3 \cite{Scharstein2014} where our robust bilateral solver is used to improve the depth map predicted by the state-of-the-art MC-CNN technique\cite{Zbontar2015}.
  On the top we have the depth map produced by \cite{Zbontar2015} (with zoomed in regions) which is used as the target in our solver.
  On the bottom we have the output of our solver, where we see that quality is significantly improved.
  We report the error for each depth map in terms of the percent of pixels whose disparity is off by more than 1 (``bad 1\%''), the mean-absolute-error of the disparity, and the root-mean-squared-error of the disparity.
  \label{fig:testset_middlebury1}
  }
\end{figure*}

\begin{figure*}[p]
\centering
  \begin{subfigure}[!]{4.8in}
    \includegraphics[width=3.8in]{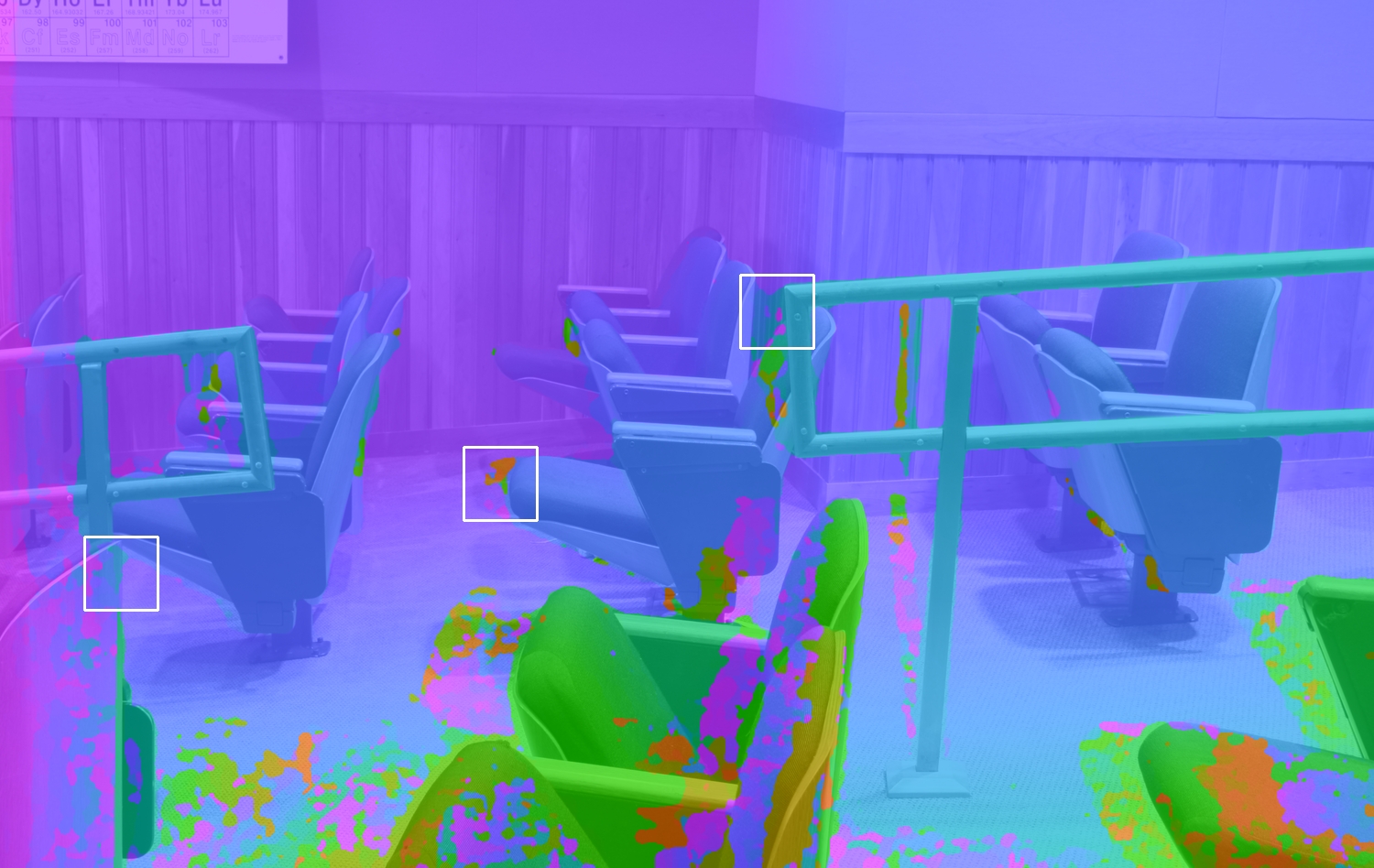}
    \includegraphics[width=0.80in]{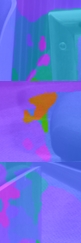}
    \caption{
    MC-CNN\cite{Zbontar2015} \\
    bad 1\% = $23.4$ \quad
    MAE = $17.1$ \quad
    RMSE = $67.4$
    }
  \end{subfigure}
  \\
  \begin{subfigure}[!]{4.8in}
    \includegraphics[width=3.8in]{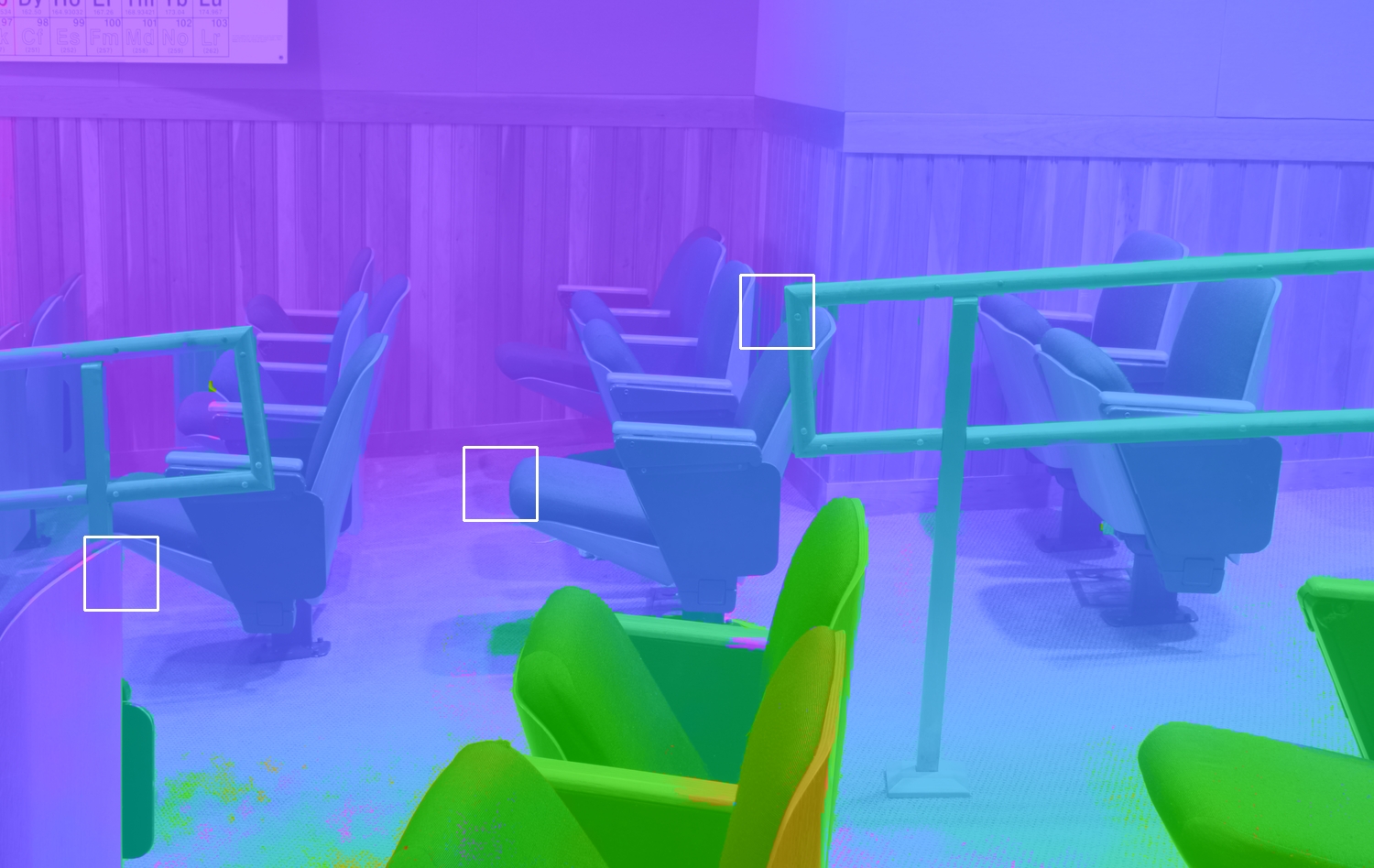}
    \includegraphics[width=0.80in]{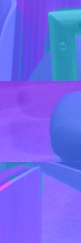}
    \caption{
    MC-CNN\cite{Zbontar2015} + RBS \\
    bad 1\% = $24.4$ \quad
    MAE = $4.30$ \quad
    RMSE = $19.9$
    }
  \end{subfigure}
  \caption{More results on the test set of the Middlebury Stereo Dataset V3 \cite{Scharstein2014} in the same format as Figure~\ref{fig:testset_middlebury1}.
  \label{fig:testset_middlebury2}
  }
\end{figure*}

\begin{figure*}[p]
\centering
  \begin{subfigure}[!]{4.8in}
    \includegraphics[width=3.76in]{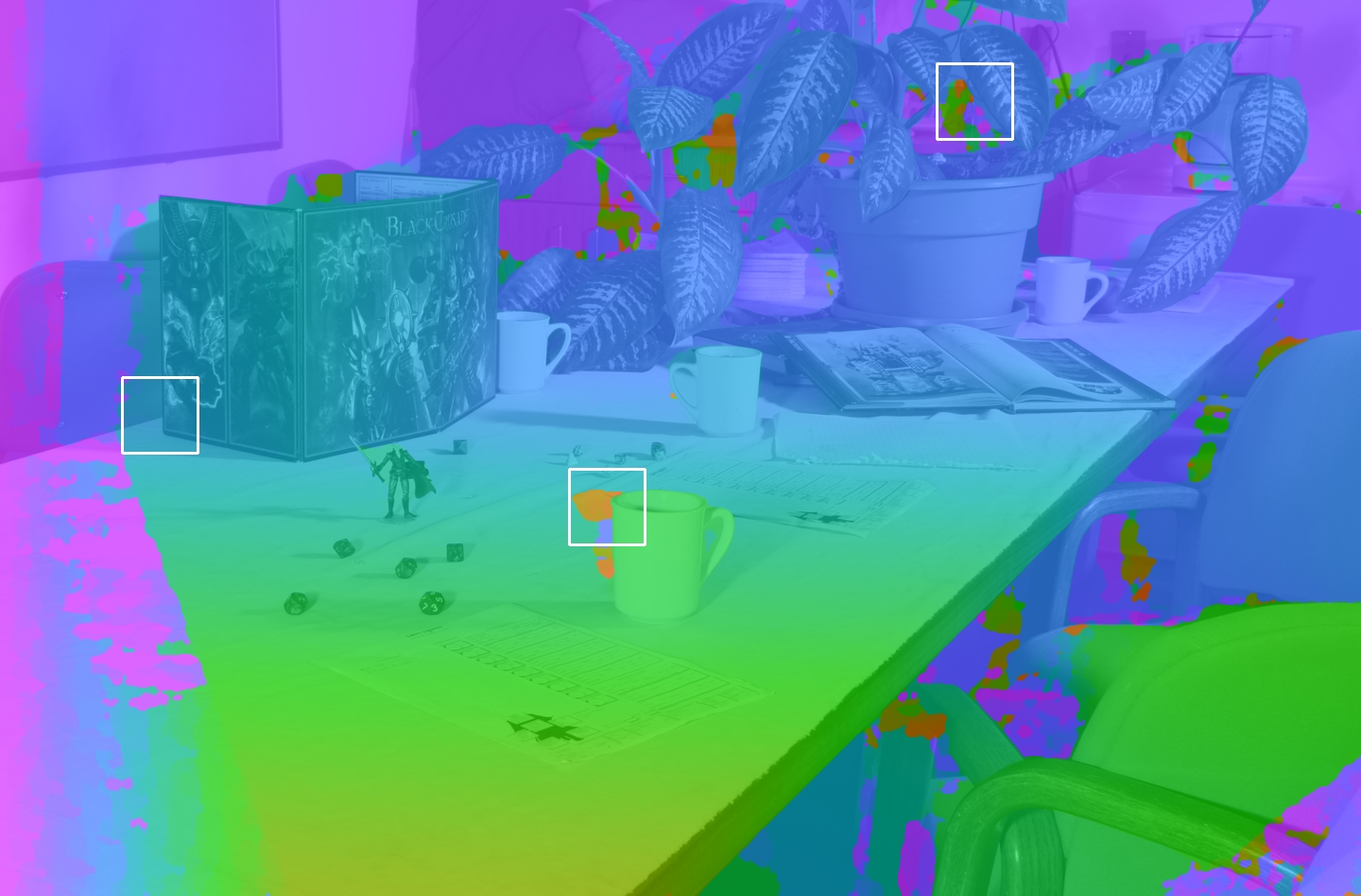}
    \includegraphics[width=0.83in]{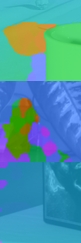}
    \caption{
    MC-CNN\cite{Zbontar2015} \\
    bad 1\% = $25.0$ \quad
    MAE = $2.47$ \quad
    RMSE = $23.2$
    }
  \end{subfigure}
  \\
  \begin{subfigure}[!]{4.8in}
    \includegraphics[width=3.76in]{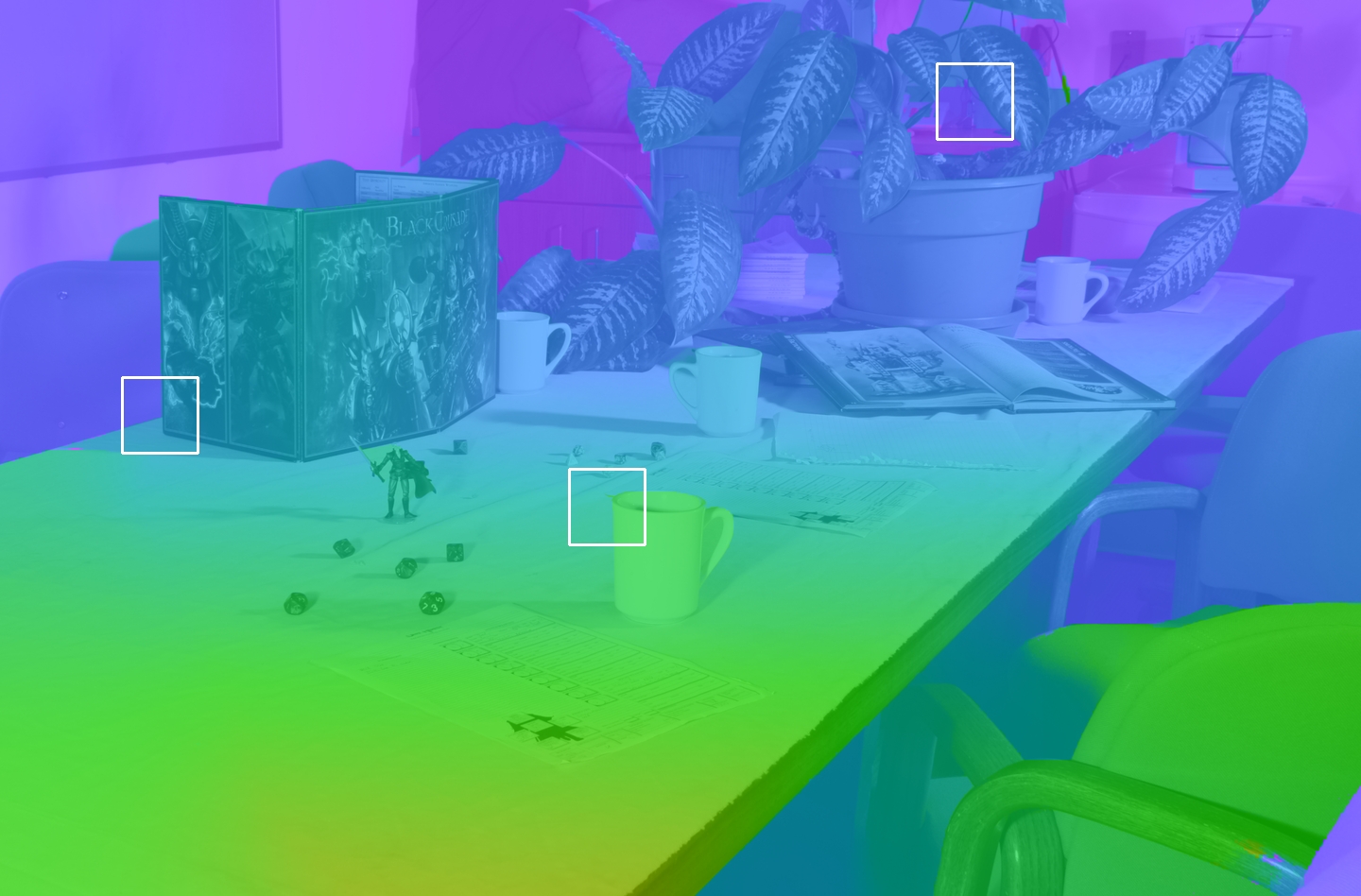}
    \includegraphics[width=0.83in]{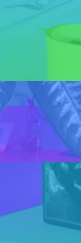}
    \caption{
    MC-CNN\cite{Zbontar2015} + RBS \\
    bad 1\% = $27.1$ \quad
    MAE = $2.81$ \quad
    RMSE = $24.2$
    }
  \end{subfigure}
  \caption{More results on the test set of the Middlebury Stereo Dataset V3 \cite{Scharstein2014} in the same format as Figure~\ref{fig:testset_middlebury1}.
  \label{fig:testset_middlebury3}
  }
\end{figure*}

\begin{figure*}[p]
\centering
  \begin{subfigure}[!]{4.8in}
    \includegraphics[width=3.7in]{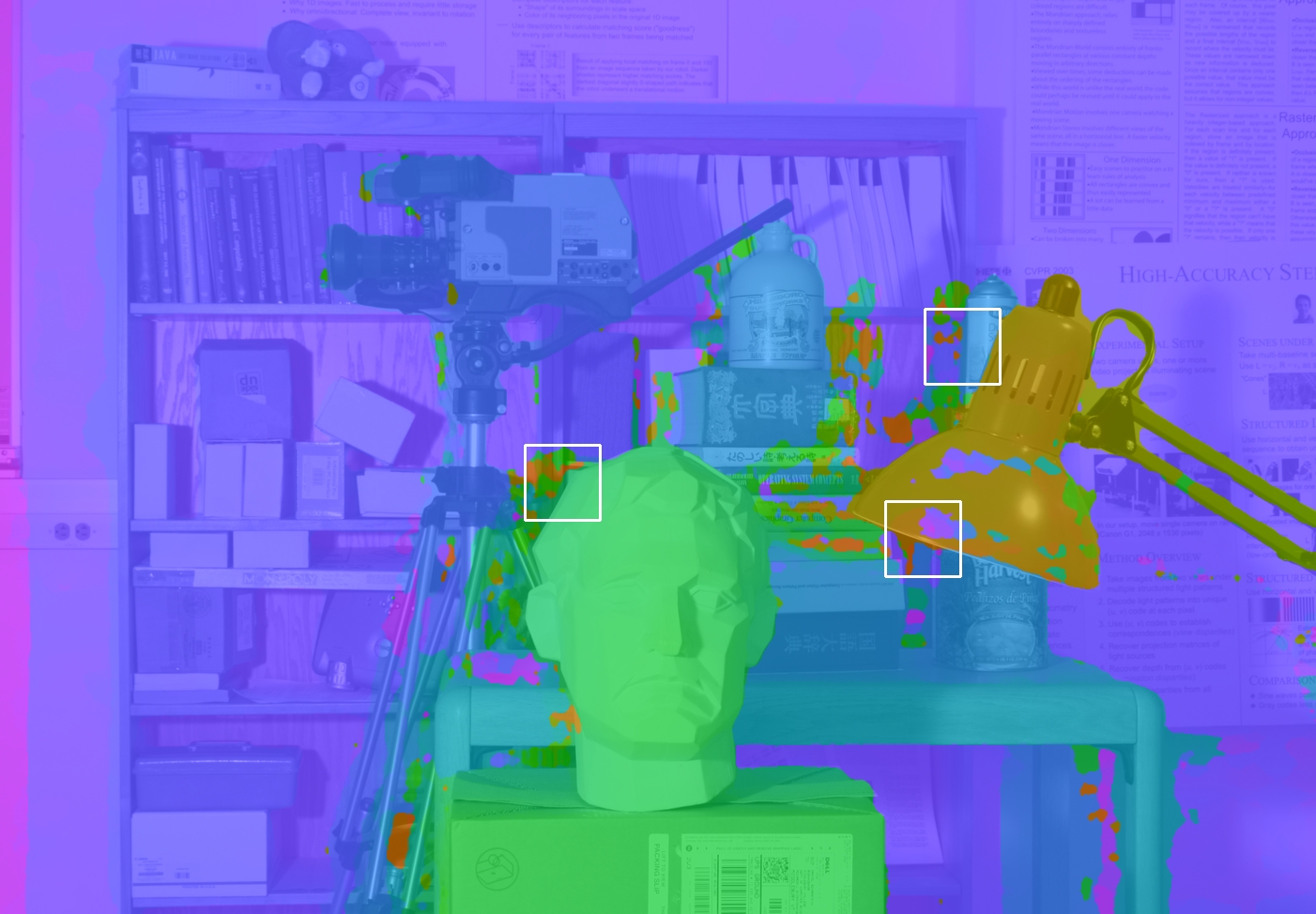}
    \includegraphics[width=0.86in]{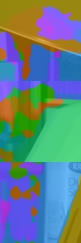}
    \caption{
    MC-CNN\cite{Zbontar2015} \\
    bad 1\% = $22.5$ \quad
    MAE = $6.00$ \quad
    RMSE = $38.8$
    }
  \end{subfigure}
  \\
  \begin{subfigure}[!]{4.8in}
    \includegraphics[width=3.7in]{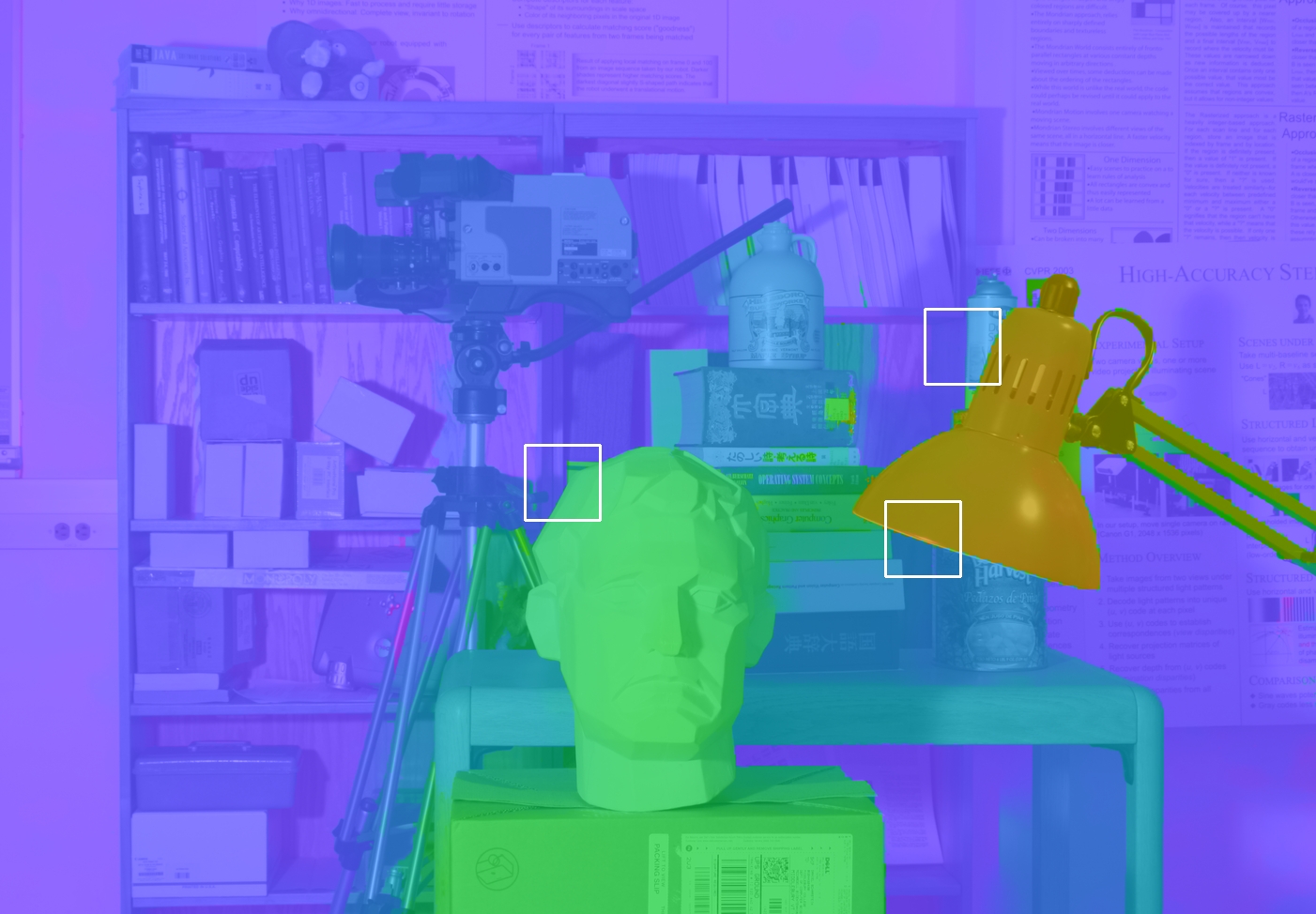}
    \includegraphics[width=0.86in]{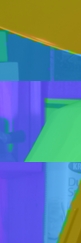}
    \caption{
    MC-CNN\cite{Zbontar2015} + RBS \\
    bad 1\% = $22.8$ \quad
    MAE = $3.02$ \quad
    RMSE = $17.9$
    }
  \end{subfigure}
  \caption{More results on the test set of the Middlebury Stereo Dataset V3 \cite{Scharstein2014} in the same format as Figure~\ref{fig:testset_middlebury1}.
  \label{fig:testset_middlebury4}
  }
\end{figure*}

\begin{figure*}[p]
\centering
  \begin{subfigure}[!]{4.8in}
    \includegraphics[width=3.76in]{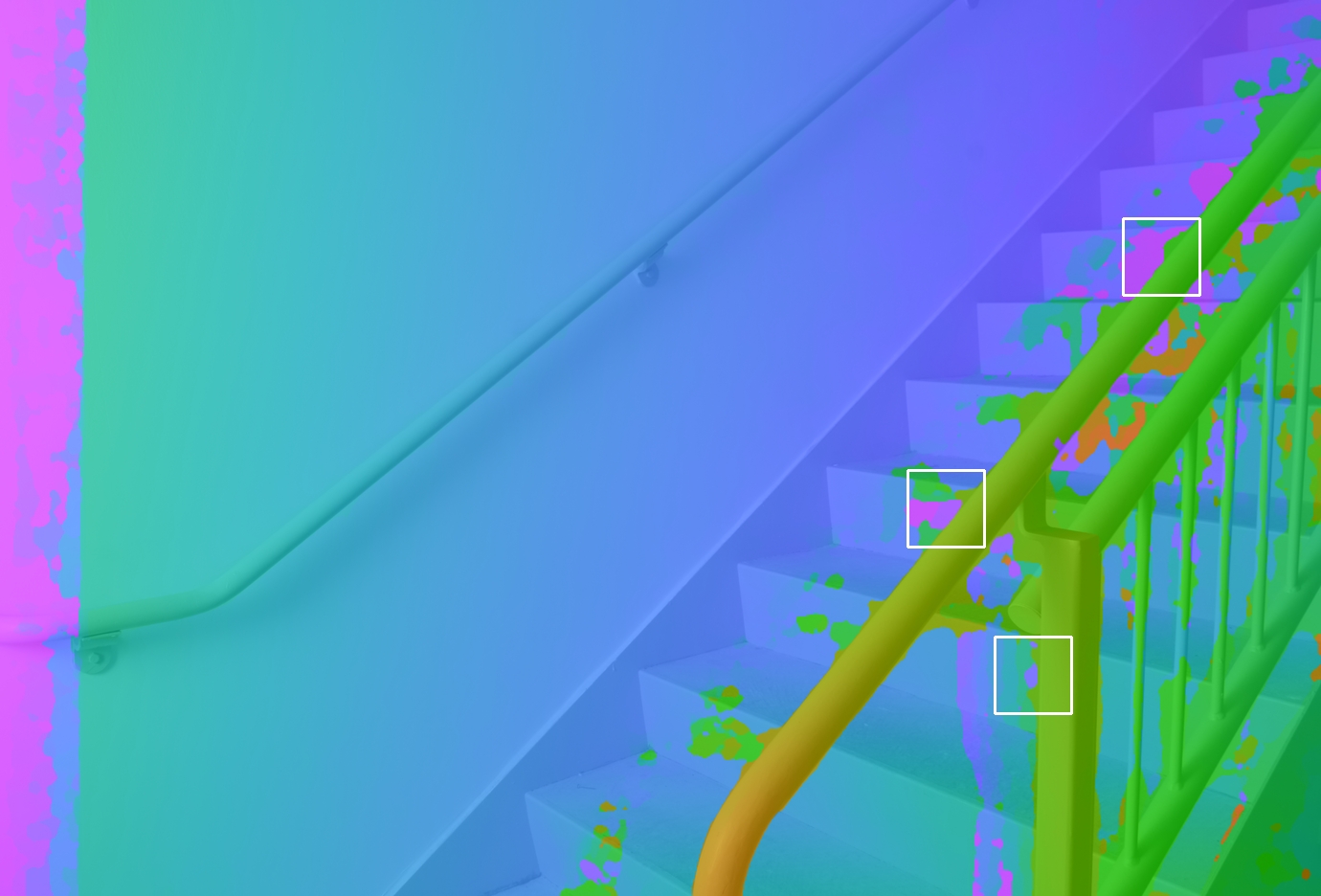}
    \includegraphics[width=0.85in]{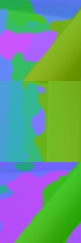}
    \caption{
    MC-CNN\cite{Zbontar2015} \\
    bad 1\% = $22.0$ \quad
    MAE = $5.66$ \quad
    RMSE = $28.6$
    }
  \end{subfigure}
  \\
  \begin{subfigure}[!]{4.8in}
    \includegraphics[width=3.76in]{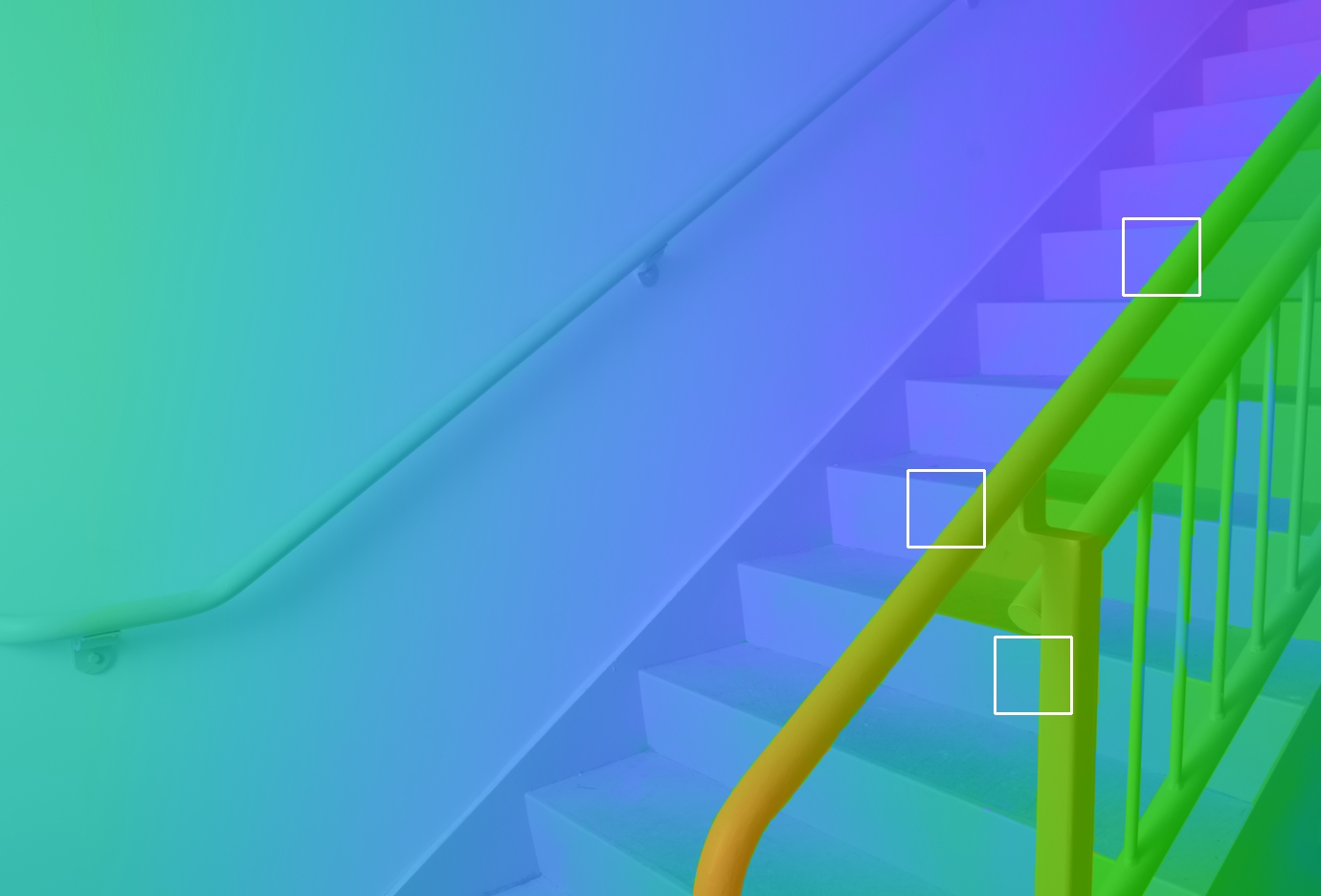}
    \includegraphics[width=0.85in]{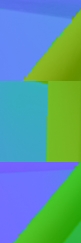}
    \caption{
    MC-CNN\cite{Zbontar2015} + RBS \\
    bad 1\% = $21.6$ \quad
    MAE = $3.19$ \quad
    RMSE = $17.9$
    }
  \end{subfigure}
  \caption{More results on the test set of the Middlebury Stereo Dataset V3 \cite{Scharstein2014} in the same format as Figure~\ref{fig:testset_middlebury1}.
  \label{fig:testset_middlebury5}
  }
\end{figure*}

\begin{figure*}[p]
\centering
  \begin{subfigure}[!]{2.3in}
    \includegraphics[width=1.83in]{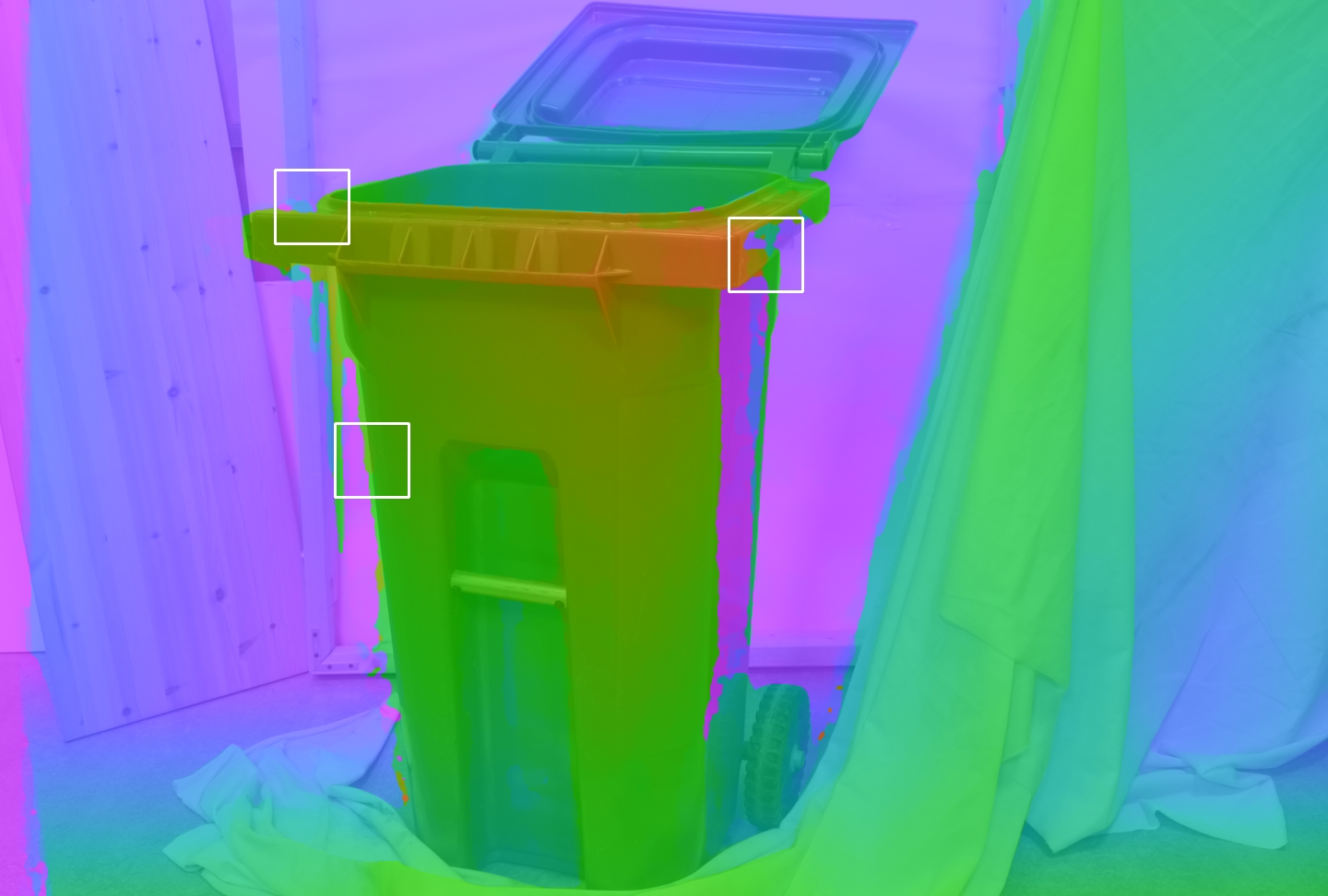}
    \includegraphics[width=0.41in]{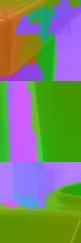}
    \caption{MC-CNN\cite{Zbontar2015}}
  \end{subfigure}
  \begin{subfigure}[!]{2.3in}
    \includegraphics[width=1.83in]{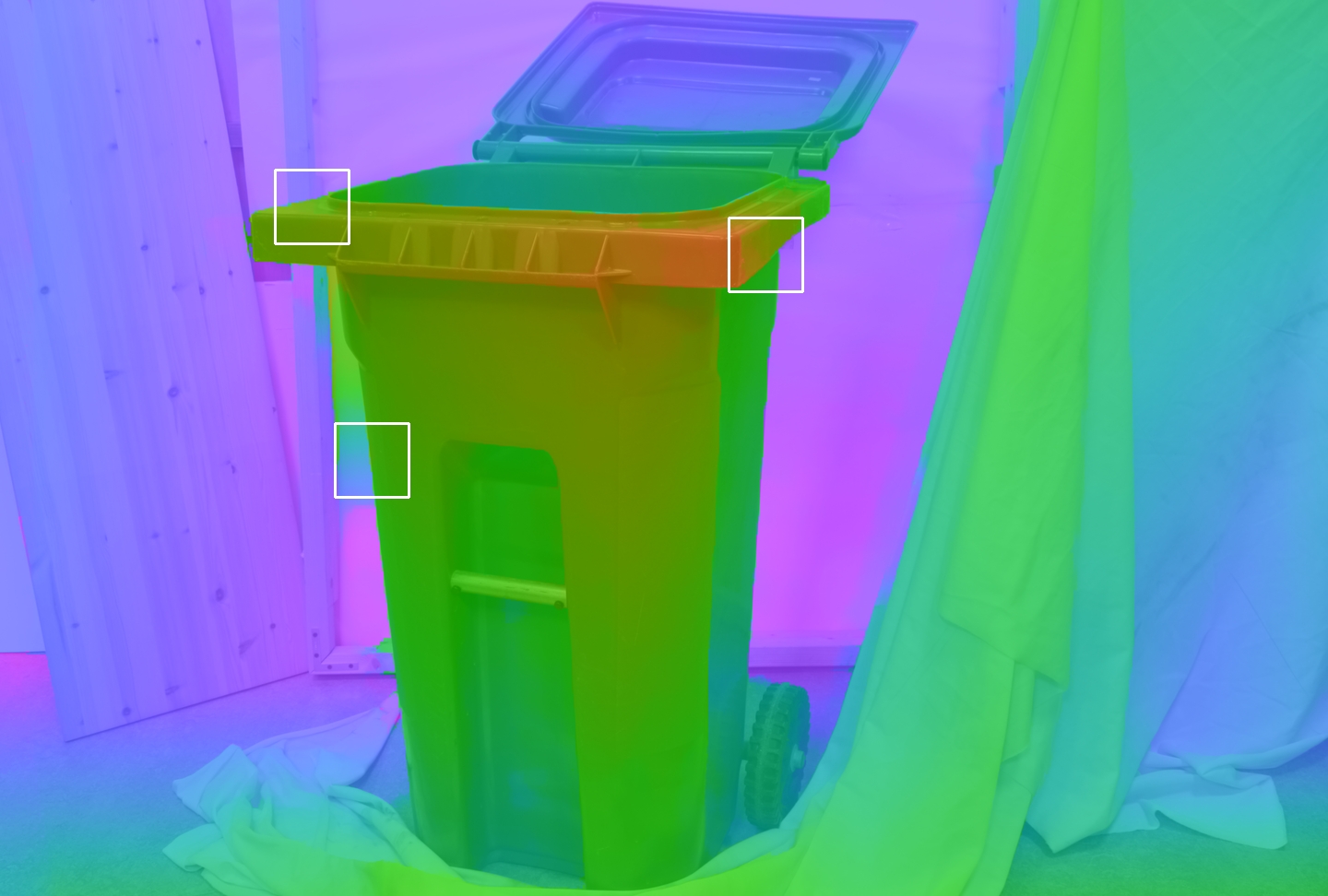}
    \includegraphics[width=0.41in]{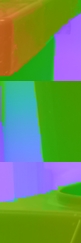}
    \caption{MC-CNN\cite{Zbontar2015} + RBS}
  \end{subfigure}
  \begin{subfigure}[!]{2.3in}
    \includegraphics[width=1.83in]{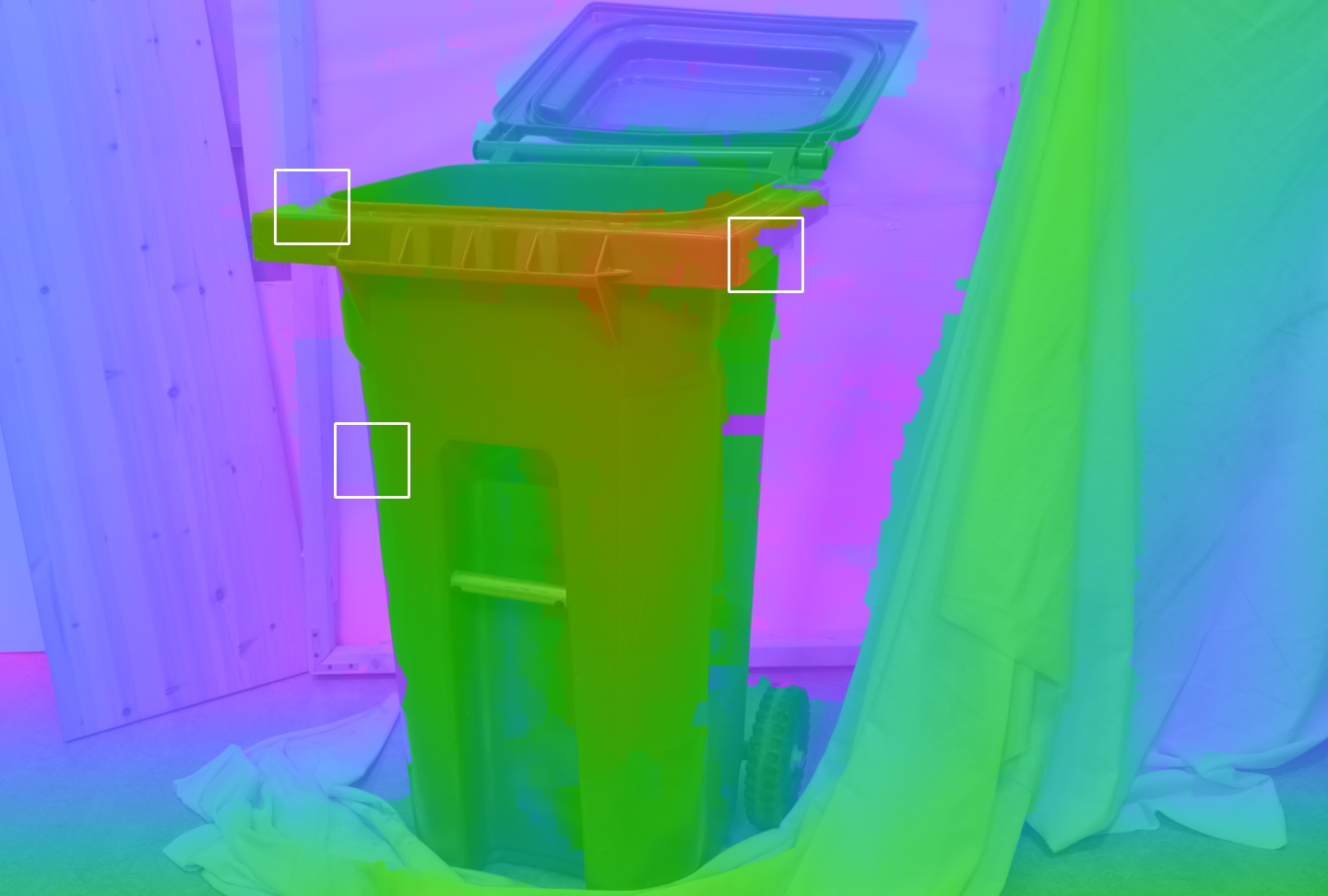}
    \includegraphics[width=0.41in]{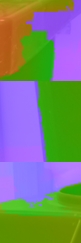}
    \caption{MeshStereo\cite{Zhang2015}}
  \end{subfigure}
  \begin{subfigure}[!]{2.3in}
    \includegraphics[width=1.83in]{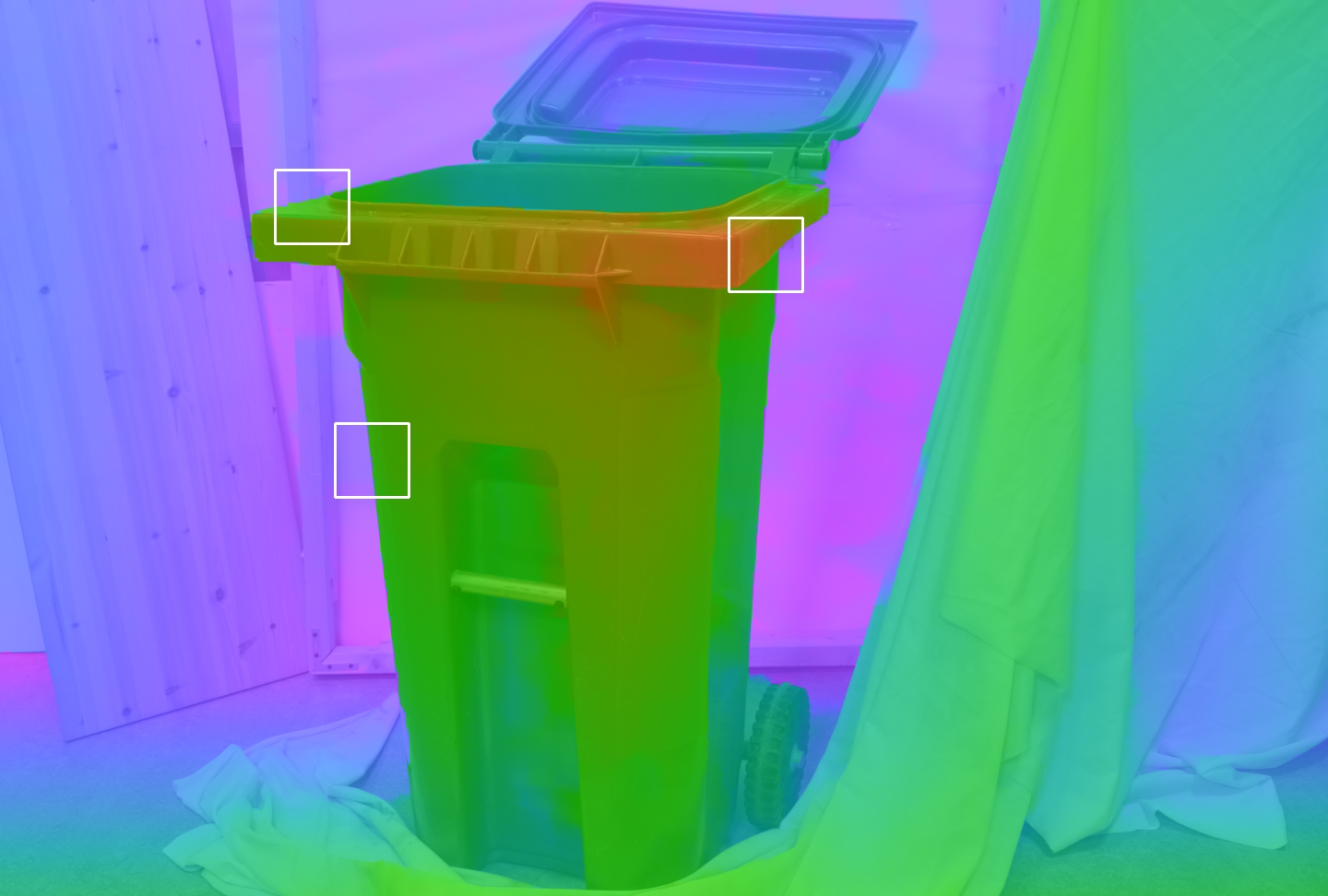}
    \includegraphics[width=0.41in]{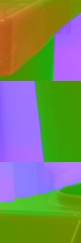}
    \caption{TMAP\cite{Zhang2015} + RBS}
  \end{subfigure}
  \begin{subfigure}[!]{2.3in}
    \includegraphics[width=1.83in]{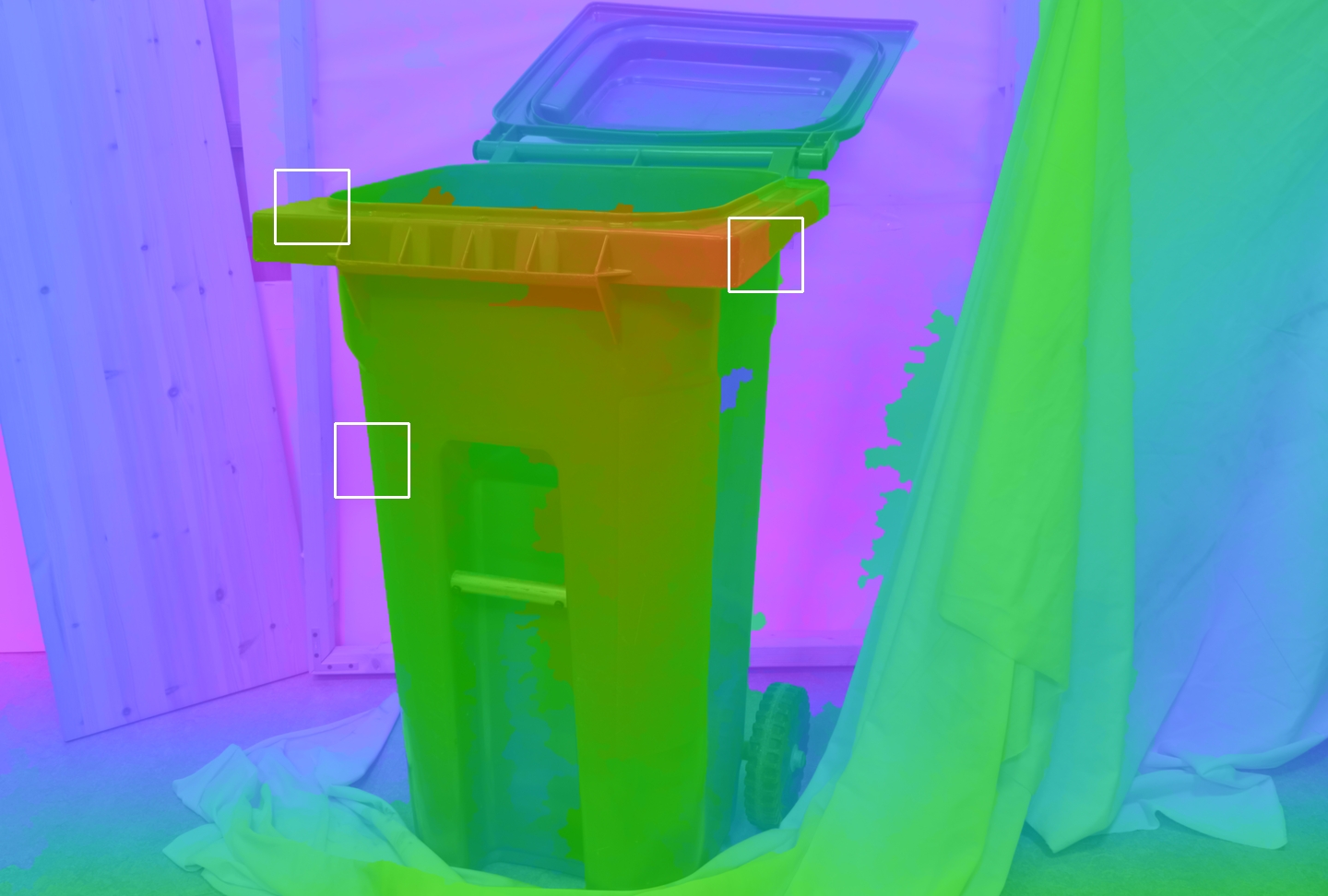}
    \includegraphics[width=0.41in]{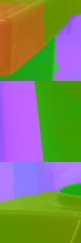}
    \caption{TMAP\cite{Psota2015}}
  \end{subfigure}
  \begin{subfigure}[!]{2.3in}
    \includegraphics[width=1.83in]{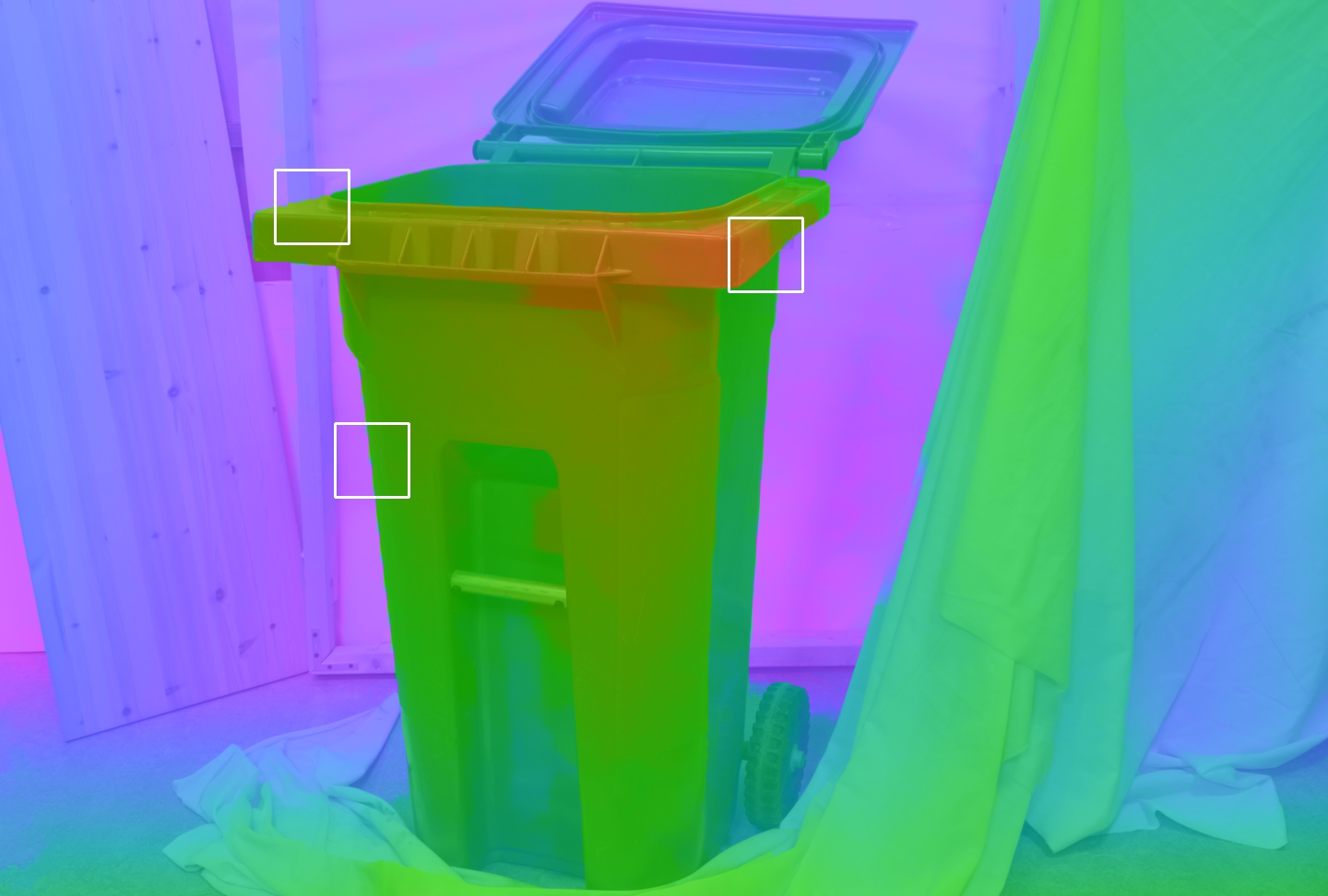}
    \includegraphics[width=0.41in]{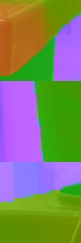}
    \caption{TMAP\cite{Psota2015} + RBS}
  \end{subfigure}
  \begin{subfigure}[!]{2.3in}
    \includegraphics[width=1.83in]{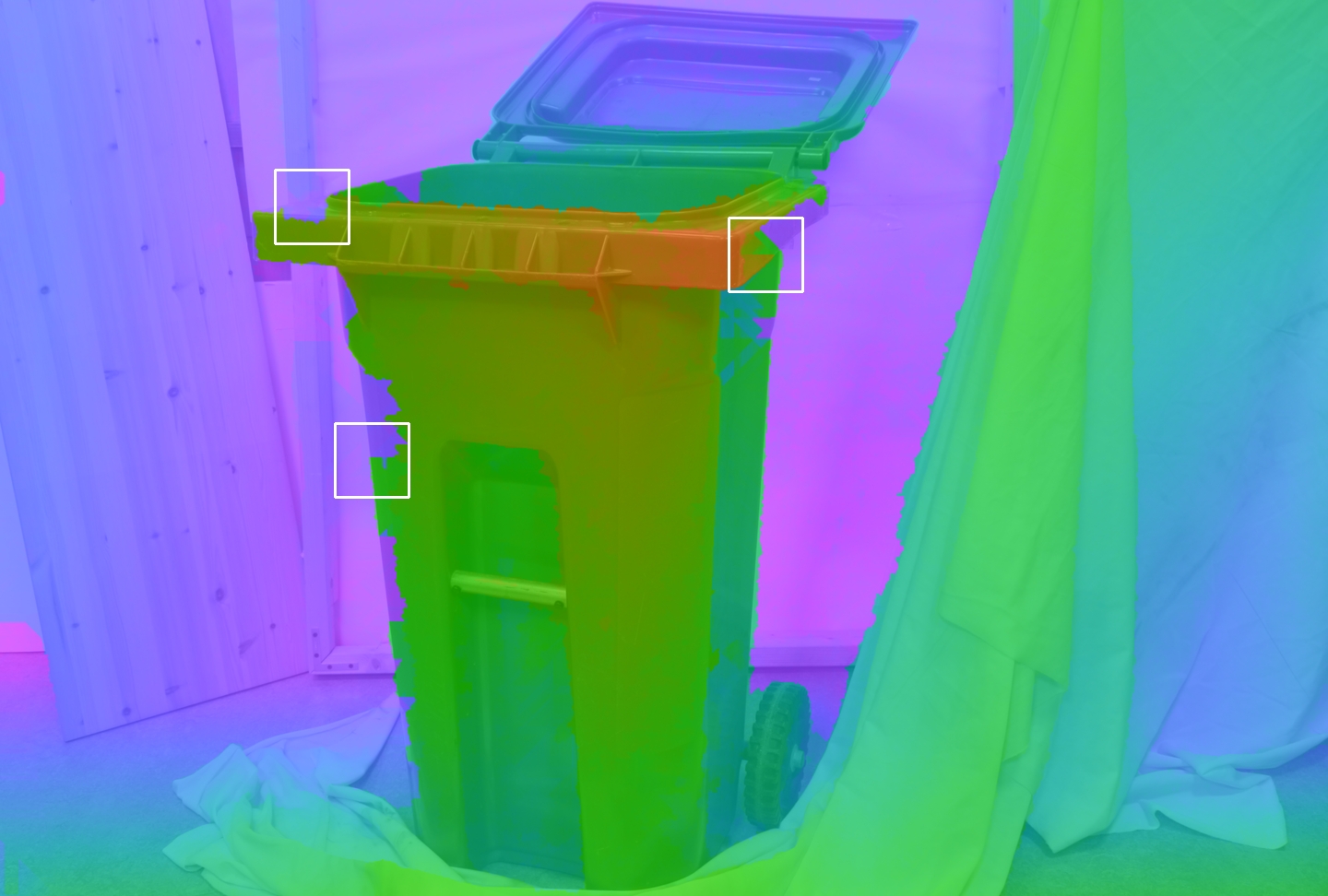}
    \includegraphics[width=0.41in]{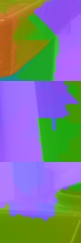}
    \caption{SGM\cite{Hirschmuller05accurateand}}
  \end{subfigure}
  \begin{subfigure}[!]{2.3in}
    \includegraphics[width=1.83in]{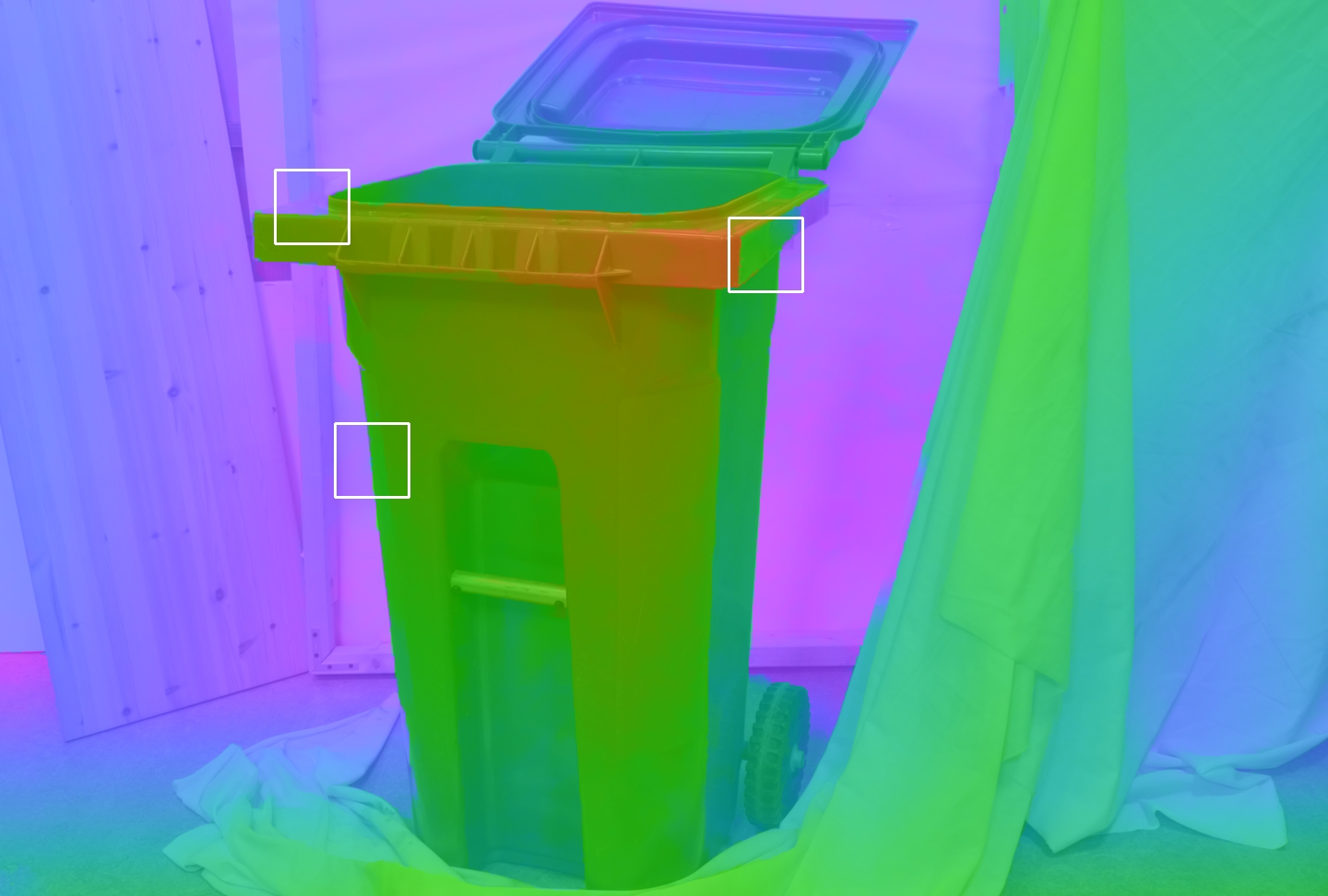}
    \includegraphics[width=0.41in]{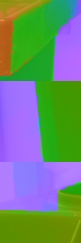}
    \caption{SGM\cite{Hirschmuller05accurateand} + RBS}
  \end{subfigure}
  \caption{
  Results on the training set of the Middlebury Stereo Dataset V3 \cite{Scharstein2014} where our robust bilateral solver is used to improve the depth map predicted by four top-performing stereo algorithms.
  On the left we have the depth map produced by each stereo algorithm (with zoomed in regions) which is used as the target in our solver.
  On the right we have the output of our solver, where we see that quality is significantly improved.
  \label{fig:middlebury1}
  }
\end{figure*}

\begin{figure*}[p]
\centering
  \begin{subfigure}[!]{2.3in}
    \includegraphics[width=1.83in]{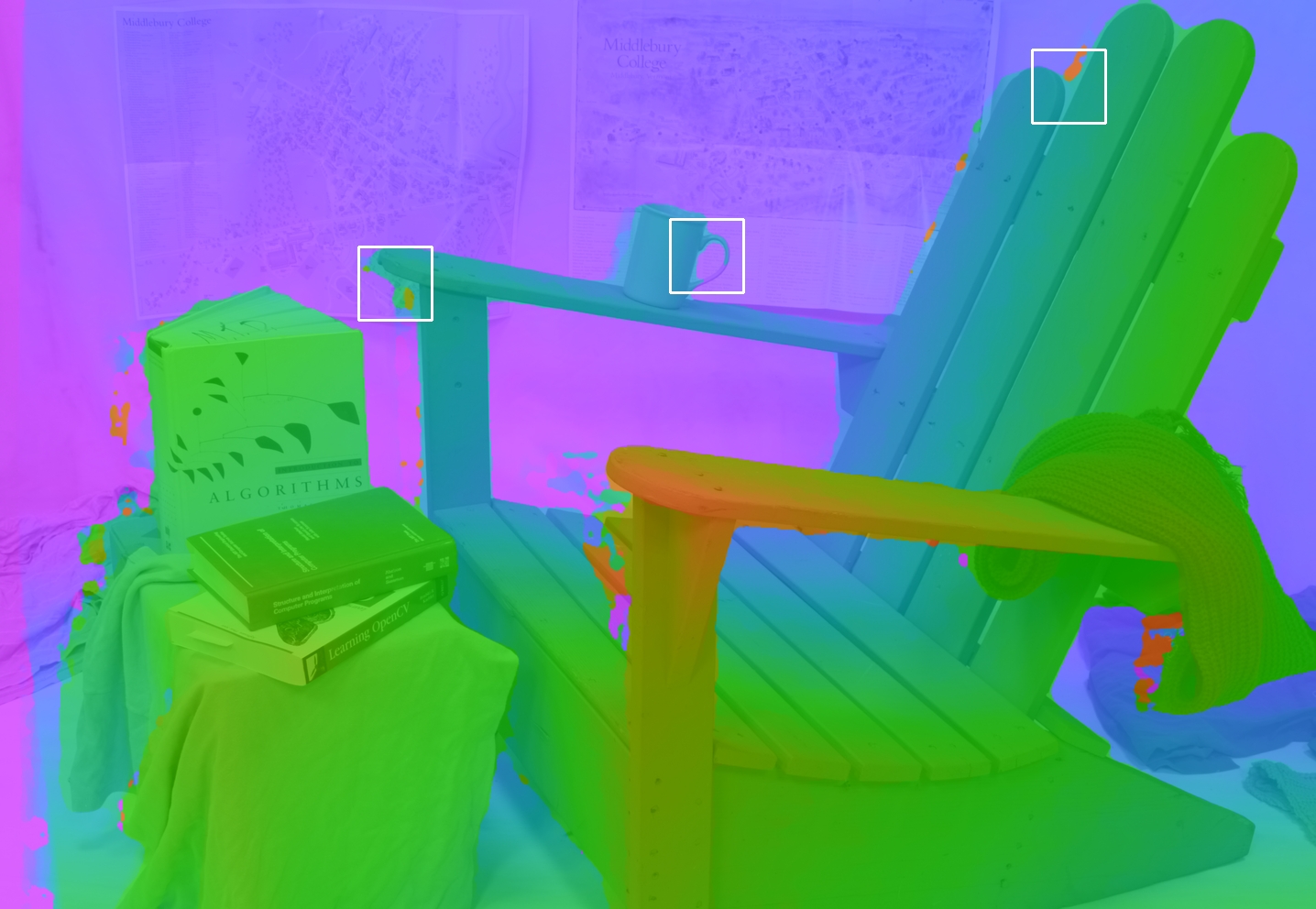}
    \includegraphics[width=0.42in]{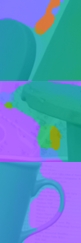}
    \caption{MC-CNN\cite{Zbontar2015}}
  \end{subfigure}
  \begin{subfigure}[!]{2.3in}
    \includegraphics[width=1.83in]{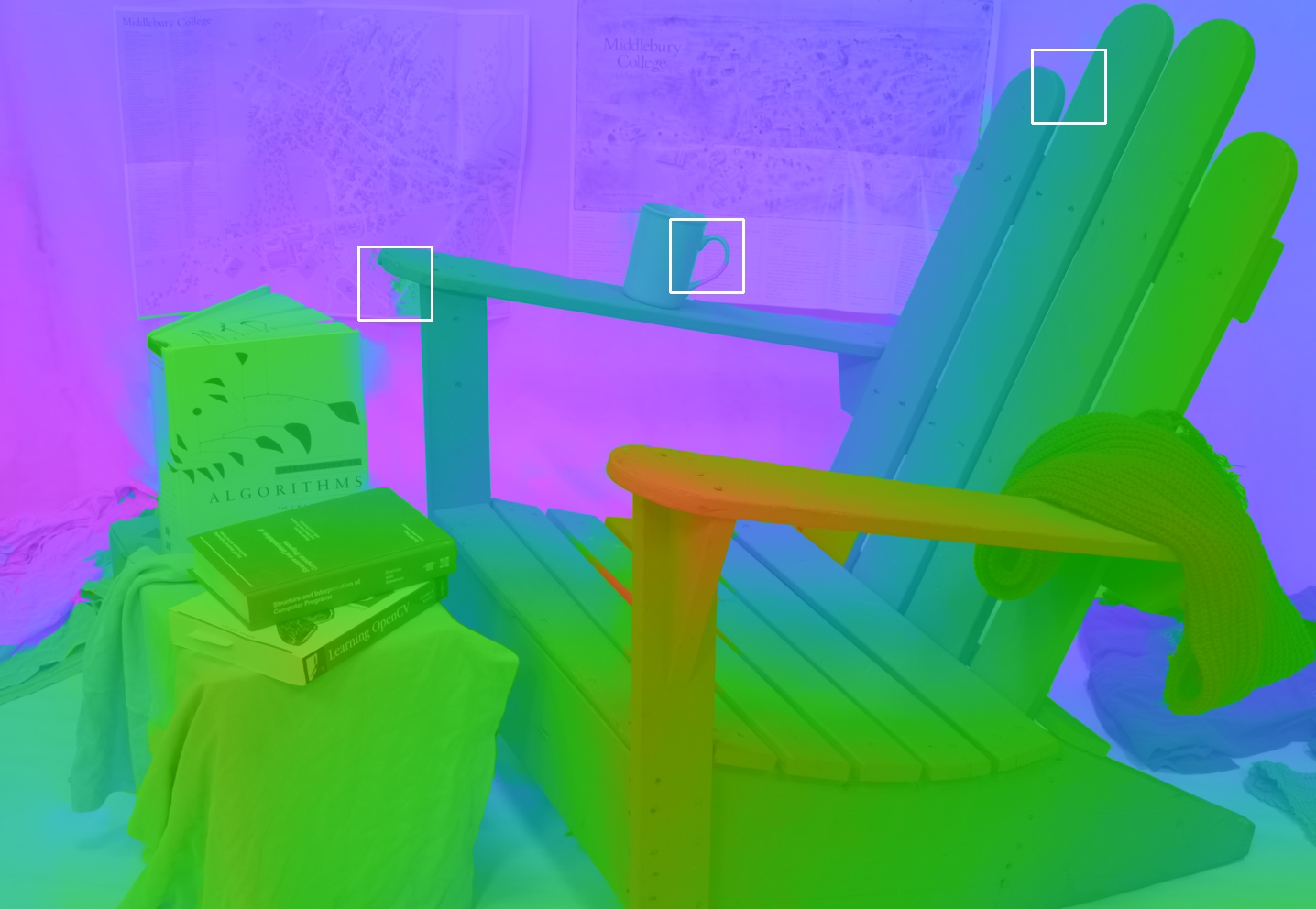}
    \includegraphics[width=0.42in]{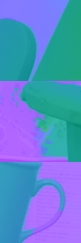}
    \caption{MC-CNN\cite{Zbontar2015} + RBS}
  \end{subfigure}
  \begin{subfigure}[!]{2.3in}
    \includegraphics[width=1.83in]{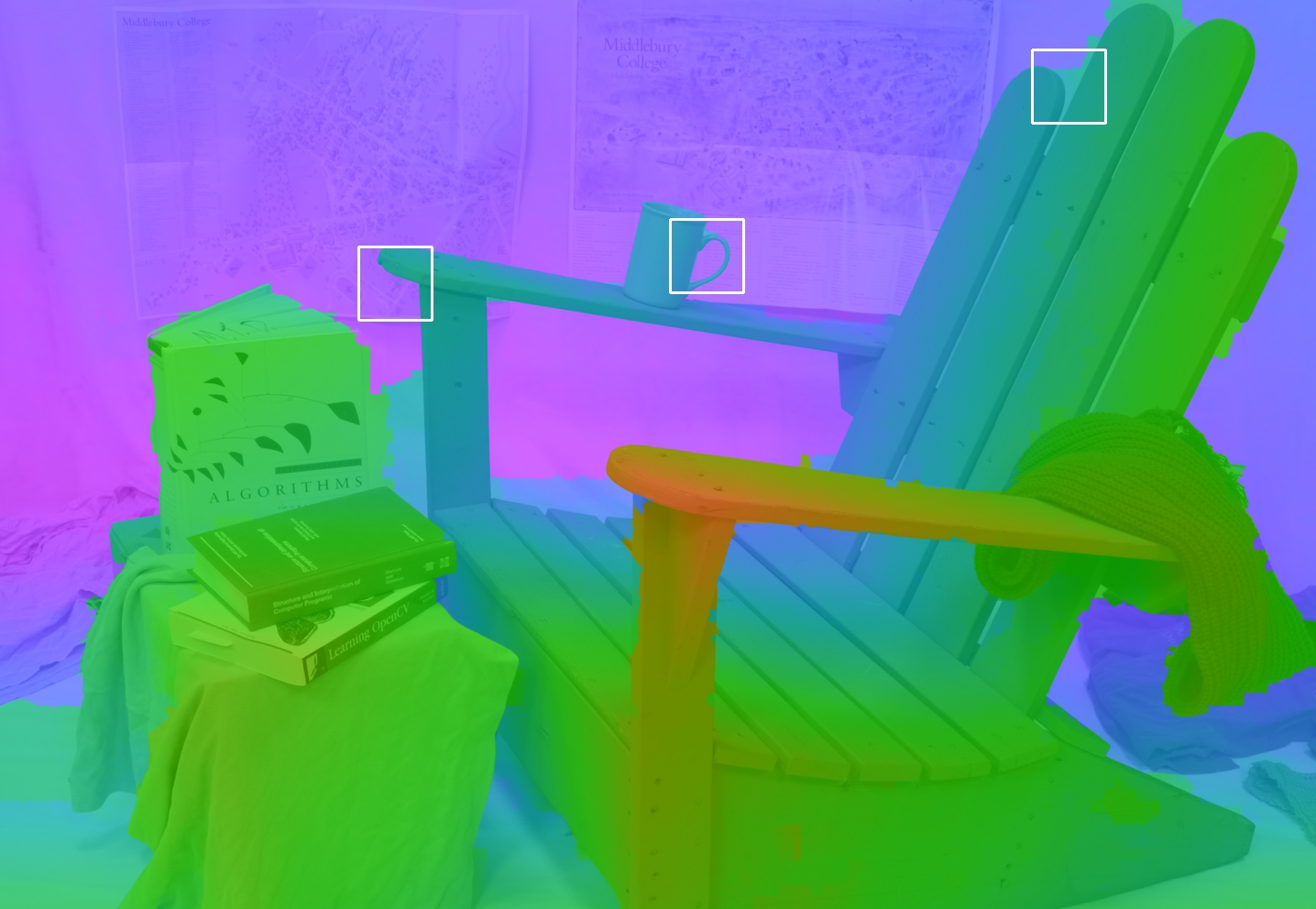}
    \includegraphics[width=0.42in]{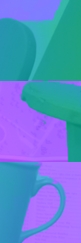}
    \caption{MeshStereo\cite{Zhang2015}}
  \end{subfigure}
  \begin{subfigure}[!]{2.3in}
    \includegraphics[width=1.83in]{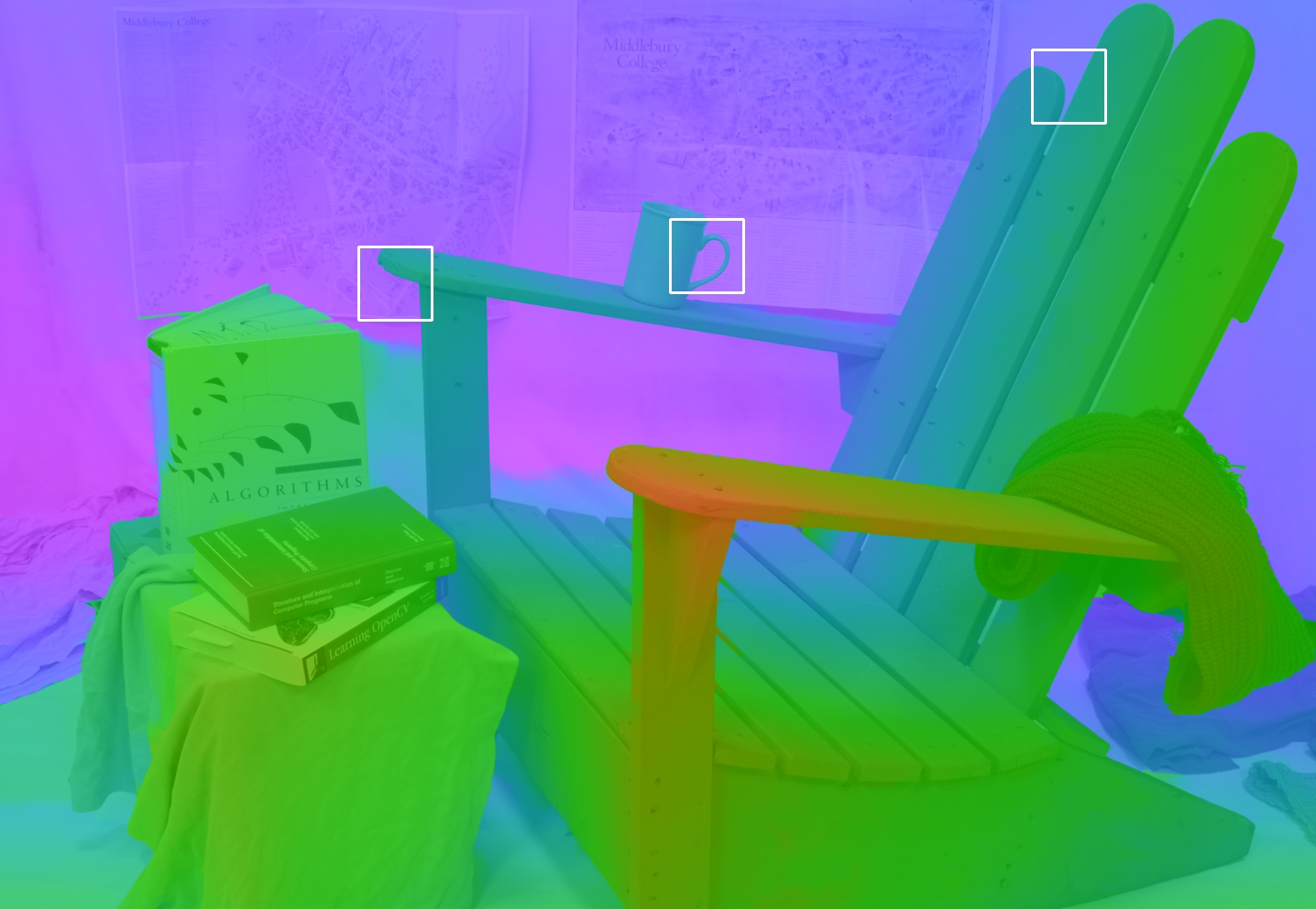}
    \includegraphics[width=0.42in]{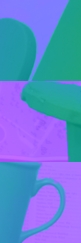}
    \caption{TMAP\cite{Zhang2015} + RBS}
  \end{subfigure}
  \begin{subfigure}[!]{2.3in}
    \includegraphics[width=1.83in]{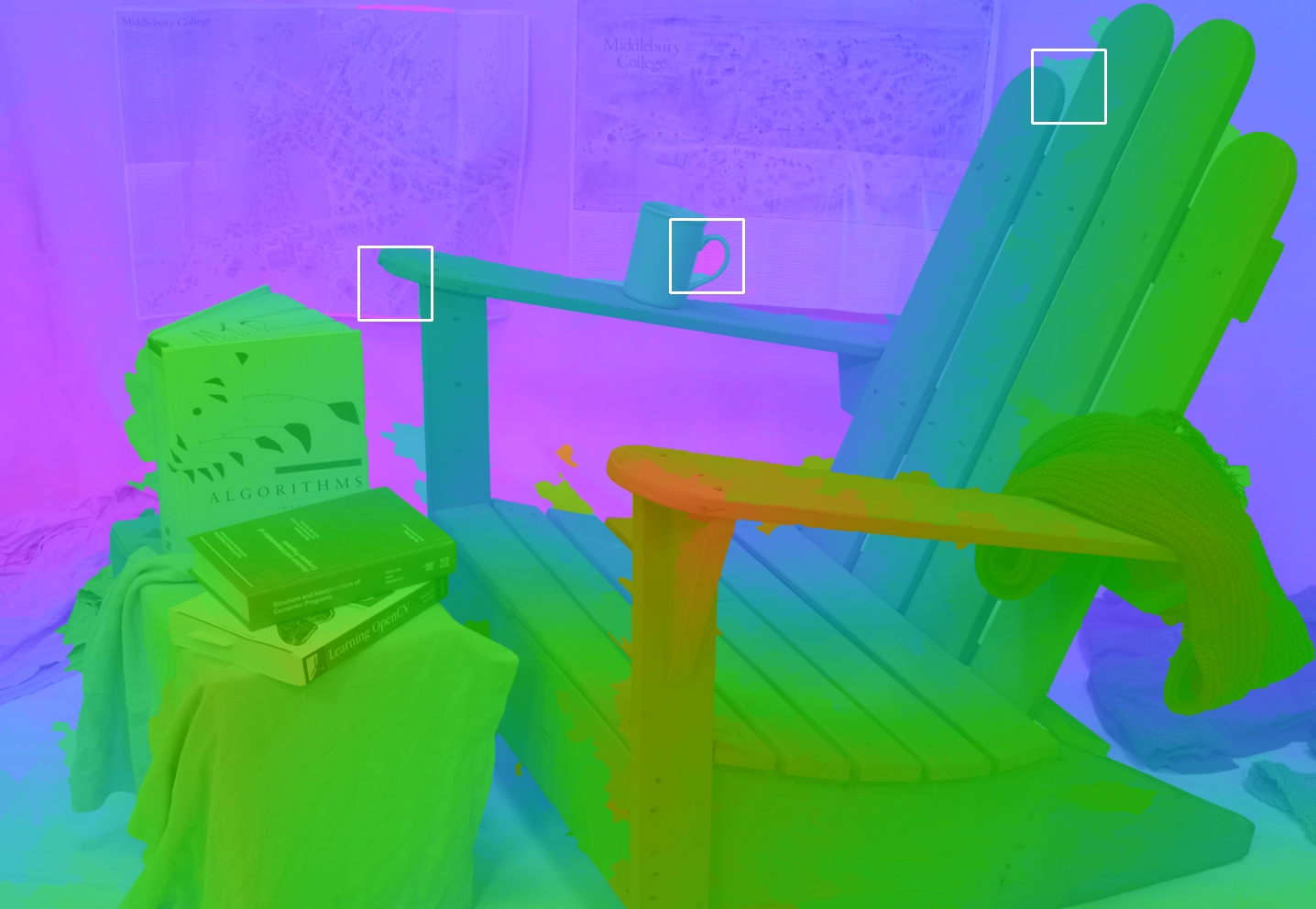}
    \includegraphics[width=0.42in]{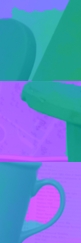}
    \caption{TMAP\cite{Psota2015}}
  \end{subfigure}
  \begin{subfigure}[!]{2.3in}
    \includegraphics[width=1.83in]{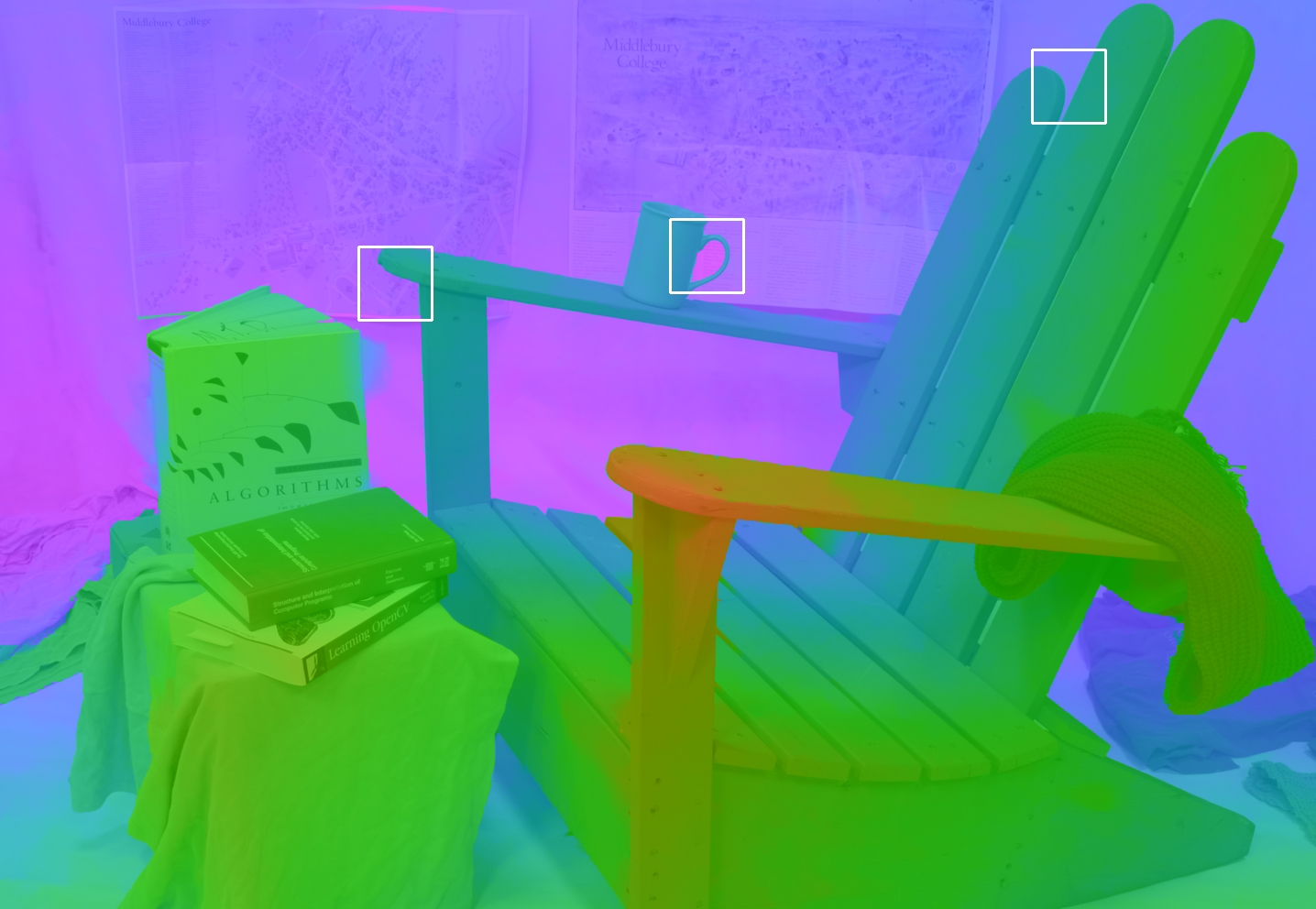}
    \includegraphics[width=0.42in]{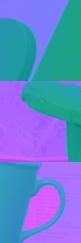}
    \caption{TMAP\cite{Psota2015} + RBS}
  \end{subfigure}
  \begin{subfigure}[!]{2.3in}
    \includegraphics[width=1.83in]{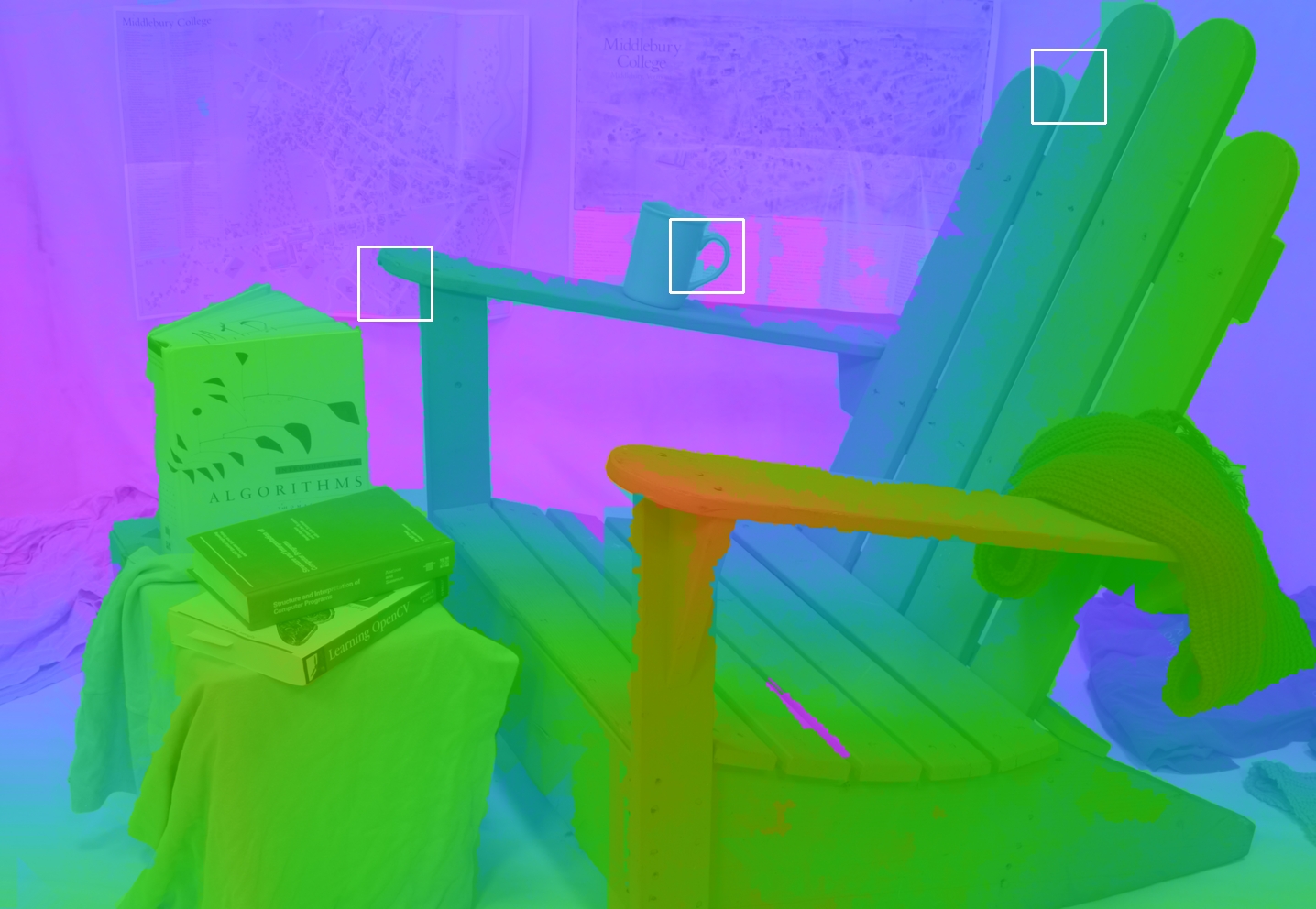}
    \includegraphics[width=0.42in]{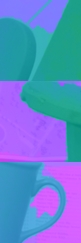}
    \caption{SGM\cite{Hirschmuller05accurateand}}
  \end{subfigure}
  \begin{subfigure}[!]{2.3in}
    \includegraphics[width=1.83in]{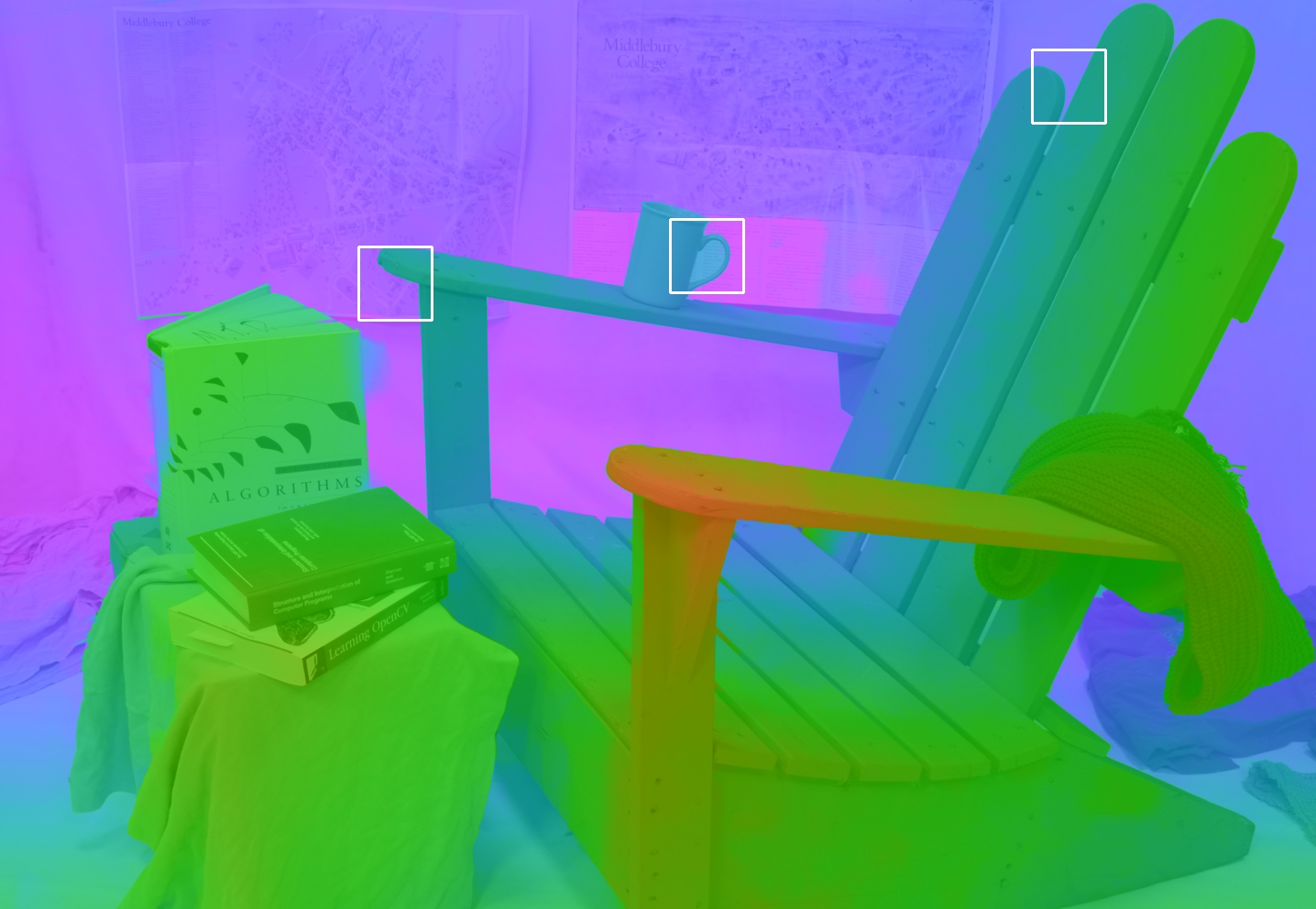}
    \includegraphics[width=0.42in]{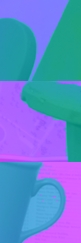}
    \caption{SGM\cite{Hirschmuller05accurateand} + RBS}
  \end{subfigure}
  \caption{More results on the training set of the Middlebury Stereo Dataset V3 \cite{Scharstein2014} in the same format as Figure~\ref{fig:middlebury1}.
  \label{fig:middlebury2}
  }
\end{figure*}

\begin{figure*}[p]
\centering
  \begin{subfigure}[!]{2.3in}
    \includegraphics[width=1.83in]{trainset_figures/12/MC-CNN_input_overlay_boxes.jpg}
    \includegraphics[width=0.41in]{trainset_figures/12/MC-CNN_input_tiles.jpg}
    \caption{MC-CNN\cite{Zbontar2015}}
  \end{subfigure}
  \begin{subfigure}[!]{2.3in}
    \includegraphics[width=1.83in]{trainset_figures/12/MC-CNN_output_overlay_boxes.jpg}
    \includegraphics[width=0.41in]{trainset_figures/12/MC-CNN_output_tiles.jpg}
    \caption{MC-CNN\cite{Zbontar2015} + RBS (Ours)}
  \end{subfigure}
  \begin{subfigure}[!]{2.3in}
    \includegraphics[width=1.83in]{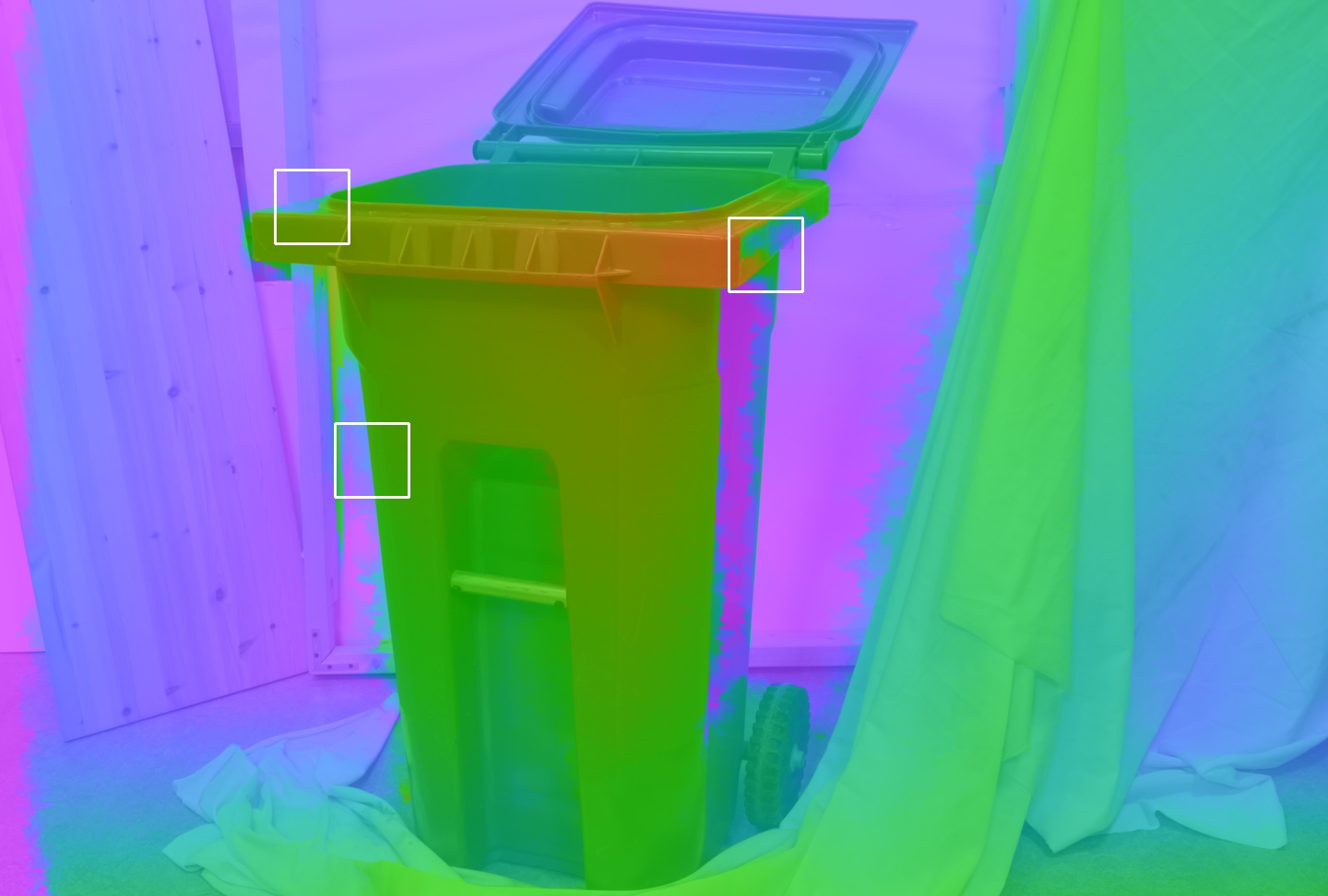}
    \includegraphics[width=0.41in]{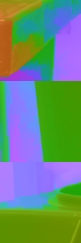}
    \caption{MC-CNN\cite{Zbontar2015} + TF \cite{Yang2015}}
  \end{subfigure}
  \begin{subfigure}[!]{2.3in}
    \includegraphics[width=1.83in]{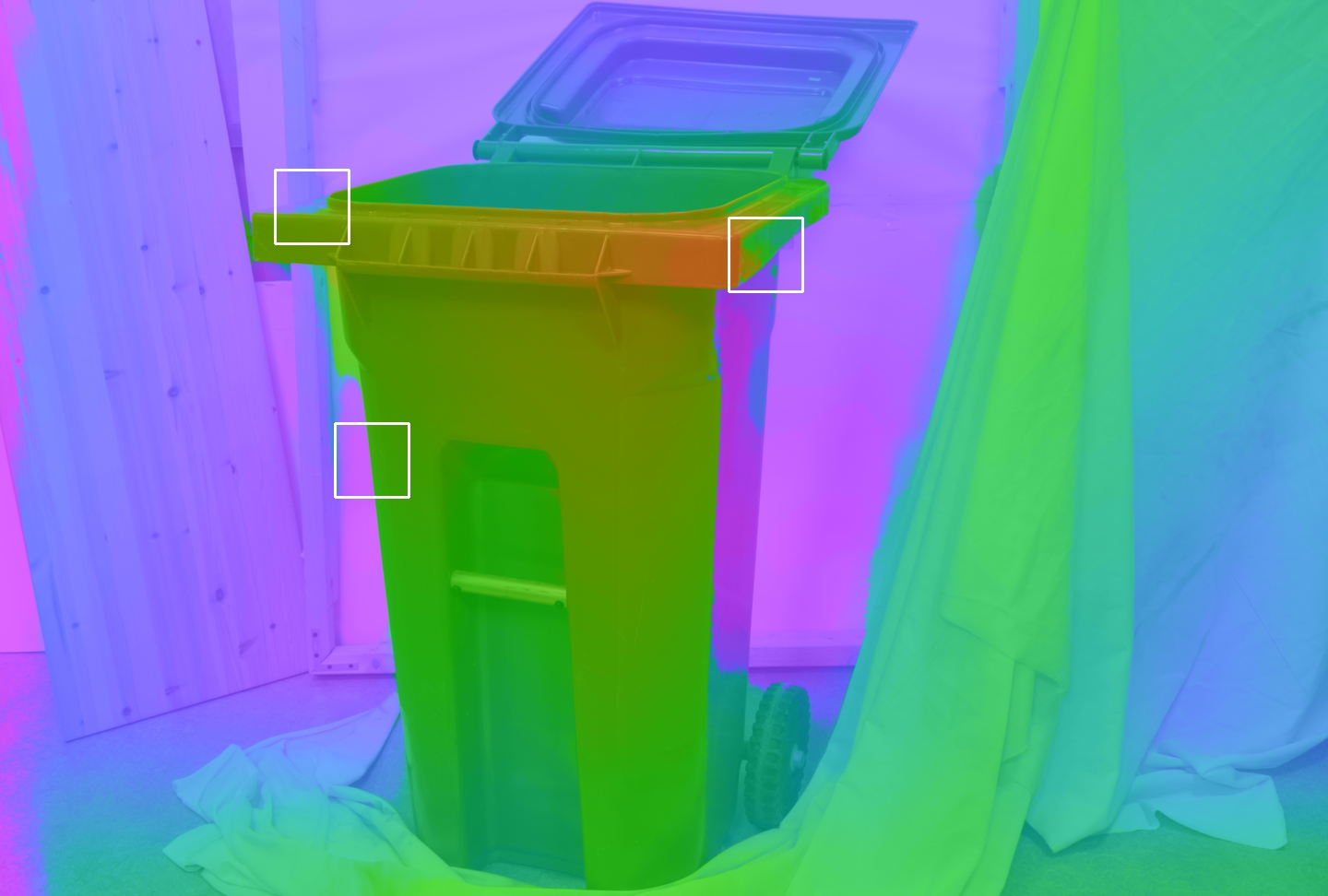}
    \includegraphics[width=0.41in]{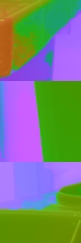}
    \caption{MC-CNN\cite{Zbontar2015} + WMF \cite{Ma2013}}
  \end{subfigure}
  \begin{subfigure}[!]{2.3in}
    \includegraphics[width=1.83in]{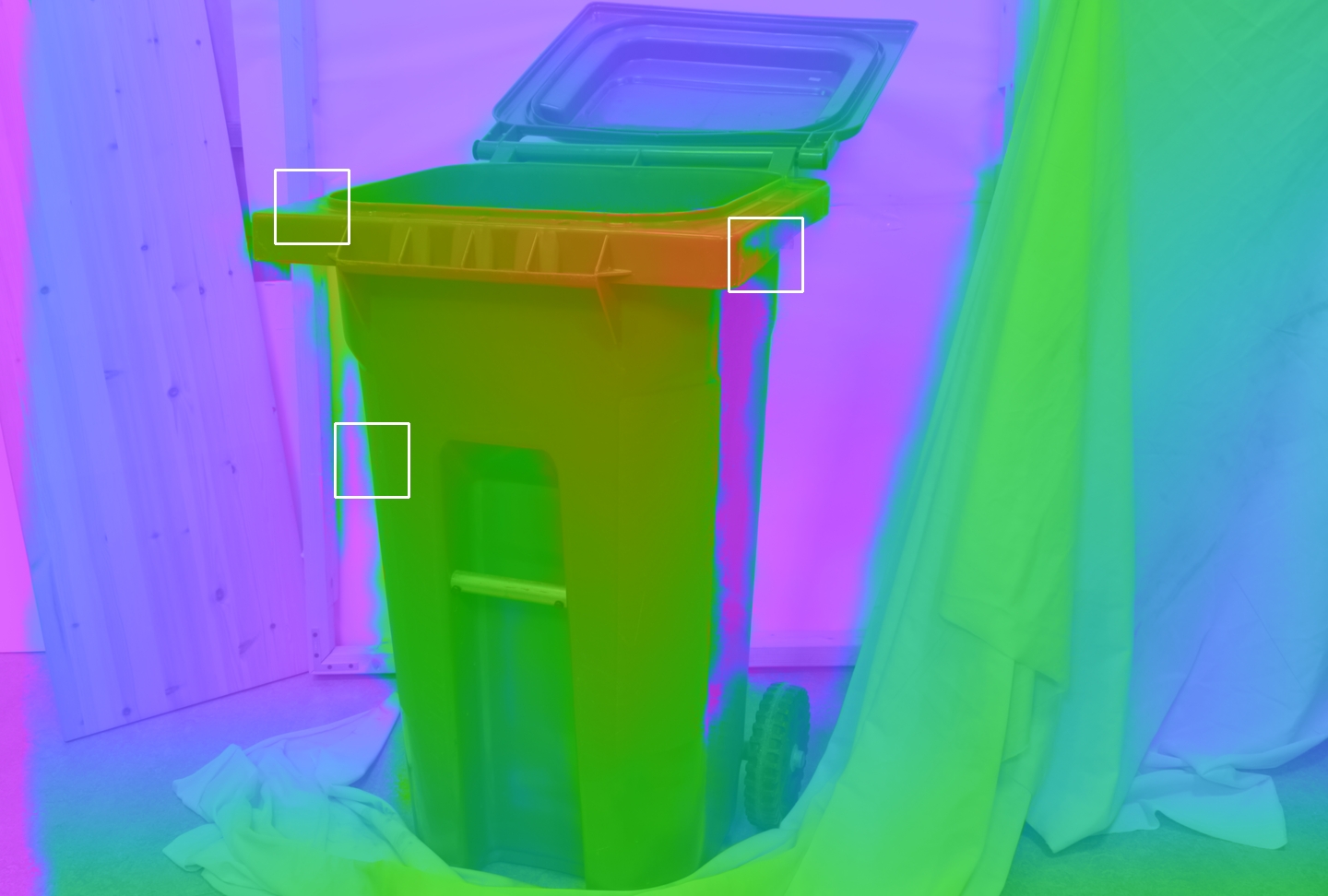}
    \includegraphics[width=0.41in]{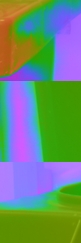}
    \caption{MC-CNN\cite{Zbontar2015} + FGF \cite{He2015}}
  \end{subfigure}
  \begin{subfigure}[!]{2.3in}
    \includegraphics[width=1.83in]{trainset_figures/12/MC-CNN_fgf_overlay_boxes.jpg}
    \includegraphics[width=0.41in]{trainset_figures/12/MC-CNN_fgf_tiles.jpg}
    \caption{MC-CNN\cite{Zbontar2015} + DT \cite{GastalOliveira2011DomainTransform}}
  \end{subfigure}
  \caption{
  Results on the training set of the Middlebury Stereo Dataset V3 \cite{Scharstein2014}, in which we compare our robust bilateral solver against several baseline techniques for post-processing depth maps.
  Our model's output exhibits much higher quality than the input or any baseline, especially at the discontinuities shown in the cropped regions.
  \label{fig:middleburyAlgo1}
  }
\end{figure*}

\begin{figure*}[p]
\centering
  \begin{subfigure}[!]{2.3in}
    \includegraphics[width=1.83in]{trainset_figures/01/MC-CNN_input_overlay_boxes.jpg}
    \includegraphics[width=0.42in]{trainset_figures/01/MC-CNN_input_tiles.jpg}
    \caption{MC-CNN\cite{Zbontar2015}}
  \end{subfigure}
  \begin{subfigure}[!]{2.3in}
    \includegraphics[width=1.83in]{trainset_figures/01/MC-CNN_output_overlay_boxes.jpg}
    \includegraphics[width=0.42in]{trainset_figures/01/MC-CNN_output_tiles.jpg}
    \caption{MC-CNN\cite{Zbontar2015} + RBS (Ours)}
  \end{subfigure}
  \begin{subfigure}[!]{2.3in}
    \includegraphics[width=1.83in]{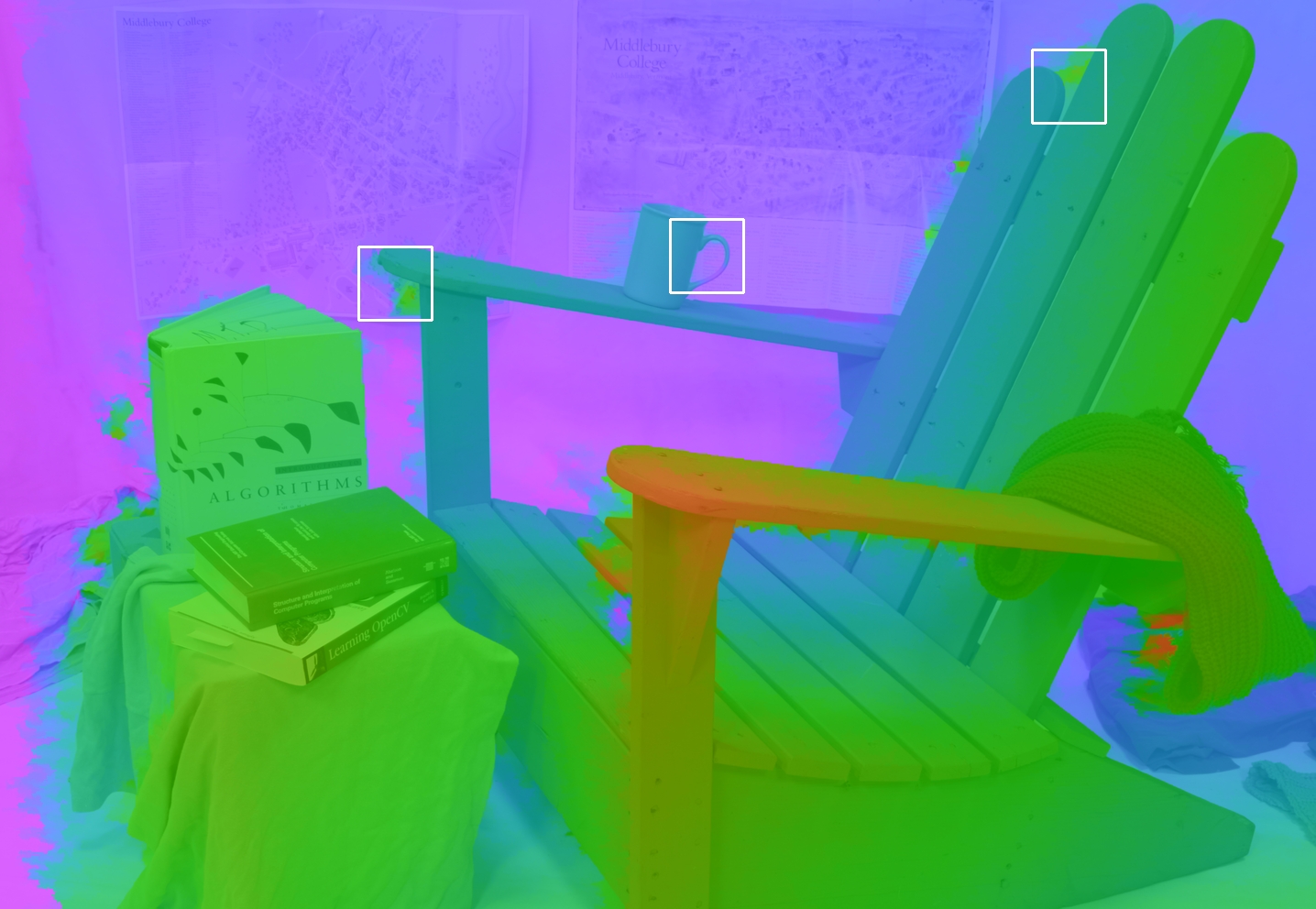}
    \includegraphics[width=0.42in]{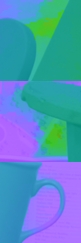}
    \caption{MC-CNN\cite{Zbontar2015} + TF \cite{Yang2015}}
  \end{subfigure}
  \begin{subfigure}[!]{2.3in}
    \includegraphics[width=1.83in]{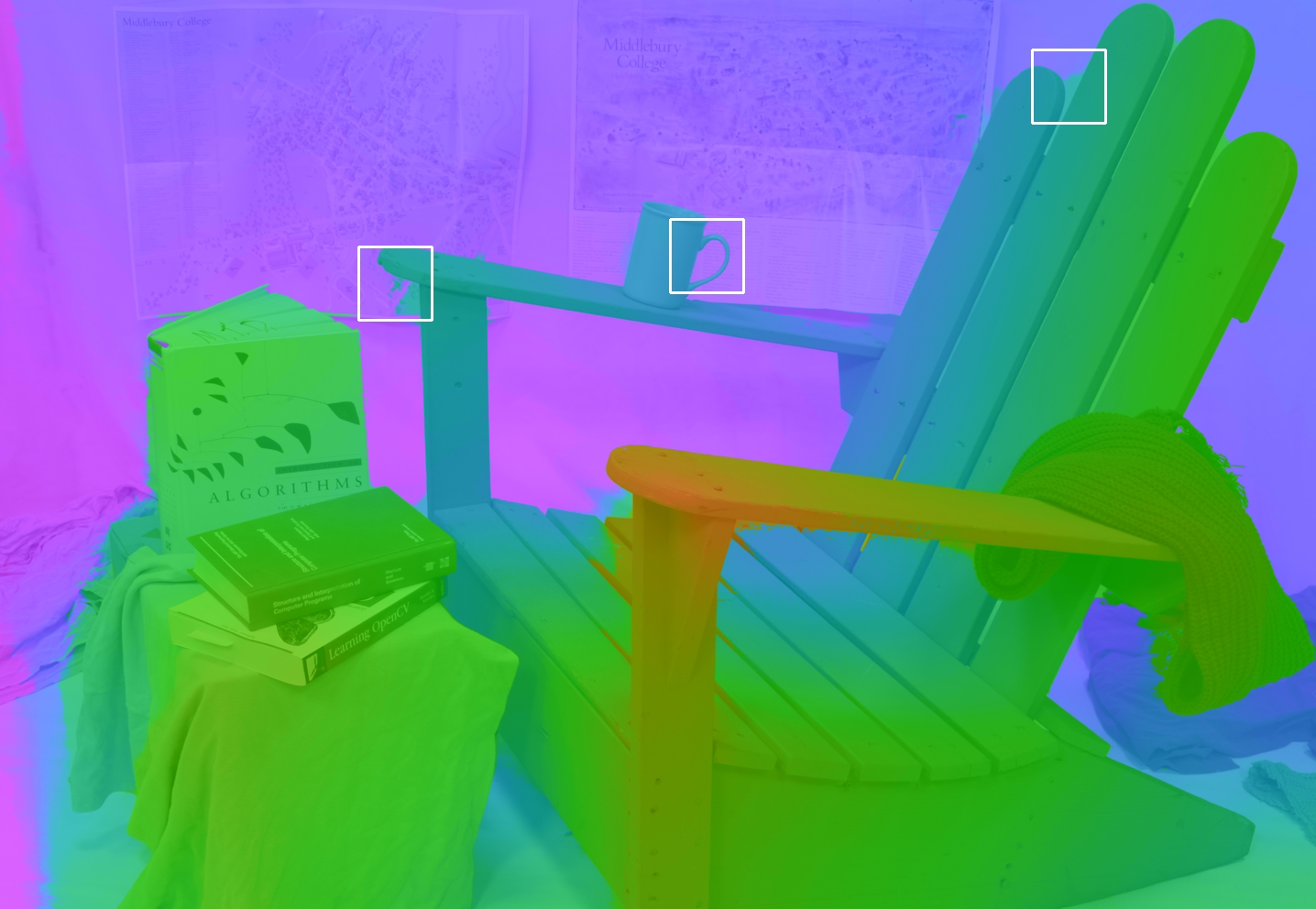}
    \includegraphics[width=0.42in]{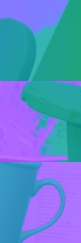}
    \caption{MC-CNN\cite{Zbontar2015} + WMF \cite{Ma2013}}
  \end{subfigure}
  \begin{subfigure}[!]{2.3in}
    \includegraphics[width=1.83in]{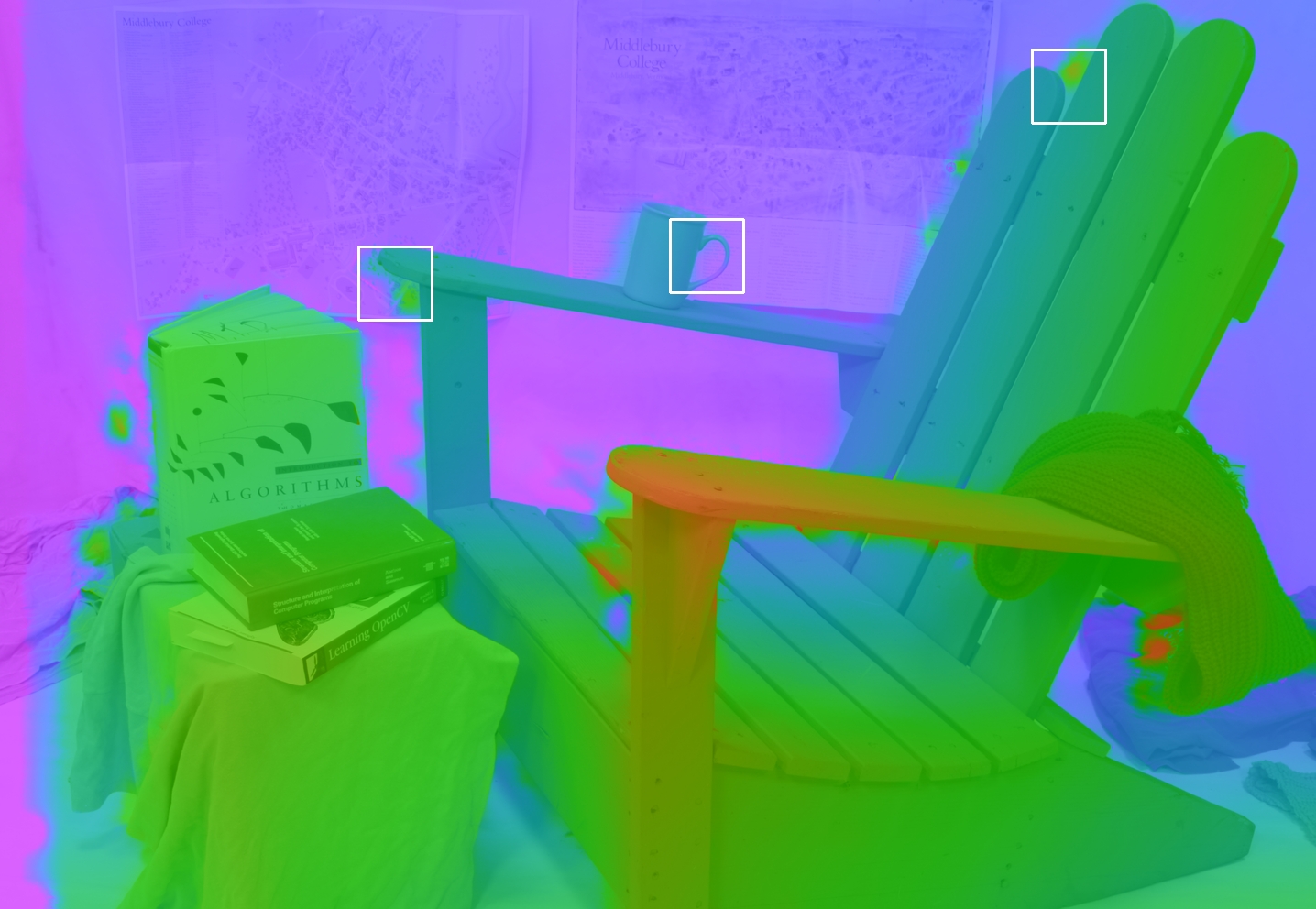}
    \includegraphics[width=0.42in]{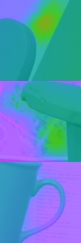}
    \caption{MC-CNN\cite{Zbontar2015} + FGF \cite{He2015}}
  \end{subfigure}
  \begin{subfigure}[!]{2.3in}
    \includegraphics[width=1.83in]{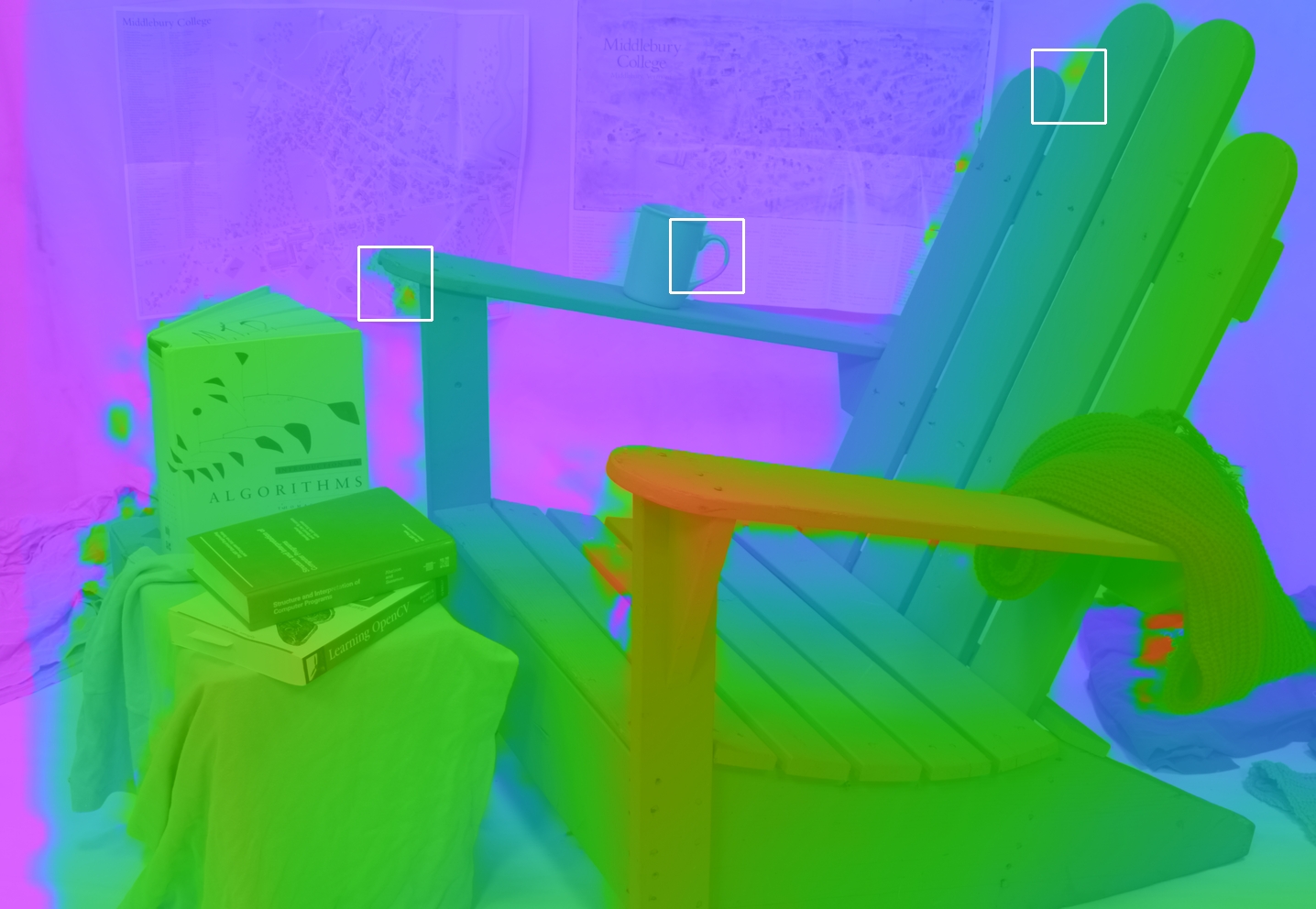}
    \includegraphics[width=0.42in]{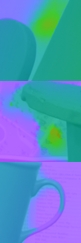}
    \caption{MC-CNN\cite{Zbontar2015} + DT \cite{GastalOliveira2011DomainTransform}}
  \end{subfigure}
  \caption{More results on the training set of the Middlebury Stereo Dataset V3 \cite{Scharstein2014} in the same format as Figure~\ref{fig:middleburyAlgo1}.
  \label{fig:middleburyAlgo2}
  }
\end{figure*}

In the paper we demonstrated the value of our robust bilateral solver as a post-processing procedure for stereo algorithms, using the Middlebury Stereo Benchmark V3 \cite{Scharstein2014}.
See Figures~\ref{fig:testset_middlebury1}-\ref{fig:testset_middlebury5} for examples of test set images from the Middlebury dataset in which we post-process the output of the state-of-the-art MC-CNN stereo technique\cite{Zbontar2015}, which we obtained using the publicly available source accompanying the technique.
See Figures~\ref{fig:middlebury1} and \ref{fig:middlebury2} for additional results on training set images from the Middlebury dataset, where we evaluate against a wider selection of stereo algorithms.
See Figures~\ref{fig:middleburyAlgo1} and \ref{fig:middleburyAlgo2} for results in which we have post-processed the training-set output of those these top-performing stereo algorithms with baseline edge-aware filtering or depth-map enhancement techniques.
We present these results on the training set because the Middlebury benchmark makes the training-set output of other stereo algorithms freely available, while obtaining test-set results requires the code or cooperation of the authors of each technique, or re-implementing these techniques.
The depth maps corresponding to each baseline stereo technique were therefore downloaded from the Middlebury website.
The output corresponding to each post-processing technique we compare against were produced by us, while taking care to tune all parameters to maximize performance on the Middlebury training set (the geometric mean of all 6 error metrics) for the MC-CNN output (these parameters are then used for all other stereo techniques).
For complete transparency, we will detail the procedure used to produce each baseline results:
\begin{description}[align=left]
\item [TF] This is the tree-filtering technique of Yang \cite{Yang2015}, which we ran ourselves using the publicly available code\footnote{\url{http://www.cs.cityu.edu.hk/~qiyang/publications/software/tree_filter.zip}}. The parameter settings used in these experiments are $\sigma_r = 0.25$ with nonlocal filtering, which experimentation showed to be the optimal parameter settings for this task.
\item [FGF] This is our own Matlab implementation of the (color) fast guided filter \cite{He2015}, which we found to be faster than the implementation built into Matlab 2015, and significantly faster than the released code\footnote{\url{http://research.microsoft.com/en-us/um/people/kahe/eccv10/fast-guided-filter-code-v1.rar}} while producing identical results. The parameters were tuned for this task, with a box filter size of $8$, $\epsilon = 0.01^2$, and a subsampling factor of $4$.
\item [WMF] This is the weighted median filter approach of Ma \etal \cite{Ma2013}, which we ran using the publicly available code\footnote{\url{http://research.microsoft.com/en-us/um/people/kahe/iccv13wmf/matlab_wmf_release_v1.rar}}. Because this technique was designed for a similar use case to its use here, we used the same parameter settings as were use in the paper: $r = \mathrm{max}(\mathit{width}, \mathit{height})$, $\epsilon=0.01^2$, followed by a median filter.
\item [DT] This is the recursive formulation of the domain transform \cite{GastalOliveira2011DomainTransform} with optimally-tuned parameters for this task ($\sigma_r = 64$, $\sigma_s = 32$).
\end{description}

\begin{table}[b!]
\begin{center}
\caption{
Our RBS improves depth map quality for a variety of state-of-the-art stereo algorithms on the Middlebury Stereo Dataset V3 \cite{Scharstein2014}.
Test-set numbers were taken from the Middlebury website, while training-set numbers were produced by our own evaluation code.
\label{table:middlebury}
}
\begin{tabular}{ l | c c c | c c c }
$\mathrm{Method}$ & \multicolumn{3}{c|}{$\mathrm{All}$} & \multicolumn{3}{c}{$\mathrm{NoOcc}$}  \\
 & $\mathrm{bad}\,1\%$ & $\mathrm{MAE}$ & $\mathrm{RMSE}$ & $\mathrm{bad}\,1\%$ & $\mathrm{MAE}$ & $\mathrm{RMSE}$  \\
\hline
\multicolumn{7}{l}{\quad Test Set} \\
\hline
$\mathrm{                   MC-CNN}$\cite{Zbontar2015} & $ 28.1 $ \cellcolor{Yellow}  & $ 17.9 $                     & $ 55.0 $                     & $ 18.0 $ \cellcolor{Yellow} & $ 3.82 $                     & $ 21.3 $                    \\
$\mathrm{             MC-CNN}$\cite{Zbontar2015}$\mathrm{ + RBS}$ & $ 28.2 $                     & $ 8.19 $ \cellcolor{Yellow}  & $ 29.9 $ \cellcolor{Yellow}  & $ 18.9 $                    & $ 2.67 $ \cellcolor{Yellow}  & $ 15.0 $ \cellcolor{Yellow} \\
\multicolumn{7}{c}{} \\
\hline
\multicolumn{7}{l}{\quad Training Set} \\
\hline
    $\mathrm{MC}$-$\mathrm{CNN}$\cite{Zbontar2015}                        & $ 20.07 $  & $ 5.93 $  & $ 18.36 $  & $ 10.42 $ \cellcolor{Yellow}  & $ 1.94 $  & $ 9.07 $  \\
    $\mathrm{MC}$-$\mathrm{CNN}$\cite{Zbontar2015} + $\mathrm{TF}$\cite{Yang2015} & $ 29.15 $  & $ 5.67 $  & $ 16.18 $  & $ 20.15 $  & $ 2.17 $  & $ 7.71 $  \\
    $\mathrm{MC}$-$\mathrm{CNN}$\cite{Zbontar2015} + $\mathrm{FGF}$\cite{He2015} & $ 32.29 $  & $ 5.91 $  & $ 16.32 $  & $ 23.62 $  & $ 2.42 $  & $ 7.98 $  \\
    $\mathrm{MC}$-$\mathrm{CNN}$\cite{Zbontar2015} + $\mathrm{WMF}$\cite{Ma2013} & $ 33.37 $  & $ 5.30 $  & $ 15.62 $  & $ 26.29 $  & $ 2.32 $  & $ 8.22 $  \\
    $\mathrm{MC}$-$\mathrm{CNN}$\cite{Zbontar2015} + $\mathrm{DT}$\cite{GastalOliveira2011DomainTransform} & $ 25.17 $  & $ 5.69 $  & $ 16.53 $  & $ 15.53 $  & $ 2.01 $  & $ 7.72 $  \\
    $\mathrm{MC}$-$\mathrm{CNN}$\cite{Zbontar2015} + $\mathrm{RBS\,(Ours)}$ & $ 19.49 $ \cellcolor{Yellow}  & $ 2.81 $ \cellcolor{Yellow}  & $ 8.44 $ \cellcolor{Yellow}  & $ 11.33 $  & $ 1.40 $ \cellcolor{Yellow}  & $ 5.23 $ \cellcolor{Yellow}  \\
\hline
             $\mathrm{MeshStereo}$\cite{Zhang2015}                        & $ 21.25 $ \cellcolor{Yellow}  & $ 3.83 $  & $ 10.75 $  & $ 15.13 $ \cellcolor{Yellow}  & $ 2.21 $  & $ 7.86 $  \\
             $\mathrm{MeshStereo}$\cite{Zhang2015} + $\mathrm{TF}$\cite{Yang2015} & $ 27.37 $  & $ 3.81 $  & $ 9.91 $  & $ 21.70 $  & $ 2.26 $  & $ 7.03 $  \\
             $\mathrm{MeshStereo}$\cite{Zhang2015} + $\mathrm{FGF}$\cite{He2015} & $ 29.28 $  & $ 3.96 $  & $ 10.03 $  & $ 23.44 $  & $ 2.38 $  & $ 7.12 $  \\
             $\mathrm{MeshStereo}$\cite{Zhang2015} + $\mathrm{WMF}$\cite{Ma2013} & $ 32.16 $  & $ 3.87 $  & $ 10.10 $  & $ 27.18 $  & $ 2.39 $  & $ 7.30 $  \\
             $\mathrm{MeshStereo}$\cite{Zhang2015} + $\mathrm{DT}$\cite{GastalOliveira2011DomainTransform} & $ 23.97 $  & $ 3.77 $  & $ 10.12 $  & $ 17.92 $  & $ 2.18 $  & $ 7.23 $  \\
             $\mathrm{MeshStereo}$\cite{Zhang2015} + $\mathrm{RBS\,(Ours)}$ & $ 21.43 $  & $ 3.22 $ \cellcolor{Yellow}  & $ 8.72 $ \cellcolor{Yellow}  & $ 15.52 $  & $ 2.03 $ \cellcolor{Yellow}  & $ 6.68 $ \cellcolor{Yellow}  \\
\hline
                   $\mathrm{TMAP}$\cite{Psota2015}                        & $ 23.08 $  & $ 3.98 $  & $ 11.55 $  & $ 16.29 $  & $ 2.24 $  & $ 7.61 $  \\
                   $\mathrm{TMAP}$\cite{Psota2015} + $\mathrm{TF}$\cite{Yang2015} & $ 27.47 $  & $ 3.94 $  & $ 10.90 $  & $ 20.88 $  & $ 2.26 $  & $ 7.04 $  \\
                   $\mathrm{TMAP}$\cite{Psota2015} + $\mathrm{FGF}$\cite{He2015} & $ 32.16 $  & $ 4.17 $  & $ 10.79 $  & $ 25.66 $  & $ 2.50 $  & $ 6.99 $  \\
                   $\mathrm{TMAP}$\cite{Psota2015} + $\mathrm{WMF}$\cite{Ma2013} & $ 33.90 $  & $ 4.11 $  & $ 11.01 $  & $ 28.39 $  & $ 2.56 $  & $ 7.61 $  \\
                   $\mathrm{TMAP}$\cite{Psota2015} + $\mathrm{DT}$\cite{GastalOliveira2011DomainTransform} & $ 24.93 $  & $ 3.86 $  & $ 10.92 $  & $ 17.99 $  & $ 2.15 $  & $ 7.02 $  \\
                   $\mathrm{TMAP}$\cite{Psota2015} + $\mathrm{RBS\,(Ours)}$ & $ 22.79 $ \cellcolor{Yellow}  & $ 3.31 $ \cellcolor{Yellow}  & $ 9.44 $ \cellcolor{Yellow}  & $ 16.09 $ \cellcolor{Yellow}  & $ 2.06 $ \cellcolor{Yellow}  & $ 6.72 $ \cellcolor{Yellow}  \\
\hline
    $\mathrm{SGM}$\cite{Hirschmuller05accurateand}                        & $ 24.37 $  & $ 3.85 $  & $ 10.68 $  & $ 17.91 $ \cellcolor{Yellow}  & $ 2.44 $  & $ 8.04 $  \\
    $\mathrm{SGM}$\cite{Hirschmuller05accurateand} + $\mathrm{TF}$\cite{Yang2015} & $ 31.47 $  & $ 3.82 $  & $ 9.55 $  & $ 25.36 $  & $ 2.51 $  & $ 6.95 $ \cellcolor{Yellow}  \\
    $\mathrm{SGM}$\cite{Hirschmuller05accurateand} + $\mathrm{FGF}$\cite{He2015} & $ 34.40 $  & $ 4.05 $  & $ 9.66 $  & $ 28.42 $  & $ 2.72 $  & $ 7.09 $  \\
    $\mathrm{SGM}$\cite{Hirschmuller05accurateand} + $\mathrm{WMF}$\cite{Ma2013} & $ 35.47 $  & $ 3.97 $  & $ 9.99 $  & $ 30.48 $  & $ 2.76 $  & $ 7.72 $  \\
    $\mathrm{SGM}$\cite{Hirschmuller05accurateand} + $\mathrm{DT}$\cite{GastalOliveira2011DomainTransform} & $ 28.78 $  & $ 3.85 $  & $ 9.90 $  & $ 22.24 $  & $ 2.49 $  & $ 7.28 $  \\
    $\mathrm{SGM}$\cite{Hirschmuller05accurateand} + $\mathrm{RBS\,(Ours)}$ & $ 24.18 $ \cellcolor{Yellow}  & $ 3.44 $ \cellcolor{Yellow}  & $ 9.21 $ \cellcolor{Yellow}  & $ 17.95 $  & $ 2.31 $ \cellcolor{Yellow}  & $ 7.06 $
\end{tabular}
\end{center}
\end{table}

For all Middlebury experiments the parameters of our RBS were: $\sigma_{\mathit{xy}} = \sigma_{\mathit{l}} = \sigma_{\mathit{uv}} = 4$, $\lambda = 0.25$, $\sigma'_{\mathit{xy}} = \sigma'_{\mathit{rgb}} = 4$, $\sigma_{\mathit{gm}} = 1$, (the scale of the Geman-McClure loss function) and we performed $32$ iterations of IRLS.

On our test-set results, we present six error metrics for each image: bad-$1$\% (the percent of pixels whose disparity is wrong by more than $1$), MAE (the mean absolute error of the disparity map) and RMSE (the root mean squared error of each disparity map), for all pixels and for only non-occluded pixels.
We generally see a reduction in the MAE and RMSE error metrics, usually by around $50\%$, and a relatively unchanged bad-$1$\% metrics, which suggests that our solver has a substantial and positive impact on quality.
This improvement is quite clear when visualizing the output depth maps, as shown in Figures~\ref{fig:testset_middlebury1}--\ref{fig:testset_middlebury5}, suggesting that the ``all or nothing''  bad-$1$\% metric does not seem to correlate well with the visual quality of the depth map.
Our error reduction is roughly consistent across all choices of stereo techniques used to produce the input depth maps to our algorithm (though for all post-processing technique the improvement is most significant for MC-CNN, as that is the environment in which parameters were tuned).
The baseline depth post-processing techniques we evaluate against often do reduce some error measures, though all tend to increase the $\mathrm{bad}\,1\%$ error measure, and none produce as substantial a reduction of MAE and RMSE as our approach.
The improvement of our approach over those baseline approaches is evident upon visual inspection, as can be seen in Figures~\ref{fig:middleburyAlgo1} and \ref{fig:middleburyAlgo2}.
Our test-set errors were taken from the Middlebury website, while training-set errors were produced with our own evaluation code.
We do not report runtime for this task, as the runtime of our technique and all baselines is dominated by the time taken by the MC-CNN technique common among all entries.

\subsection{Stereo-based Defocus}

Though the primary focus of this work is the bilateral solver and not the ``defocus'' task of Barron \etal \cite{Barron2015A}, we would be remiss to omit a comparison of our technique with the optimization technique presented in \cite{Barron2015A}, given the similarities between the two techniques.
The stereo algorithm of \cite{Barron2015A} performs brittle block-matching on a rectified stereo pair to produce, for each pixel, an interval (lower and upper values $[l_i, u_i]$ parametrizing an interval) of likely depths for that pixel.
This data term appears to have been chosen for its efficacy, and because its simple piecewise-linear form allowed this pixel-space loss to be easily ``splatted'' into bilateral space to construct a convex (though non-linear) optimization problem.
This data term is not compatible with our bilateral solver, and so we must convert it to the expected input: a per-pixel ``target'' value and ``confidence'' measure.
We simply use the average of the upper and lower interval as the target and the exponentiated negative length of the interval as the confidence:
\begin{equation}
t_i = (l_i + u_i)/2 \quad\quad c_i = \exp(l_i - u_i)
\end{equation}
This simple reparametrization combined with our bilateral solver produces an effective stereo technique for this ``defocus'' task, while being roughly $3.5\times$ faster than the already-efficient solver of \cite{Barron2015A}.
See Figures~\ref{fig:stereo_defocus_supp1} and \ref{fig:stereo_defocus_supp2} for a visualization of our performance relative to \cite{Barron2015A}, using some of the stereo pairs from \cite{Barron2015A}.
Though this is only a modest improvement over \cite{Barron2015A}, it is reassuring that our general-purpose bilateral solver not only applies to a host of problems in computer vision, but also performs at least as well as its much more specialized precursor technique.
Note that \cite{Barron2015A} reported large errors on the Middlebury V2 dataset, while our technique can be used for both accurate depth maps and pleasing graphical effects.

\subsection{Depth Superresolution}

\begin{table}[b!]
\caption{
Performance on the depth superresolution task of \cite{ferstl2013b}.
We report root-means-squared error, as was done in \cite{Kwon2015}, along with the geometric mean of those errors over the entire dataset.
Times other than our own are indicated by colors:
Green runtimes are from \cite{ferstl2013b}, blue runtimes are from \cite{Kwon2015}, pink runtimes are from \cite{Lu2015}, and the teal runtime is from \cite{Ma2013}. The beige runtimes were produced by us, but on a different, faster computer.
Algorithms which use external training data are indicated with a dagger.
\label{table:super_supp}
}
\begin{center}
\resizebox{4.8in}{!}{
\large
\begin{tabular}{r@{}r@{}l | c c c c | c c c c | c c c c | c | c }
 && $\mathrm{Method}$ & \multicolumn{4}{c|}{$\mathrm{Art}$} & \multicolumn{4}{c|}{$\mathrm{Books}$} & \multicolumn{4}{c|}{$\mathrm{Moebius}$} & $\mathrm{Avg.}$ & $\mathrm{Time\,(sec)}$ \\
 \hline
A) & &                               $\mathrm{Nearest\,Neighbor}$   &  $ 6.55 $  &  $ 7.41 $  &  $ 8.87 $  &  $ 11.24 $  &  $ 6.16 $  &  $ 6.32 $  &  $ 6.63 $  &  $ 7.36 $  &  $ 6.59 $  &  $ 6.78 $  &  $ 6.98 $  &  $ 7.48 $  &  $ 7.26 $  &  $ 0.003 $  \\
B) & &                                         $\mathrm{Bicubic}$   &  $ 5.32 $  &  $ 6.00 $  &  $ 7.15 $  &  $ 9.35 $  &  $ 5.00 $  &  $ 5.17 $  &  $ 5.46 $  &  $ 5.98 $  &  $ 5.34 $  &  $ 5.52 $  &  $ 5.66 $  &  $ 6.07 $  &  $ 5.91 $  &  $ 0.007 $  \\
C) & $\dagger$ &               $\mathrm{Kiechle\,\etal }$\cite{Kiechle2013}   & \cellcolor{Orange}  $ 2.82 $  &  $ 5.10 $  &  $ 6.83 $  &  $ 10.80 $  &  $ 3.83 $  &  $ 5.10 $  &  $ 6.12 $  &  $ 8.43 $  &  $ 4.50 $  &  $ 5.73 $  &  $ 6.64 $  &  $ 8.96 $  &  $ 5.86 $  &  \cellcolor{Blue} $ 450 $  \\
D) & &                                        $\mathrm{Bilinear}$   &  $ 4.57 $  &  $ 5.53 $  &  $ 6.99 $  &  $ 9.45 $  &  $ 3.94 $  &  $ 4.31 $  &  $ 4.71 $  &  $ 5.38 $  &  $ 4.19 $  &  $ 4.55 $  &  $ 4.83 $  &  $ 5.37 $  &  $ 5.16 $  &  $ 0.004 $  \\
E) & &                      $\mathrm{Liu\,\etal\,}$\cite{Liu2013}   &  $ 4.10 $  &  $ 5.43 $  &  $ 7.69 $  &  $ 11.36 $  &  $ 3.08 $  &  $ 3.87 $  &  $ 4.82 $  &  $ 6.46 $  &  $ 3.18 $  &  $ 4.04 $  &  $ 5.11 $  &  $ 6.62 $  &  $ 5.10 $  &  $ 16.60 $  \\
F) & &                    $\mathrm{Shen\,\etal\,}$\cite{Shen2015}   &  $ 3.49 $  &  $ 4.62 $  &  $ 6.13 $  &  $ 8.68 $  &  $ 2.86 $  &  $ 3.48 $  &  $ 4.43 $  &  $ 5.57 $  &  $ 2.29 $  &  $ 3.07 $  &  $ 4.22 $  &  $ 5.43 $  &  $ 4.24 $  &   $ 31.48 $  \\
G) & &             $\mathrm{Diebel\,\&\,Thrun\,}$\cite{Diebel05b}   &  $ 3.49 $  &  $ 4.41 $  &  $ 6.24 $  &  $ 9.11 $  &  $ 2.06 $  &  $ 3.00 $  &  $ 4.06 $  &  $ 5.13 $  &  $ 2.13 $  &  $ 3.10 $  &  $ 4.14 $  &  $ 5.12 $  &  $ 3.98 $  &  $ - $  \\
H) & &                     $\mathrm{Chan\,\etal }$\cite{chan2008}   &  $ 3.44 $  &  $ 4.38 $  &  $ 5.98 $  &  $ 8.41 $  &  $ 2.09 $  &  $ 2.77 $  &  $ 3.78 $  &  $ 5.45 $  &  $ 2.08 $  &  $ 2.69 $  &  $ 3.73 $  &  $ 5.33 $  &  $ 3.83 $  &  \cellcolor{Green}$ 3.02 $   \\
I) & &         $\mathrm{Guided Filter }$\cite{He2010,ferstl2013b}   &  $ 3.55 $  &  $ 4.31 $  &  $ 5.59 $  &  $ 8.22 $  &  $ 2.37 $  &  $ 2.73 $  &  $ 3.42 $  &  $ 4.52 $  &  $ 2.48 $  &  $ 2.82 $  &  $ 3.54 $  &  $ 4.53 $  &  $ 3.76 $  &  \cellcolor{Green} $ 23.89 $   \\
J) & &                        $\mathrm{Min\,\etal}$\cite{Min2014}   &  $ 3.65 $  &  $ 4.08 $  &  $ 5.09 $  &  $ 7.91 $  &  $ 2.85 $  &  $ 2.77 $  &  $ 2.97 $  &  $ 3.81 $  &  $ 3.46 $  &  $ 3.25 $  &  $ 3.20 $  &  $ 3.86 $  &  $ 3.74 $  &  $ 0.383 $  \\
K) & $\dagger$ &                   $\mathrm{Lu\,\&\,Forsyth }$\cite{Lu2015}   &  $ 4.30 $  &  $ 5.05 $  &  $ 6.33 $  &  $ 7.94 $  &  $ 2.17 $  &  $ 2.71 $  &  $ 3.30 $  &  $ 4.29 $  &  $ 2.16 $  &  $ 2.50 $  &  $ 3.15 $  &  $ 4.10 $  &  $ 3.69 $  &  \cellcolor{Pink} $ 20 $  \\
L) & &                     $\mathrm{Park\,\etal }$\cite{Park2011}   &  $ 3.76 $  &  $ 4.48 $  &  $ 5.80 $  &  $ 8.75 $  &  $ 1.95 $  &  $ 2.60 $  &  $ 3.30 $  &  $ 4.86 $  &  $ 1.96 $  &  $ 2.49 $  &  $ 3.21 $  &  $ 4.48 $  &  $ 3.61 $  &  \cellcolor{Green}$ 24.05 $   \\
M) & &$\mathrm{Domain\,Transform\,}$\cite{GastalOliveira2011DomainTransform}   &  $ 3.95 $  &  $ 4.76 $  &  $ 6.14 $  &  $ 8.49 $  &  $ 1.80 $  &  $ 2.40 $  &  $ 3.23 $  &  $ 4.44 $  &  $ 1.83 $  &  $ 2.40 $  &  $ 3.35 $  &  $ 4.64 $  &  $ 3.56 $  &  $ 0.021 $  \\
N) & &                        $\mathrm{Ma\,\etal\,}$\cite{Ma2013}   &  $ 3.27 $  &  $ 3.99 $  &  $ 5.08 $  & \cellcolor{Yellow}  $ 7.39 $  &  $ 2.39 $  &  $ 2.70 $  &  $ 3.09 $  &  $ 3.77 $  &  $ 2.55 $  &  $ 2.84 $  &  $ 3.23 $  &  $ 3.81 $  &  $ 3.49 $  &  \cellcolor{Teal} $ 18 $  \\
O) & &            $\mathrm{Guided Filter (Matlab) }$\cite{He2010}   &  $ 3.60 $  &  $ 4.25 $  &  $ 5.49 $  &  $ 7.99 $  &  $ 2.39 $  &  $ 2.52 $  &  $ 2.89 $  &  $ 3.89 $  &  $ 2.50 $  &  $ 2.57 $  &  $ 2.90 $  &  $ 3.61 $  &  $ 3.47 $  &  $ 0.434 $  \\
P) & &                  $\mathrm{Zhang\,\etal\,}$\cite{Zhang2014}   &  $ 4.15 $  &  $ 4.22 $  &  $ 5.03 $  &  $ 7.86 $  &  $ 1.96 $  &  $ 2.24 $  &  $ 3.13 $  &  $ 4.80 $  &  $ 1.80 $  &  $ 2.19 $  &  $ 3.22 $  &  $ 4.90 $  &  $ 3.45 $  &  \cellcolor{WindowsColor} $ 1.346 $  \\
Q) & &                $\mathrm{Fast Guided Filter }$\cite{He2015}   &  $ 3.40 $  &  $ 4.16 $  &  $ 5.46 $  &  $ 7.97 $  &  $ 2.08 $  &  $ 2.51 $  &  $ 3.04 $  &  $ 3.95 $  &  $ 2.13 $  &  $ 2.55 $  &  $ 3.08 $  &  $ 3.79 $  &  $ 3.41 $  &  $ 0.225 $  \\
R) & &                     $\mathrm{Yang\,2015\,}$\cite{Yang2015}   &  $ 3.27 $  &  $ 4.15 $  &  $ 5.46 $  &  $ 7.93 $  &  $ 2.00 $  &  $ 2.38 $  &  $ 3.00 $  &  $ 4.04 $  &  $ 2.25 $  &  $ 2.57 $  &  $ 3.13 $  &  $ 4.00 $  &  $ 3.41 $  &  \cellcolor{WindowsColor} $ 0.304 $  \\
S) & &              $\mathrm{Yang\,\etal\,2007\,}$\cite{Yang2007}   &  $ 3.01 $  &  $ 3.92 $  & \cellcolor{Yellow}  $ 4.85 $  &  $ 7.57 $  &  $ 1.87 $  &  $ 2.38 $  &  $ 2.86 $  &  $ 4.26 $  &  $ 1.92 $  &  $ 2.41 $  &  $ 2.96 $  &  $ 4.37 $  &  $ 3.25 $  &  $ - $  \\
T) & &                 $\mathrm{Farbman\,\etal\,}$\cite{FFLS2008}   &  $ 3.14 $  &  $ 4.00 $  &  $ 5.30 $  &  $ 7.70 $  &  $ 1.76 $  &  $ 2.26 $  &  $ 2.90 $  &  $ 3.88 $  &  $ 1.79 $  &  $ 2.29 $  &  $ 2.98 $  &  $ 3.93 $  &  $ 3.19 $  &   $ 6.11 $  \\
U) & &                  $\mathrm{JBU\,}$\cite{Adams2010,Kopf2007}   &  $ 3.17 $  &  $ 4.02 $  &  $ 5.37 $  &  $ 7.59 $  &  $ 1.83 $  &  $ 2.18 $  &  $ 2.80 $  &  $ 4.00 $  &  $ 1.83 $  &  $ 2.13 $  &  $ 2.71 $  &  $ 3.76 $  &  $ 3.14 $  &  $ 1.98 $  \\
V) & &                $\mathrm{Ferstl\,\etal }$\cite{ferstl2013b}   &  $ 3.19 $  &  $ 4.06 $  &  $ 5.08 $  &  $ 7.61 $  &  $ 1.52 $  &  $ 2.21 $  & \cellcolor{Yellow}  $ 2.47 $  & \cellcolor{Yellow}  $ 3.54 $  &  $ 1.47 $  &  $ 2.03 $  &  $ 2.58 $  & \cellcolor{Yellow}  $ 3.50 $  &  $ 2.93 $  &  \cellcolor{Green}$ 140 $  \\
W) & $\dagger$ &                         $\mathrm{Li\,\etal }$\cite{Li2013}   &  $ 3.02 $  & \cellcolor{Orange}  $ 3.12 $  & \cellcolor{Orange}  $ 4.43 $  &  $ 7.43 $  & \cellcolor{Orange}  $ 1.18 $  & \cellcolor{Orange}  $ 1.70 $  &  $ 2.55 $  &  $ 3.58 $  & \cellcolor{Orange}  $ 1.11 $  & \cellcolor{Orange}  $ 1.59 $  & \cellcolor{Orange}  $ 2.28 $  &  $ 3.50 $  & \cellcolor{Orange}  $ 2.56 $  &  \cellcolor{Blue} $ 700 $  \\
X) & $\dagger$ &                     $\mathrm{Kwon\,\etal }$\cite{Kwon2015}   & \cellcolor{Red}  $ 0.87 $  & \cellcolor{Red}  $ 1.30 $  & \cellcolor{Red}  $ 2.05 $  & \cellcolor{Red}  $ 3.56 $  & \cellcolor{Red}  $ 0.51 $  & \cellcolor{Red}  $ 0.75 $  & \cellcolor{Red}  $ 1.14 $  & \cellcolor{Red}  $ 1.88 $  & \cellcolor{Red}  $ 0.57 $  & \cellcolor{Red}  $ 0.89 $  & \cellcolor{Red}  $ 1.37 $  & \cellcolor{Red}  $ 2.14 $  & \cellcolor{Red}  $ 1.21 $  &  \cellcolor{Blue}$ 300 $  \\
\hline
Y) & &                                      $\mathrm{BS\,(Ours)}$   & \cellcolor{Yellow}  $ 2.93 $  & \cellcolor{Yellow}  $ 3.79 $  &  $ 4.95 $  & \cellcolor{Orange}  $ 7.13 $  & \cellcolor{Yellow}  $ 1.39 $  & \cellcolor{Yellow}  $ 1.84 $  & \cellcolor{Orange}  $ 2.38 $  & \cellcolor{Orange}  $ 3.29 $  & \cellcolor{Yellow}  $ 1.38 $  & \cellcolor{Yellow}  $ 1.80 $  & \cellcolor{Yellow}  $ 2.38 $  & \cellcolor{Orange}  $ 3.23 $  & \cellcolor{Yellow}  $ 2.70 $  &  $ 0.234 $
\end{tabular}
}
\end{center}
\end{table}

In Table~\ref{table:super_supp} we present an expanded form of the depth superresolution results presented in the main paper, in which the task is broken down into its constituent images and scale factors, as was done in past work \cite{ferstl2013b,Kwon2015}.
Our bilateral solver appears to be roughly the second or third most accurate technique, after \cite{Kwon2015} and sometimes \cite{Li2013}.
But it should be noted that these top-performing techniques \cite{Kiechle2013,Kwon2015,Li2013} are all dictionary-based techniques which are trained on a great deal of data, while our technique produces competitive results despite the simplicity of our model and our lack of training.
What is perhaps the most important property of our model is its speed --- our technique $600$-$3000 \times$ faster than the three most accurate techniques, despite having comparable accuracy to all but \cite{Kwon2015}.
Our performance is most competitive when the upsampling factor is large ($16\times$) which is when the task is inherently most difficult.
Additionally, those techniques with speeds comparable to or better than the bilateral solver \cite{GastalOliveira2011DomainTransform,Yang2015,He2015,Min2014} produce error rates that are 25-40$\%$ greater than our approach.
The only techniques with significantly faster runtimes than the bilateral solver are standard image interpolation techniques (bilinear, bicubic, etc) which produce low-quality output, and our implementation of the domain transform \cite{GastalOliveira2011DomainTransform} which is a highly-optimized, vectorized, and multi-threaded implementation, unlike our own technique and all other techniques we compare against.
More conventional implementations of the domain transform appear to have runtimes comparable to our own, or that of the guided filter\cite{He2010}.
See Figures~\ref{fig:super1}-\ref{fig:super3} for example output for this task.

For transparency and completeness we evaluated against as many applicable baseline techniques as was possible.
This is difficult, as though many papers present results for denoising or upsampling different versions of the Middlebury dataset, there is not one standard benchmark which is universally adopted.
To this end, we build upon the benchmarked used in \cite{ferstl2013b}, but augment it heavily.
We inherit some results from past work \cite{ferstl2013b,Kwon2015} which is necessary due to the lack of publicly available source code for many techniques.
This means that some runtimes are produced on other hardware platforms, and so these runtimes should be considered accordingly.
Our own runtimes (indicated in Table~\ref{table:super_supp} in white) were benchmarked on a 2012 HP Z420 workstation (Intel Xeon CPU E5-1650, 3.20GHz, 32 GB RAM), and our algorithm is implemented in standard, single-threaded C++.
Runtimes highlighted with a color come from some other hardware platform other than our reference hardware.
For clarity, we will now detail the experimental conditions of each baseline technique, both in terms of parameter settings and hardware environments.
For all experiments in which we produced the model output, the parameters of each algorithm were tuned to minimize the average error on this task by starting with the baseline parameters in the publicly available code, and then performing coordinate descent on each parameter, halving and doubling each and taking the new parameter setting if the average error decreased.
For all models in which the code was ran by us, the runtime reported is the median of the runtime over all images in the dataset, unless the model's runtime is resolution-dependent, in which case we report the geometric mean of the runtimes.
\begin{description}[align=left]
\item [A] This is nearest-neighbor interpolation, and the reported runtime is that of the Matlab $\mathrm{imresize()}$ function on our workstation.
\item [B] This is bicubic interpolation, and the reported runtime is that of the Matlab $\mathrm{imresize()}$ function on our workstation.
\item [C] This is the technique of Kiechle \etal \cite{Kiechle2013}, with errors and runtimes  taken directly from \cite{Kwon2015}.
\item [D] This is bilinear interpolation, and the reported runtime is that of the Matlab $\mathrm{imresize()}$ function on our workstation.
\item [E] This is the Joint Geodesic Upsampling technique of \cite{Liu2013}, which we ran ourselves using the publicly available code\footnote{\url{http://www.merl.com/research/license/}}. This technique performs poorly for this task, possibly because it assumes noise-free input. We used the default parameters of the code, which we found optimal for this task: $\mathrm{interval} = 3$, $\sigma = 0.5$,  $\lambda_1 = 10$, $\lambda_2 = 1$.
\item [F] This is the Mutual-Structure technique of Shen \etal \cite{Shen2015}, which we ran ourselves using the publicly available code\footnote{\url{http://www.cse.cuhk.edu.hk/leojia/projects/mutualstructure/index.html}}. This technique performs poorly on this task, which appears to be due to the low-frequency nature of the depth-map noise inherent in this depth super-resolution task, unlike the high-frequency, per-pixel noise used in the experiments of \cite{Shen2015}. The parameters were tuned for optimal performance on our task: $\epsilon_I = 10^{-5}$, $\epsilon_G = 2 \times 10^{-4}$, $\lambda_I = 1$, $\lambda_G = 4$, $20$ iterations, window size of $2$. A more aggressive smoothing effect could be achieved using larger windows sizes which were a function of the upsampling factor, but these oversmoothed depths had significantly higher errors than was produced using these settings.
\item [G] This is the technique presented in Diebel \& Thrun \cite{Diebel05b}, which was used as a baseline algorithm in \cite{ferstl2013b}. We did not run this code ourselves, and produced these error rates using the precomputed output from \cite{ferstl2013b}, which is why we do not have runtimes.
\item [H] This is an implementation of Chan \etal \cite{chan2008}, which was used as a baseline algorithm in \cite{ferstl2013b}. We did not run this code ourselves, and produced these error rates using the precomputed output from \cite{ferstl2013b}, and reproduced the runtime quoted in \cite{ferstl2013b}.
\item [I] This is presumably an implementation of the guided filter \cite{He2010}, which was used as a baseline algorithm in \cite{ferstl2013b}. We did not run this code ourselves, and produced these error rates using the precomputed output from \cite{ferstl2013b}, and we took the runtime quoted in \cite{ferstl2013b}. Because the quoted runtime seems unusually slow for the guided filter, we ran two of our own evaluations of the guided filter on this task (models O and Q) which produced lower errors rates and significantly faster runtimes.
\item [J] This is the weighted least-squares approach of Min \etal \cite{Min2014}, which we ran ourselves using publicly available code\footnote{\url{https://github.com/soundsilence/ImageSmoothing}}. The parameters were tuned for optimal performance on our task: $\sigma = 0.00125 \times 2^f$ (where $f$ is the upsampling factor), $\lambda = 30^2$, $\mathrm{iteration} = 3$, $\mathrm{attenuation} = 4$.
\item [K] This is the technique presented in Lu \& Forsyth 2015 \cite{Lu2015}. The authors of \cite{Lu2015} ran their own code on the task presented here on their own computer (with comparable specifications to our own) at our request, sent us the output, and reported their approximate runtime, which we report here.
\item [L] This is the technique presented in Park \etal \cite{Park2011}, which was used as a baseline algorithm in \cite{ferstl2013b}. We did not run this code ourselves, and produced these error rates using the precomputed output from \cite{ferstl2013b}, which is why we do not have runtimes.
\item [M] This is the recursive formulation of the domain transform \cite{GastalOliveira2011DomainTransform} with optimally-tuned parameters ($\sigma_r = 64$, $\sigma_s = 8f$). We used our own highly optimized implementation of this code, implemented in Halide, with heavily parallelization and vectorization. This is unlike most other baselines and our own algorithm, which are single-threaded unoptimized code.
\item [N] This is the weighted median filter of \cite{Ma2013}, which we ran ourselves using the publicly available code\footnote{\url{http://research.microsoft.com/en-us/um/people/kahe/iccv13wmf/matlab_wmf_release_v1.rar}}. We found the color version of the code (which we used for its improved accuracy) to perform slowly, so the runtime we cite is taken by extrapolating from the quoted runtime for the CPU implementation of \cite{Ma2013}, ($60$ms per megapixel-disparity) on a comparable computer to ours, suggesting that runtime on this task would be $\sim 18$ seconds per image ($\sim 200$ disparities, $\sim 1.5$ megapixels per image). The parameters were tuned for optimal performance on our task: filter size = $2^f$ (where $f$ is the upsampling factor) and $\epsilon = 0.02^2$.
\item [O] This is the implementation of the guided filter \cite{He2010} which is built into the 2015 version of Matlab. The parameters were tuned for this task,
``NeighborhoodSize'' = $3 \times 2^f$, (where $f$ is the upsampling factor) and ``DegreeOfSmoothing'' = $0.5$.
\item [P] This is the weighted median filter of \cite{Zhang2014}, which we ran ourselves using the publicly available code\footnote{\url{http://www.cse.cuhk.edu.hk/~leojia/projects/fastwmedian/download/JointWMF_mex.zip}}. Because the code was only available as a compiled windows binary, we were forced to use a different computer for this evaluation, and so the runtime is for a significantly faster computer than was used for our other evaluation: A dual-processor Xeon CPU-E5-2690 v3 2.6GHz with 64 GB of ram. The runtime should therefore be considered a lower bound on the actual runtime for the reference hardware. We used the default parameter settings provided with the reference code for our experiments, which we found to perform optimally for our benchmark.
\item [Q] This is our own Matlab implementation of the (color) fast guided filter \cite{He2015}, which we found to be faster than the implementation built into Matlab 2015, and significantly faster than the released code\footnote{\url{http://research.microsoft.com/en-us/um/people/kahe/eccv10/fast-guided-filter-code-v1.rar}} while producing identical results. The parameters were tuned for this task, with a box filter size of $2^f$ (where $f$ is the upsampling factor), $\epsilon = 0.02^2$, and a subsampling of $2$. Our experimentation suggests that the subsampling could be made more aggressive with little drop in accuracy, but because the algorithm requires that the filter size must be divisible by the subsampling factor, the most aggressive subsampling we could use for all upsampling factors was $2$. From this we can assume that a $4\times$ speedup may be possible.
\item [R] This is the technique presented in Yang \cite{Yang2015}, which we ran ourselves using the publicly available code\footnote{\url{http://www.cs.cityu.edu.hk/~qiyang/publications/software/tree_filter.zip}}. Because the code was only available as a compiled windows binary, like in Model P this evaluation was performed on a different, faster workstation, and so the runtime should be considered a lower bound on the actual runtime for the reference hardware. The parameter settings used in these experiments are $\sigma_r = 2^{(f-6)}$ (where $f$ is the upsampling factor) and nonlocal filtering, which experimentation showed to be the optimal parameter settings for this task.
\item [S] This is the technique presented in Yang \etal \cite{Yang2007}, which was used as a baseline algorithm in \cite{ferstl2013b}. We did not run this code ourselves, and produced these error rates using the precomputed output from \cite{ferstl2013b}, which is why we do not have runtimes.
\item [T] This is the WLS model of Farbman \etal \cite{FFLS2008}, which we ran ourselves using the publicly available code\footnote{\url{http://www.cs.huji.ac.il/~danix/epd/wlsFilter.m}}, with the code was modified to use non-log color reference images. The parameter settings used in these experiments are $\lambda = 5000 \times 2^{(f-4)}$ (where $f$ is the upsampling factor) and $\alpha = 2$, which experimentation showed to be the optimal parameter settings for this task.
\item [U] This model is standard joint bilateral upsampling \cite{Kopf2007} using a permutohedral lattice \cite{Adams2010} with optimally-tuned parameters ($\sigma_{rgb} = 8$, $\sigma_x = \sigma_y = 8f$, where $f$ is the upsampling factor). These errors and runtimes were produced by us using a C++ implementation of the permutohedral lattice\footnote{\url{https://code.google.com/archive/p/imagestack/}}.
\item [V] These errors and runtimes were taken from the website\footnote{\url{https://rvlab.icg.tugraz.at/project_page/project_tofusion/project_tofsuperresolution.html}} of \cite{ferstl2013b}. Since the publication of this paper, it has been revealed that the error cited in the publication is MAE, not RMSE, so the errors in the paper are incorrect. For details, see the supplement\footnote{\url{https://sites.google.com/site/datadrivendepthcvpr2015/}} of \cite{Kwon2015}.
\item [W]  This is the technique of Li \etal \cite{Li2013}, with errors and runtimes taken directly from \cite{Kwon2015}.
\item [X] This is the technique of Kwon \etal \cite{Kwon2015}, with errors and runtimes taken directly from \cite{Kwon2015}. We attempted to obtain the output depth maps or source code from \cite{Kwon2015}, but the authors were unable to comply with our request, and so we are unable to present the results of \cite{Kwon2015} in our figures or perform a more detailed analysis of how well the algorithm performs, nor can we reproduce these results. Details on the lack of availability of the source and output from this model can be found on the project website\footnote{\url{https://sites.google.com/site/datadrivendepthcvpr2015/}}.
\end{description}

\subsection{Colorization}

See Figure~\ref{fig:colorization_supp} for a comparison of our technique on the colorization task, compared against Levin \etal \cite{Levin2004}.
The images and code from \cite{Levin2004} were taken from the paper's website\footnote{\url{www.cs.huji.ac.il/~yweiss/Colorization/}}, and were produced using the exact least-square solver presented in the paper for still image processing.
Note that \cite{Levin2004} also presents an approximate multigrid optimization which they use for colorizing videos, which is $6 \times$ faster than their exact still image approach but presumably produces inexact results for the still image task.
Several of the techniques we evaluate against for the depth super-resolution task can also be used for colorization, presumably with the same tradeoffs between accuracy and speed seen in that experiment.
Since there is traditionally no ground-truth for this task, we do not present an exhaustive evaluation here.

\subsection{Semantic Segmentation}

\begin{figure*}[p]
	\centering
	{\includegraphics[width=3.3in]{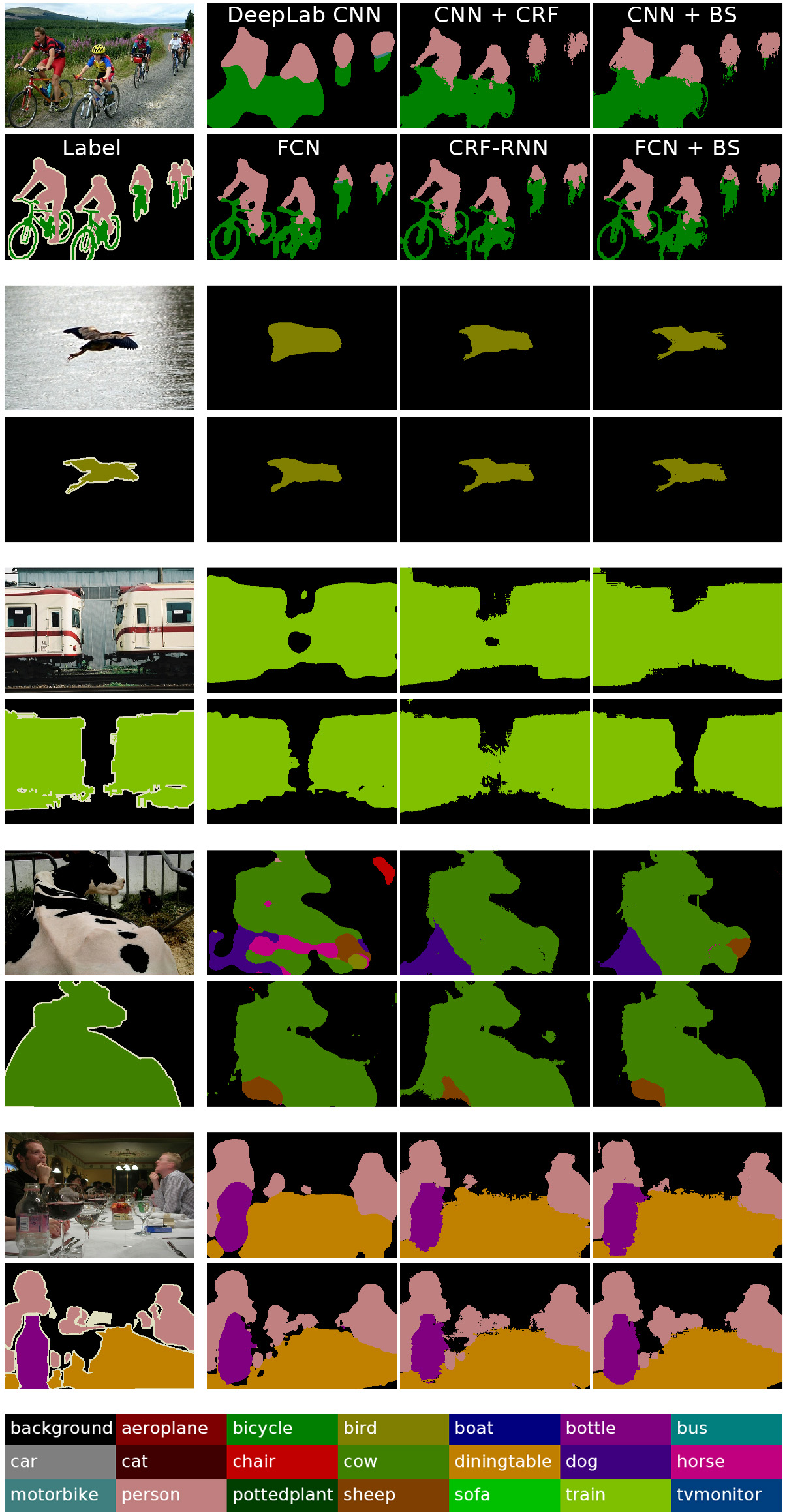}}
	\caption{
    Additional semantic segmentation results on Pascal VOC12 validation images. FCN refers to the fully convolutional network component of the end-to-end trained CRF-RNN model.
    While the CRF-augmented DeepLab model (top rows) and CRF-RNN model (bottom rows) perform best overall, the bilateral solver produces better results on isolated objects (second and third images) at a fraction of the cost.
  }
	\label{fig:supp_seg}
\end{figure*}

See Figure~\ref{fig:supp_seg} for additional examples of our bilateral solver and competing techniques applied to the semantic segmentation task. The bilateral solver produces more edge-aware results than CRF-based techniques on well-isolated objects.

\newcommand{\blurwidthA}{1.5in}
\newcommand{\blurwidthB}{2.3in}

\begin{figure*}[p]
  \centering
  \begin{subfigure}[!]{\blurwidthA}
    \includegraphics[width=\blurwidthA]{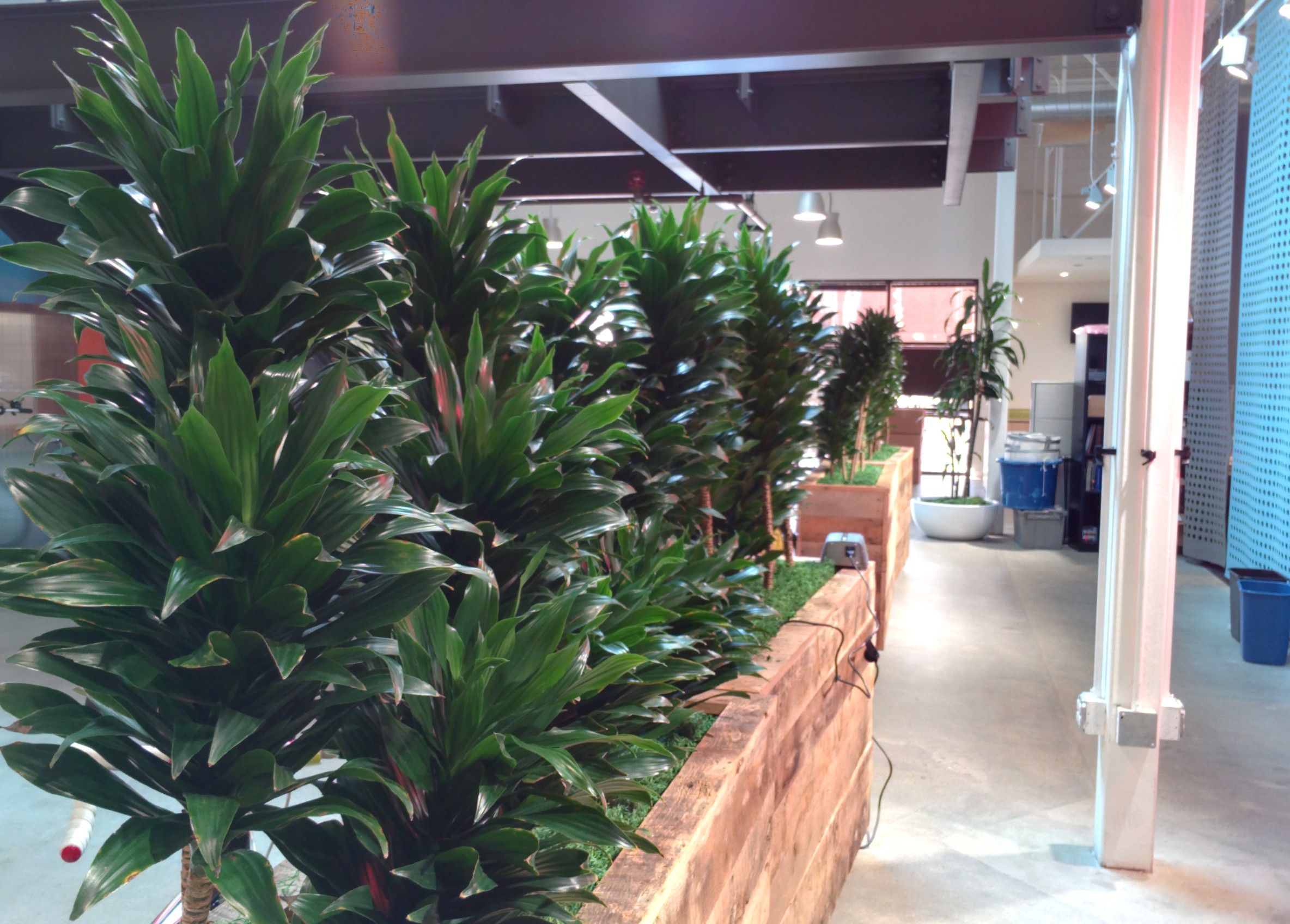}
    \caption{Input Image}
  \end{subfigure}
  \begin{subfigure}[!]{\blurwidthA}
    \includegraphics[width=\blurwidthA]{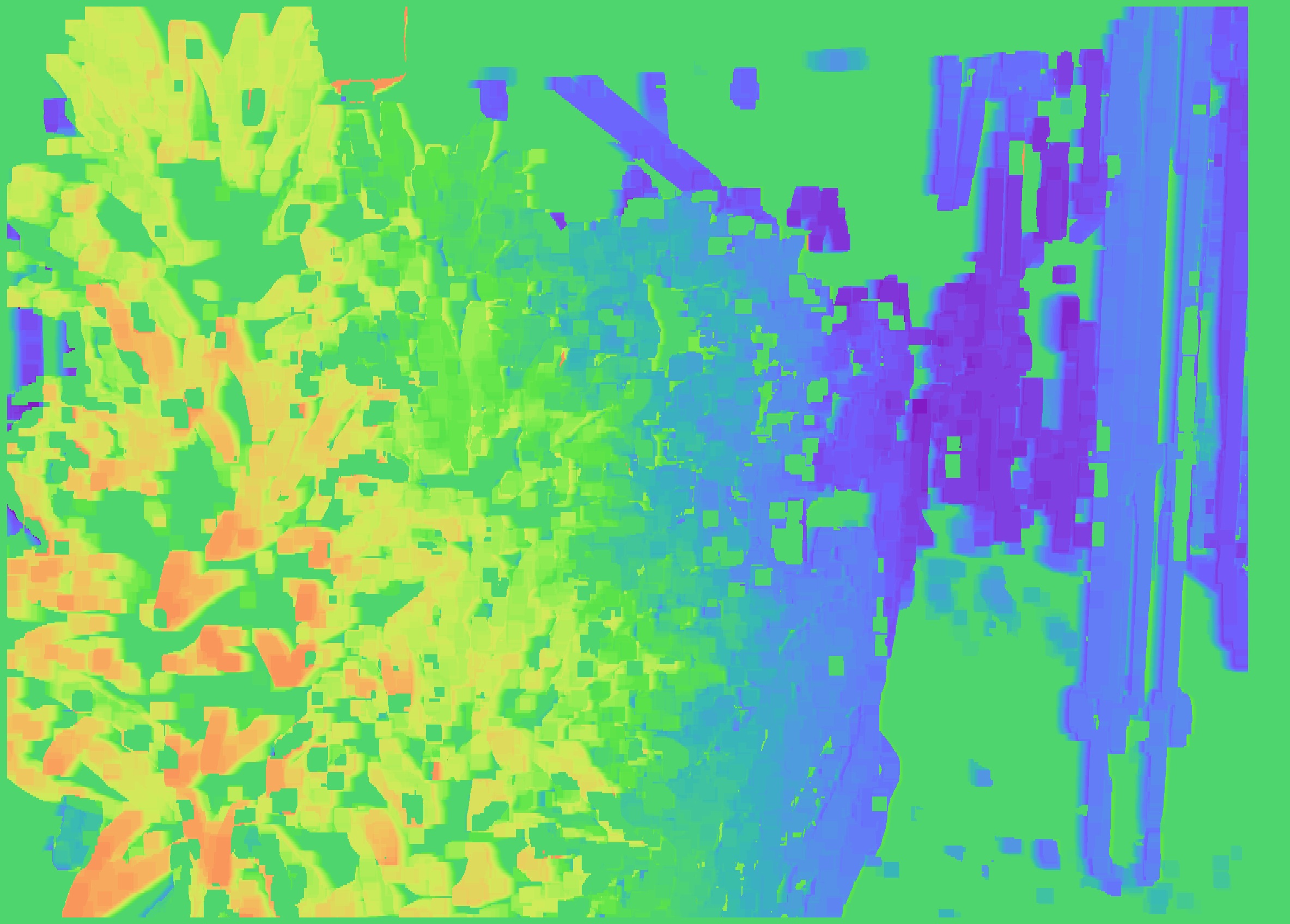}
    \caption{Depth Input \label{fig:defocus_input}}
  \end{subfigure}
  \begin{subfigure}[!]{\blurwidthA}
    \includegraphics[width=\blurwidthA]{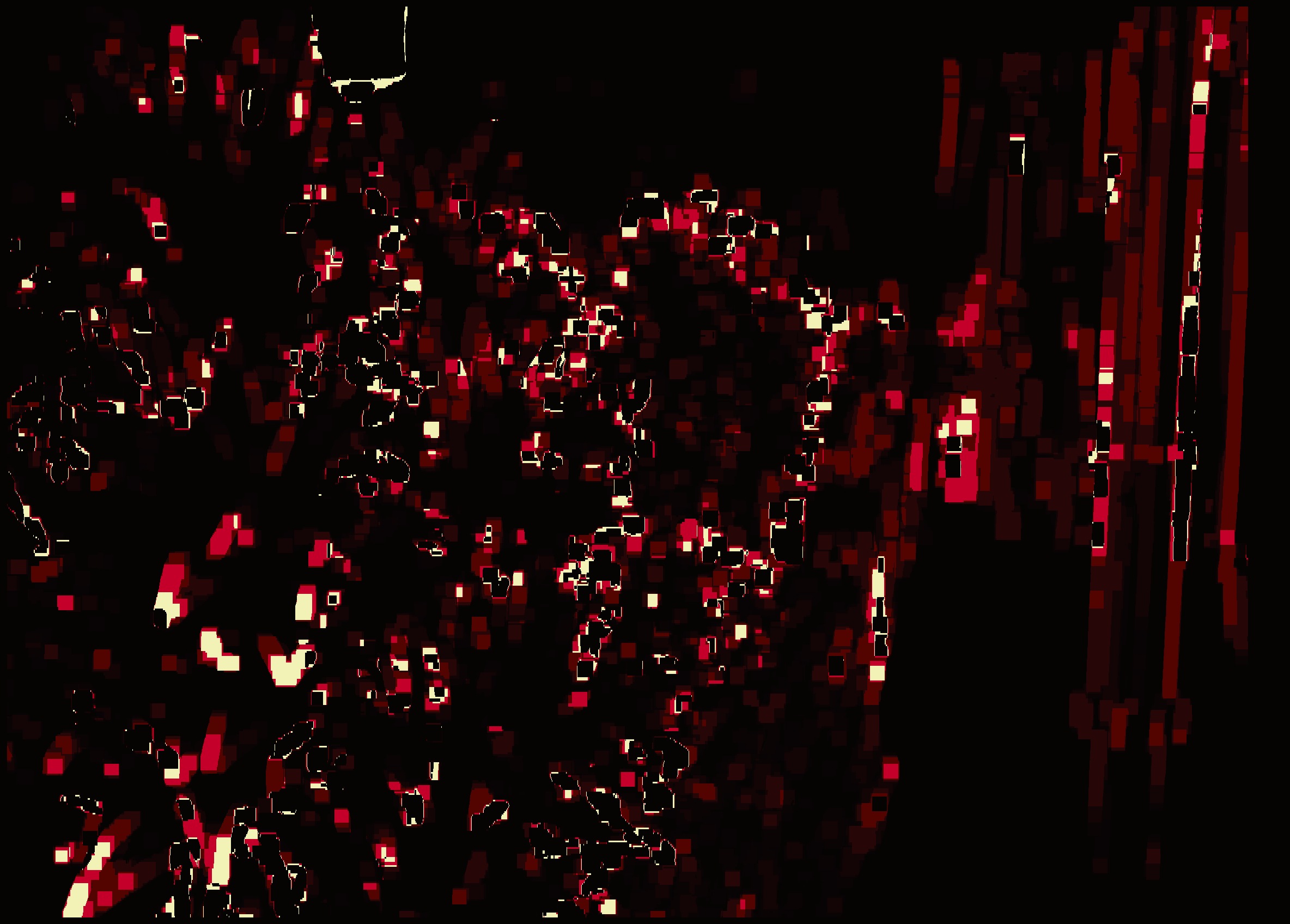}
    \caption{Confidence Input \label{fig:defocus_confidence}}
  \end{subfigure}
  \\
  \begin{subfigure}[!]{\blurwidthB}
    \includegraphics[width=\blurwidthB]{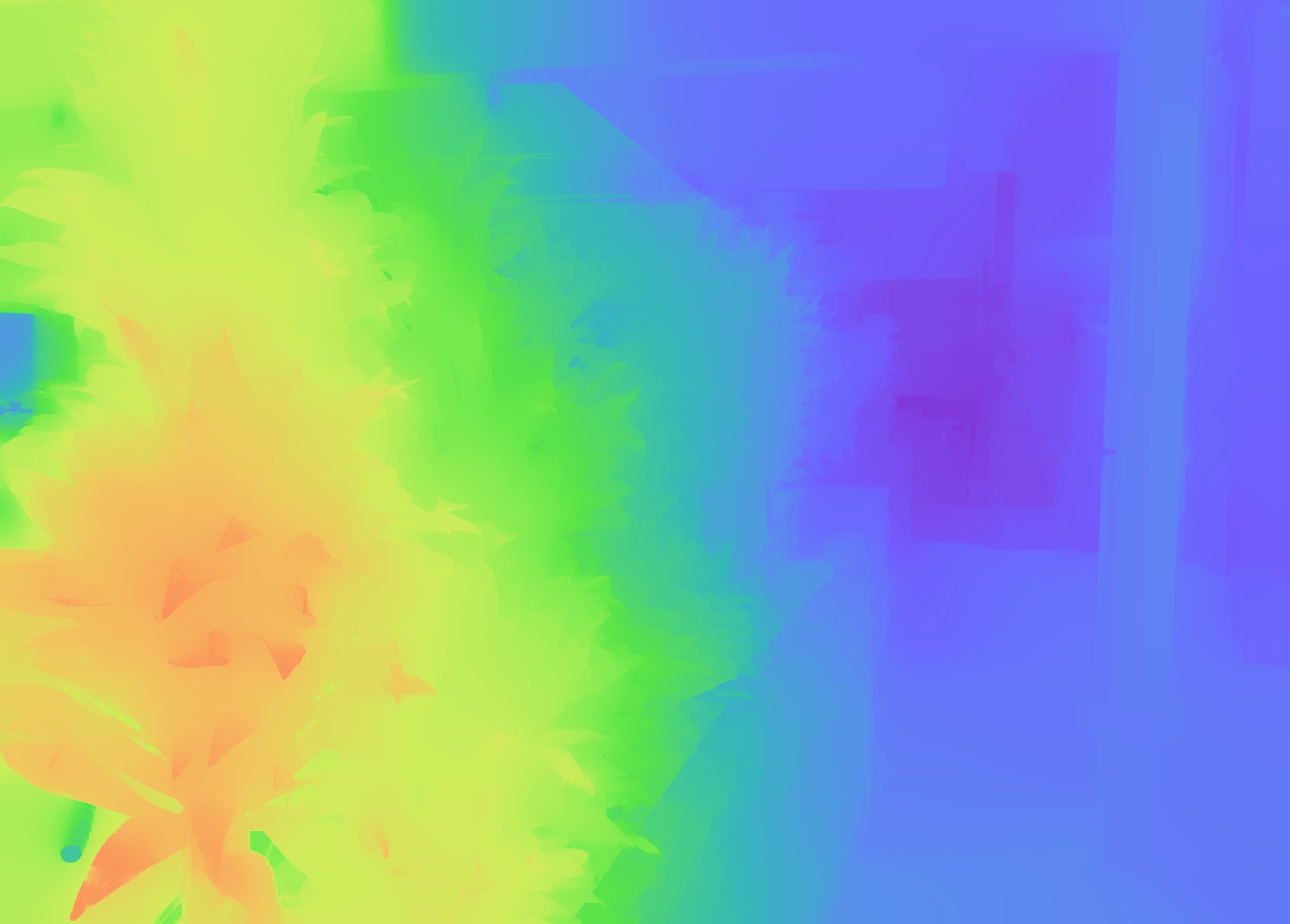}
    \caption{\cite{Barron2015A} Depth Output}
  \end{subfigure}
  \begin{subfigure}[!]{\blurwidthB}
    \includegraphics[width=\blurwidthB]{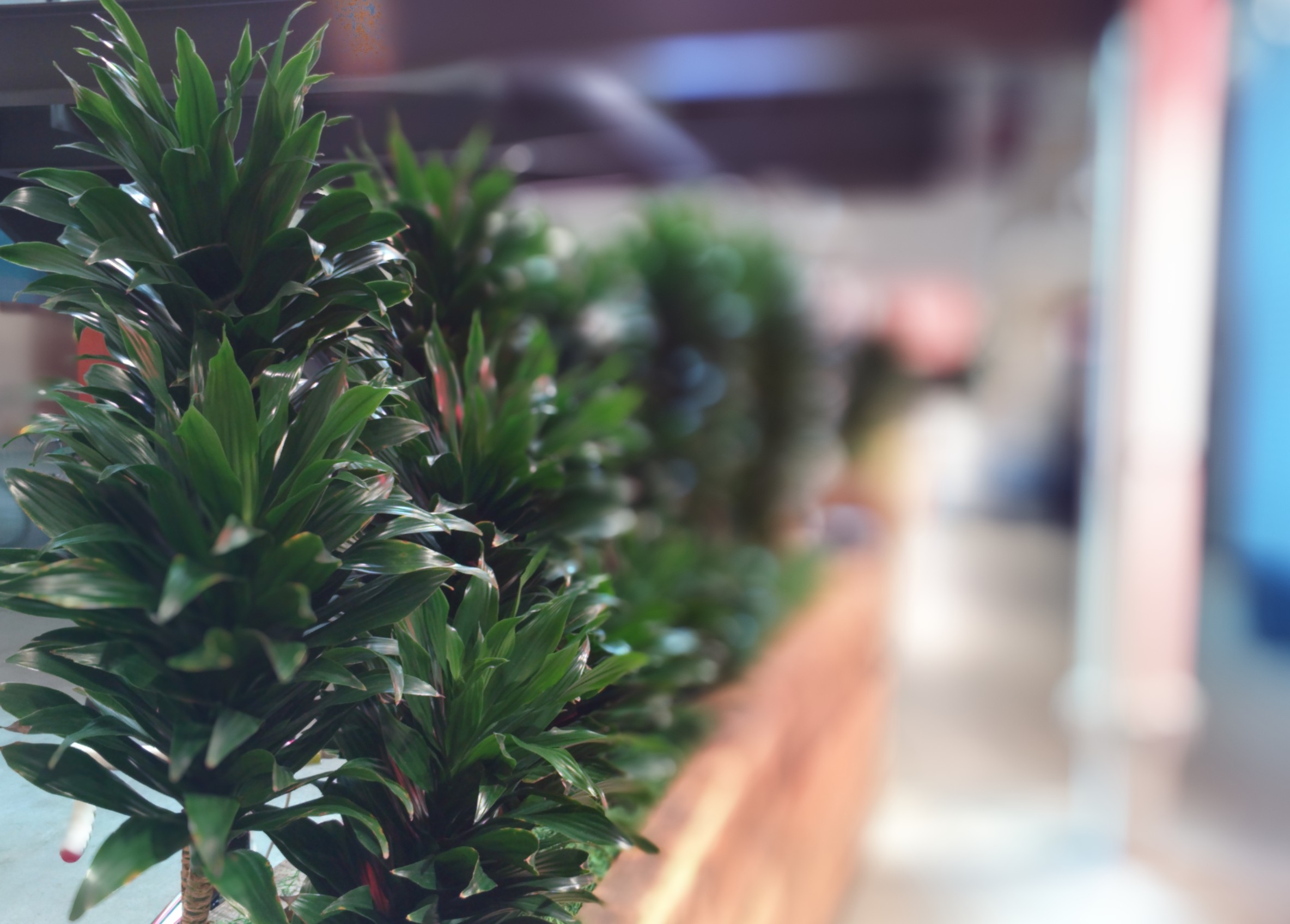}
    \caption{\cite{Barron2015A} Rendering}
  \end{subfigure}
  \\
  \begin{subfigure}[!]{\blurwidthB}
    \includegraphics[width=\blurwidthB]{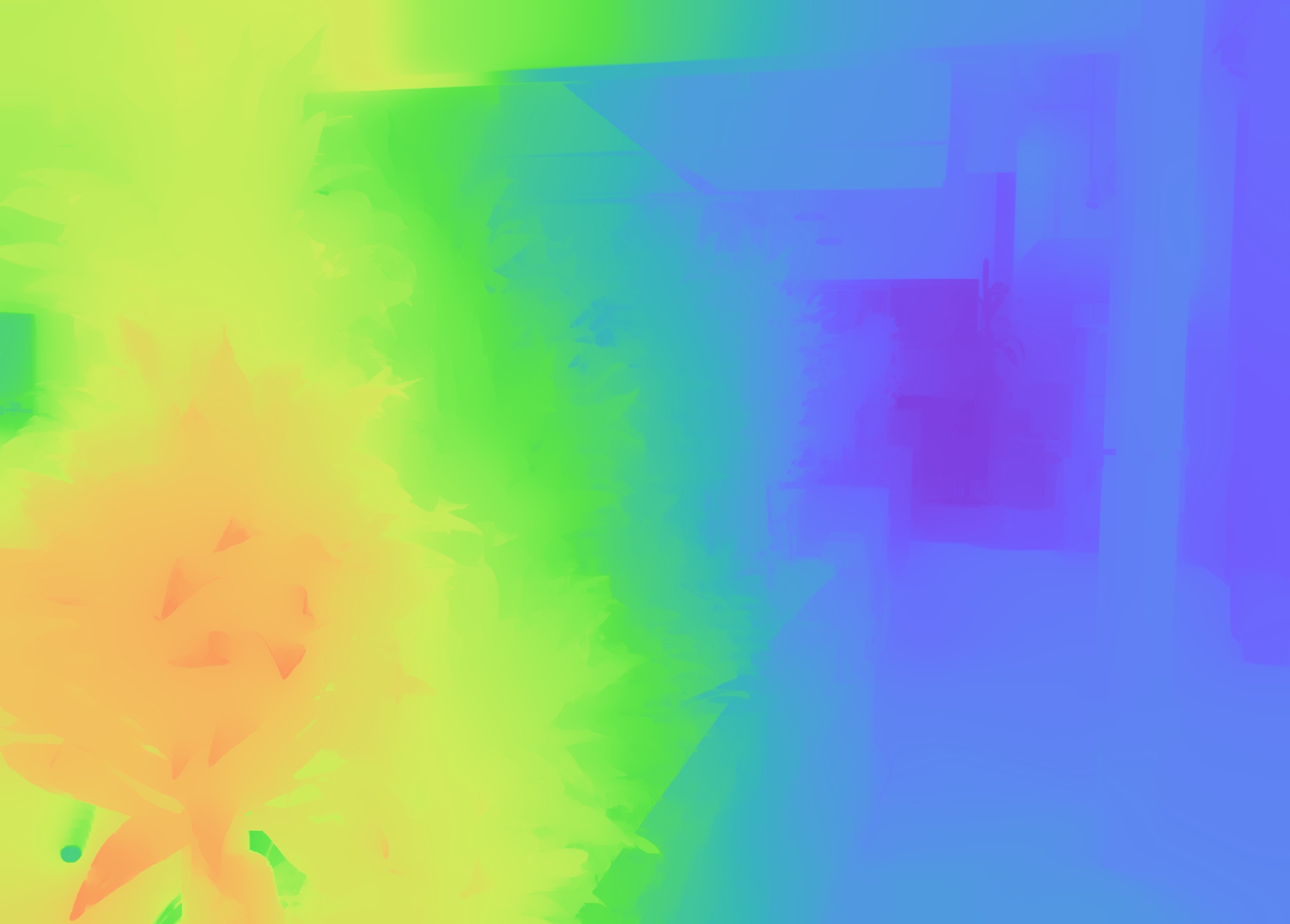}
    \caption{Our Depth Output}
  \end{subfigure}
  \begin{subfigure}[!]{\blurwidthB}
    \includegraphics[width=\blurwidthB]{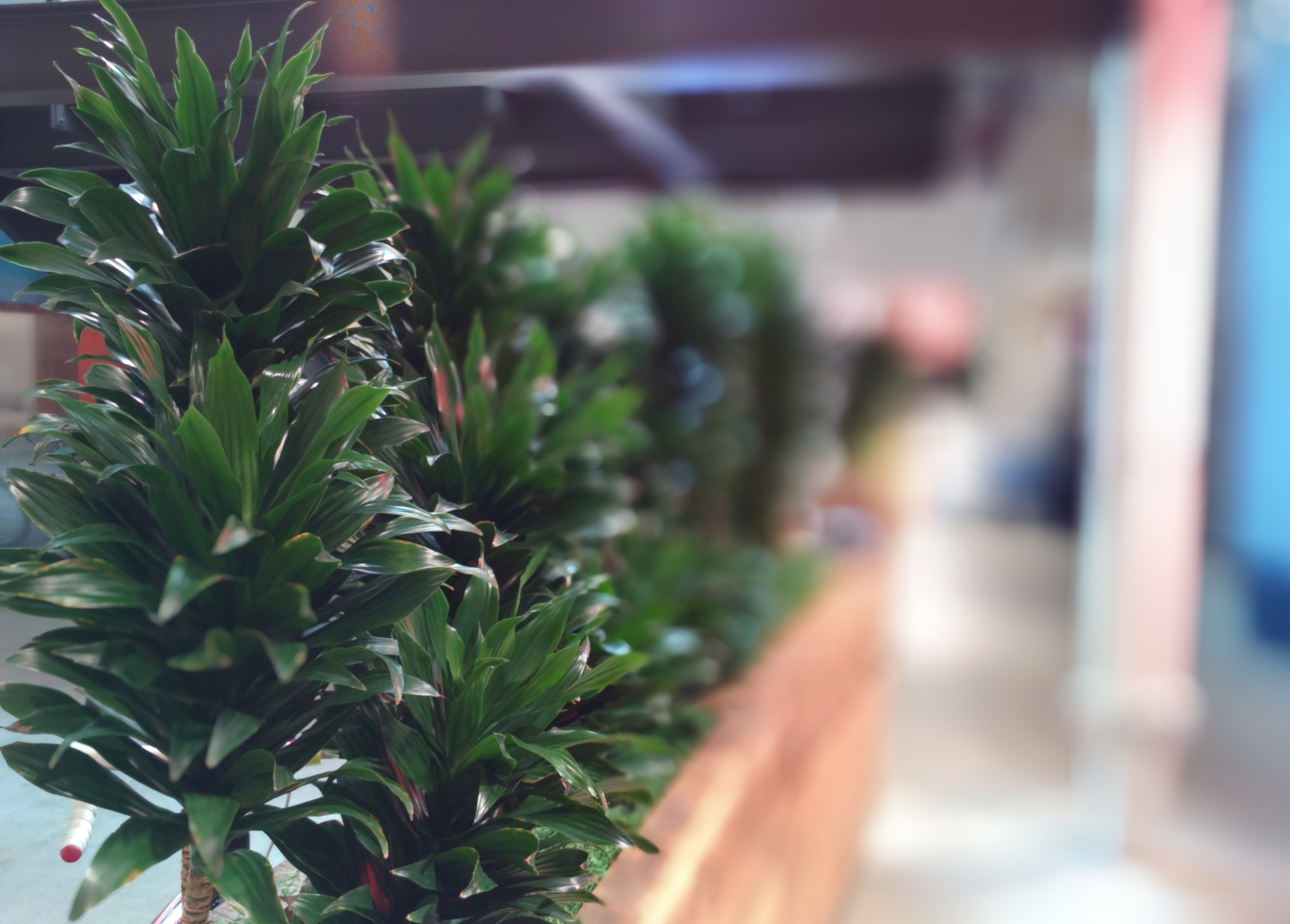}
    \caption{Our Rendering}
  \end{subfigure}
  \caption{
  Given the reparametrized input into the solver of \cite{Barron2015A} (\subref{fig:defocus_input} and \subref{fig:defocus_confidence}) our solver can produce depth maps and defocused renderings of a comparable quality to \cite{Barron2015A}, while being $3.5\times$ faster.
  \label{fig:stereo_defocus_supp1}
  }
\end{figure*}

\begin{figure*}[p]
  \centering
  \begin{subfigure}[!]{\blurwidthA}
    \includegraphics[width=\blurwidthA]{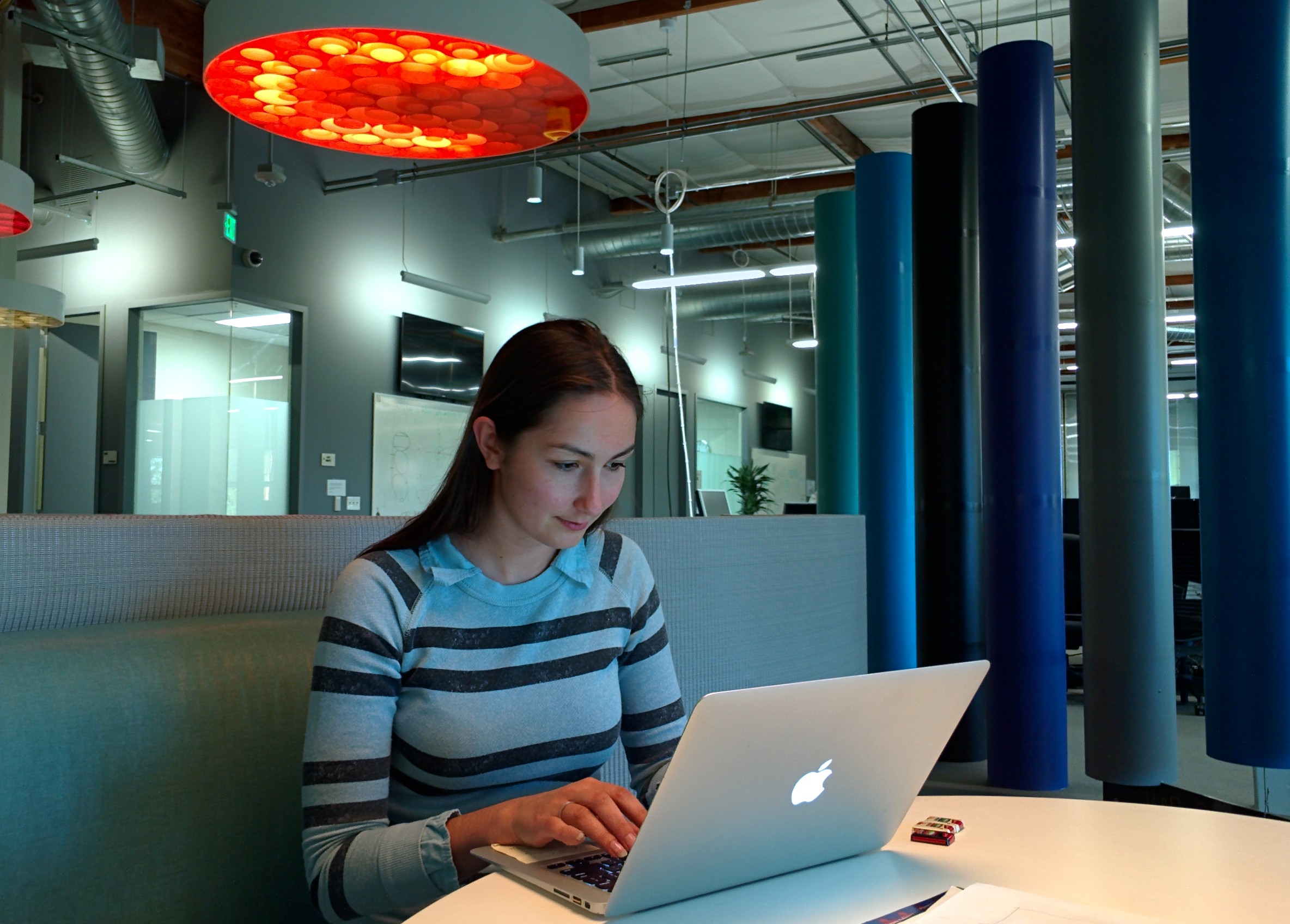}
    \caption{Input Image}
  \end{subfigure}
  \begin{subfigure}[!]{\blurwidthA}
    \includegraphics[width=\blurwidthA]{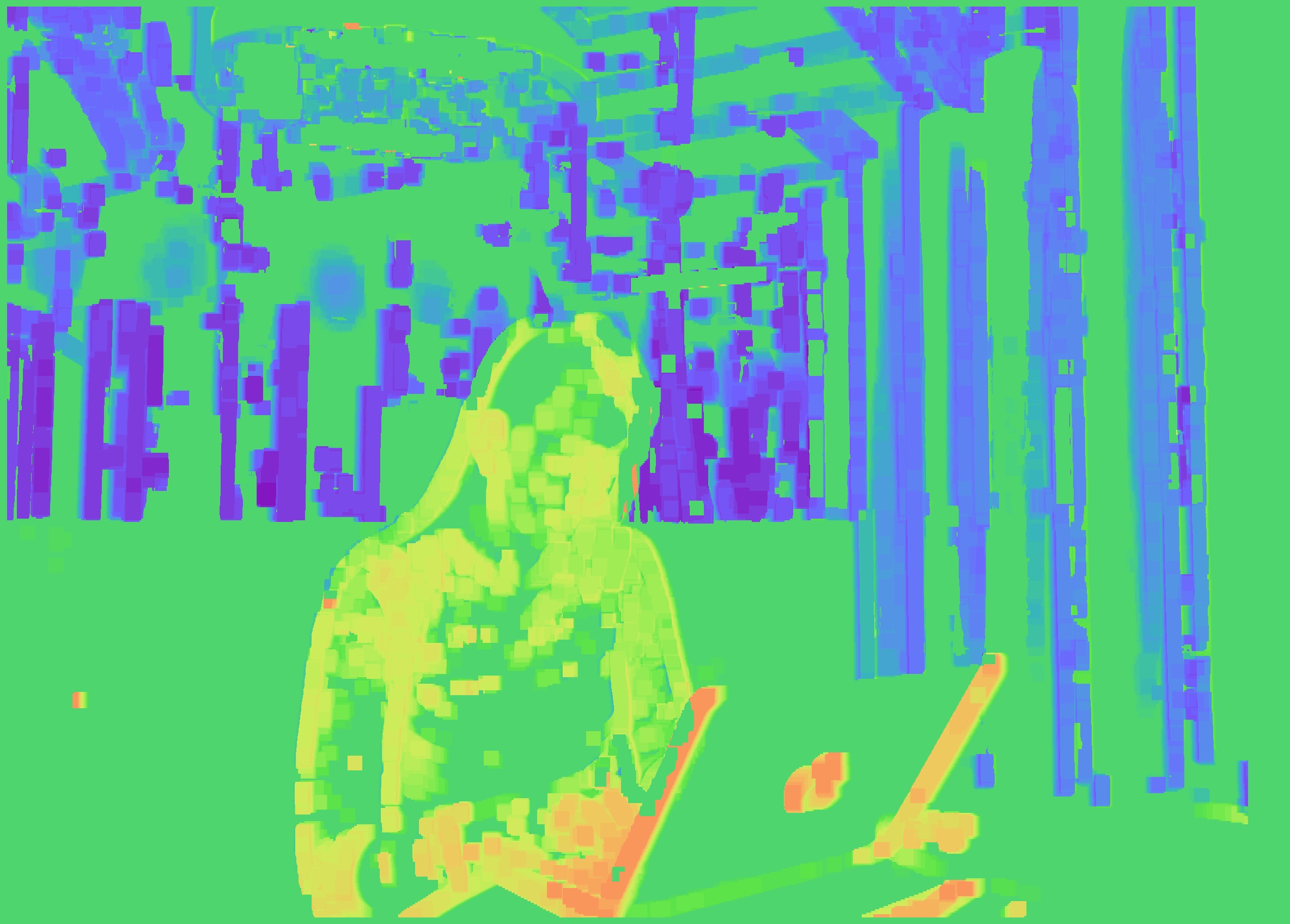}
    \caption{Depth Input}
  \end{subfigure}
  \begin{subfigure}[!]{\blurwidthA}
    \includegraphics[width=\blurwidthA]{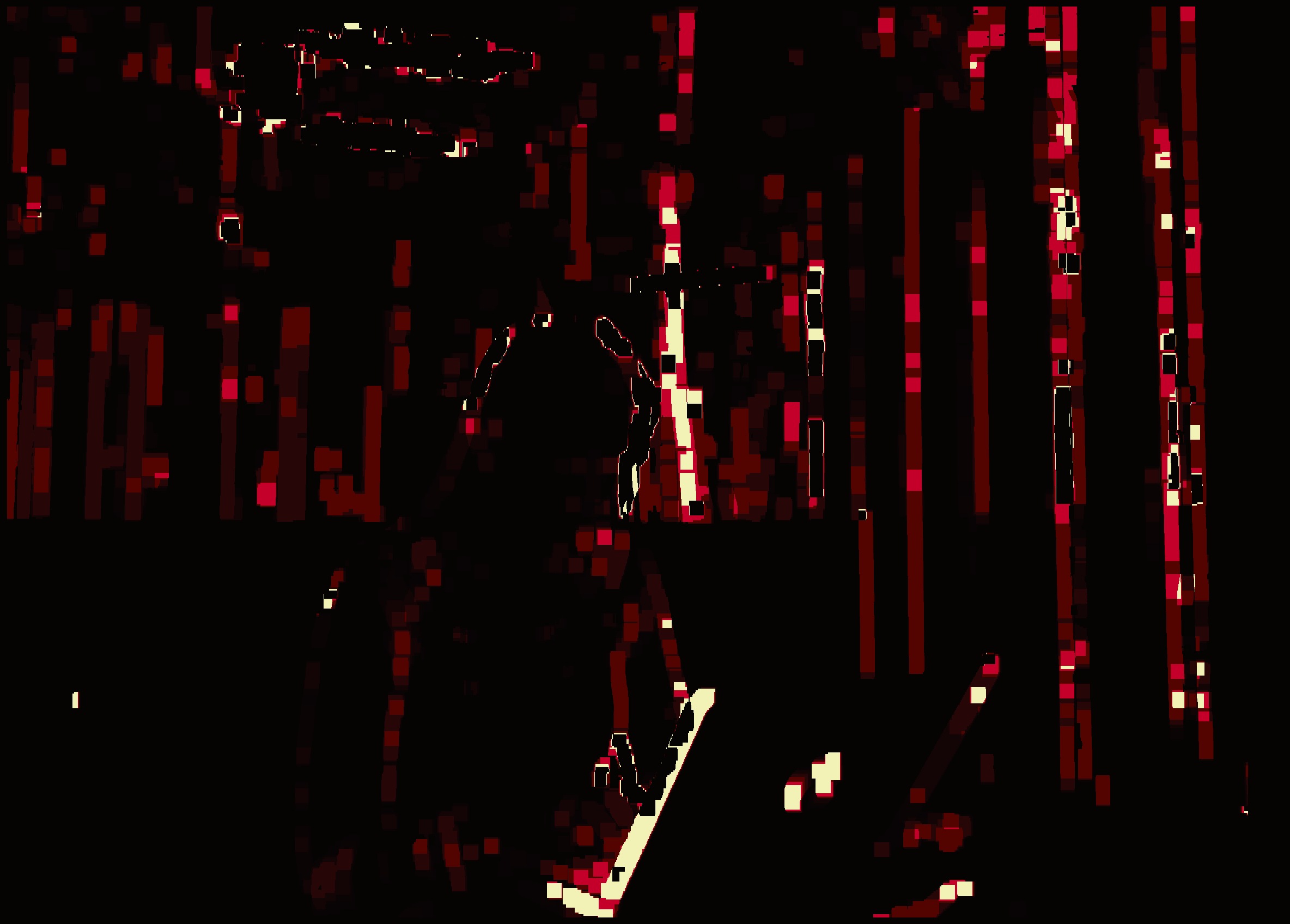}
    \caption{Confidence Input}
  \end{subfigure}
  \\
  \begin{subfigure}[!]{\blurwidthB}
    \includegraphics[width=\blurwidthB]{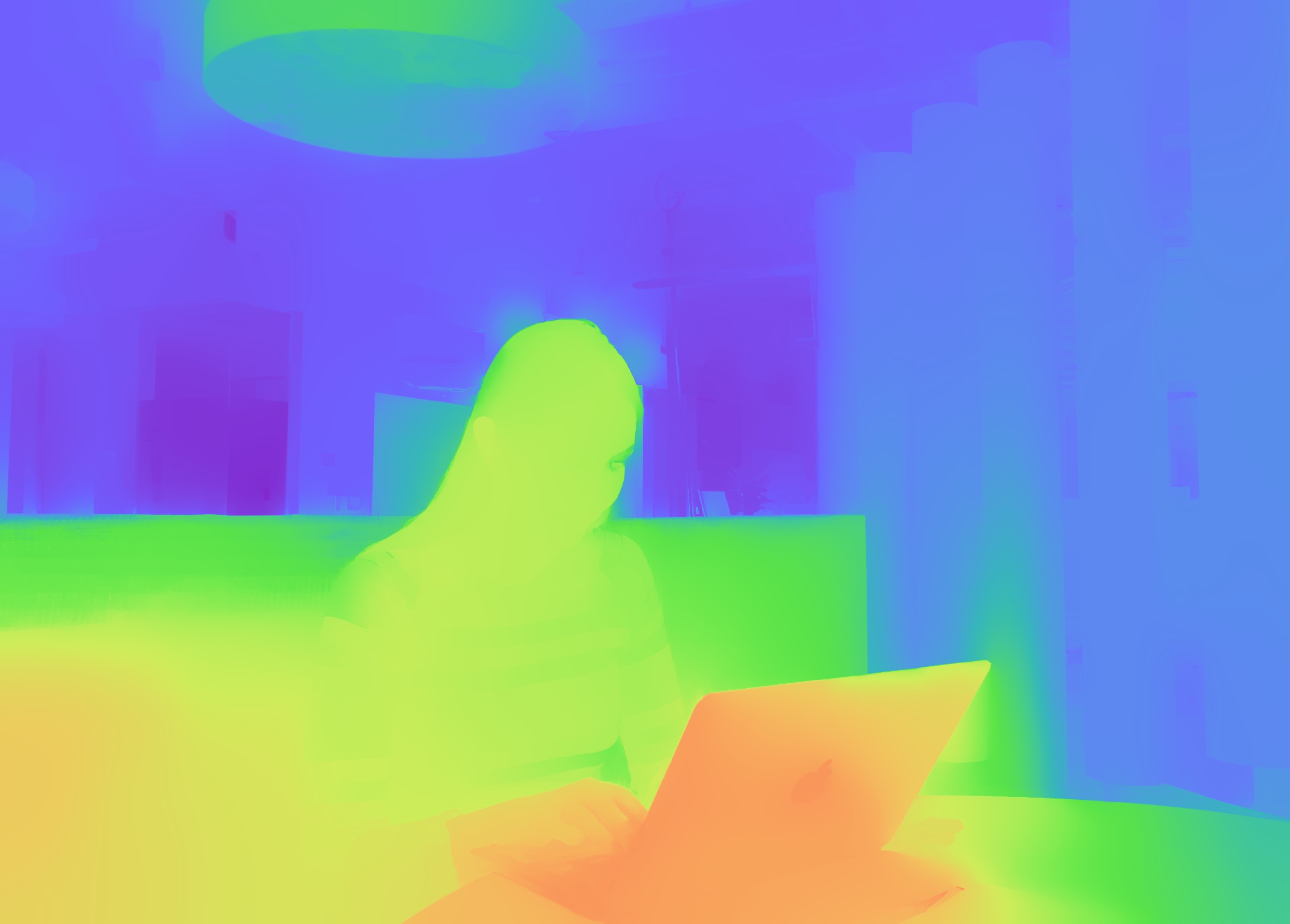}
    \caption{\cite{Barron2015A} Depth Output}
  \end{subfigure}
  \begin{subfigure}[!]{\blurwidthB}
    \includegraphics[width=\blurwidthB]{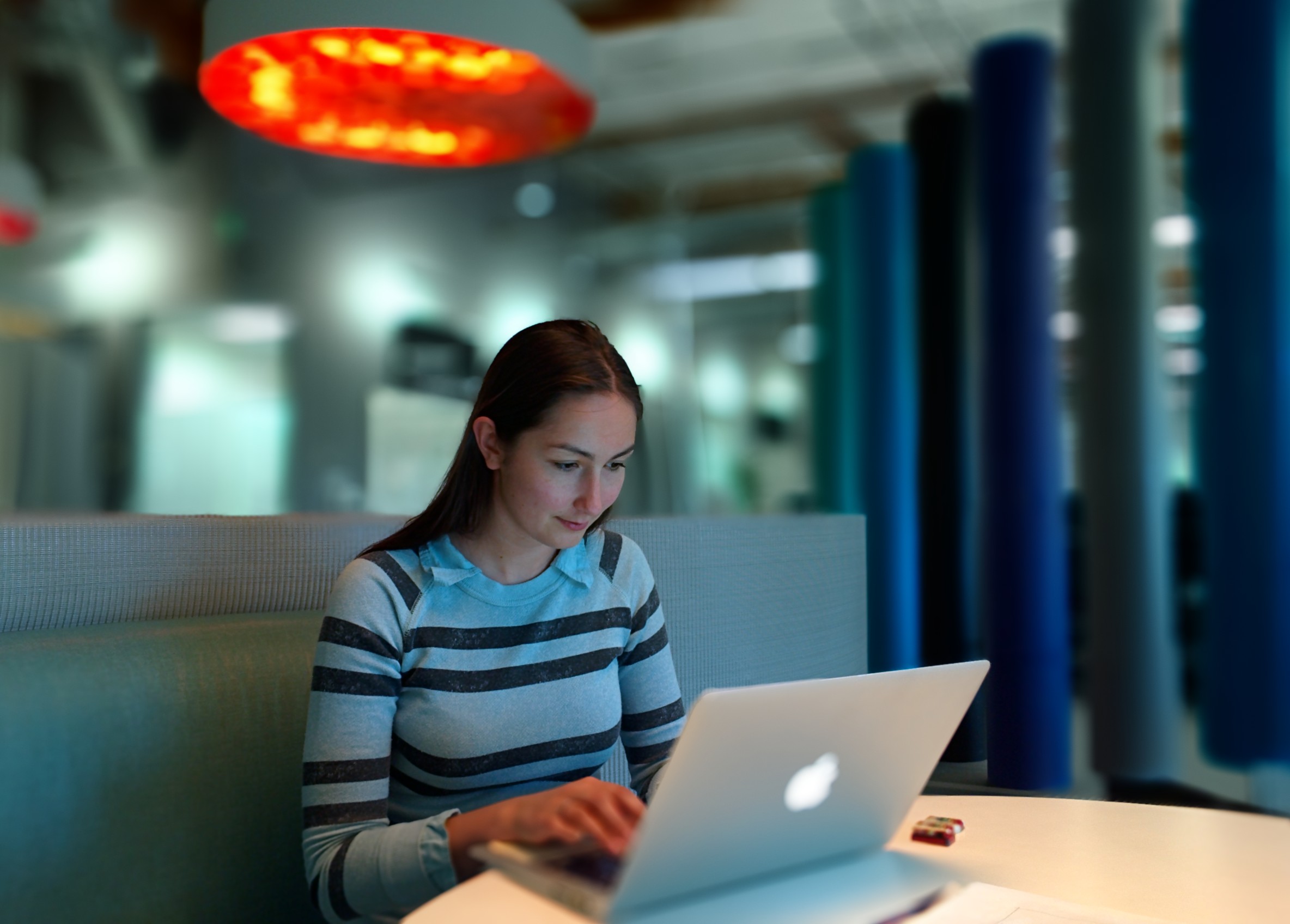}
    \caption{\cite{Barron2015A} Rendering}
  \end{subfigure}
  \\
  \begin{subfigure}[!]{\blurwidthB}
    \includegraphics[width=\blurwidthB]{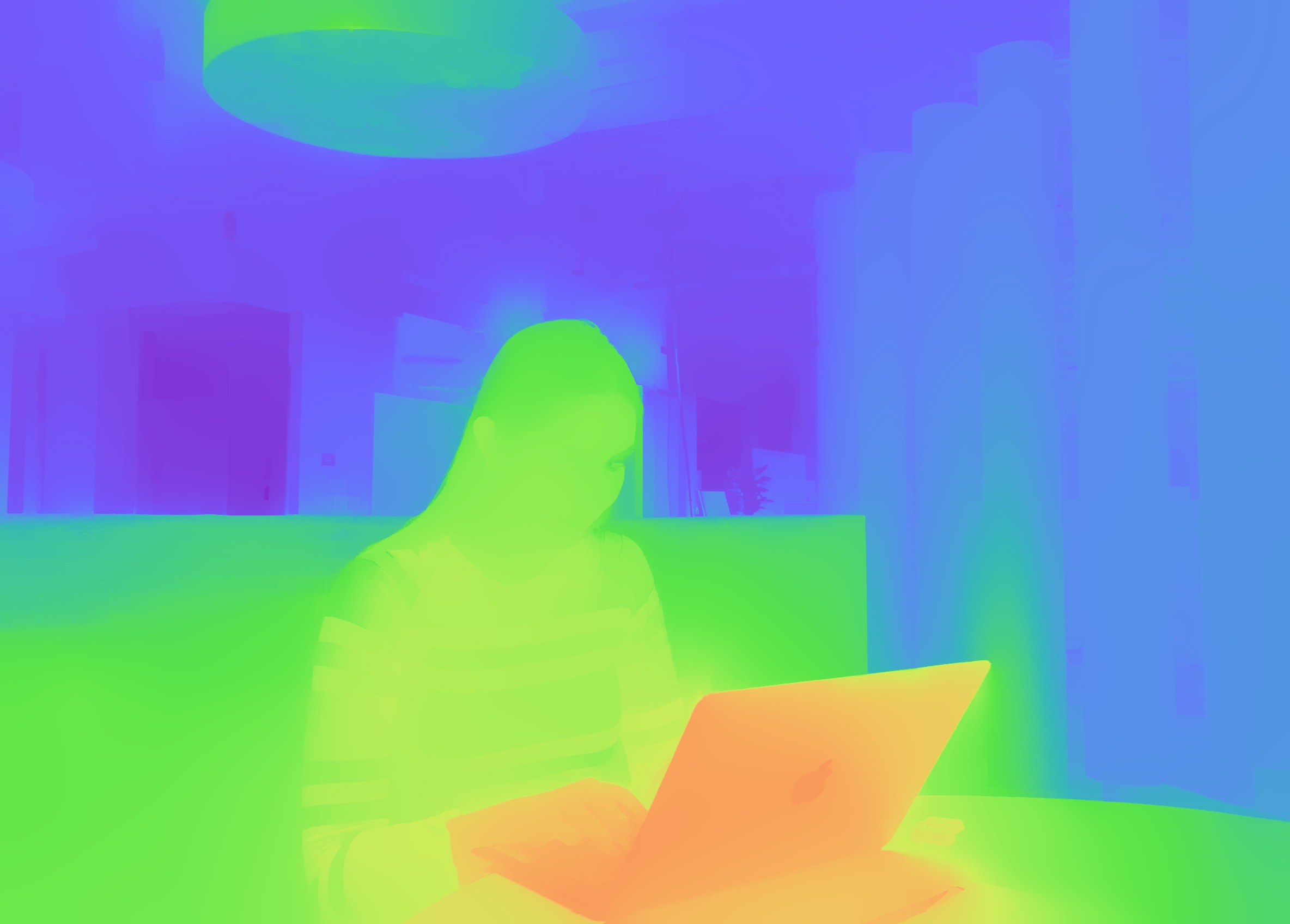}
    \caption{Our Depth Output}
  \end{subfigure}
  \begin{subfigure}[!]{\blurwidthB}
    \includegraphics[width=\blurwidthB]{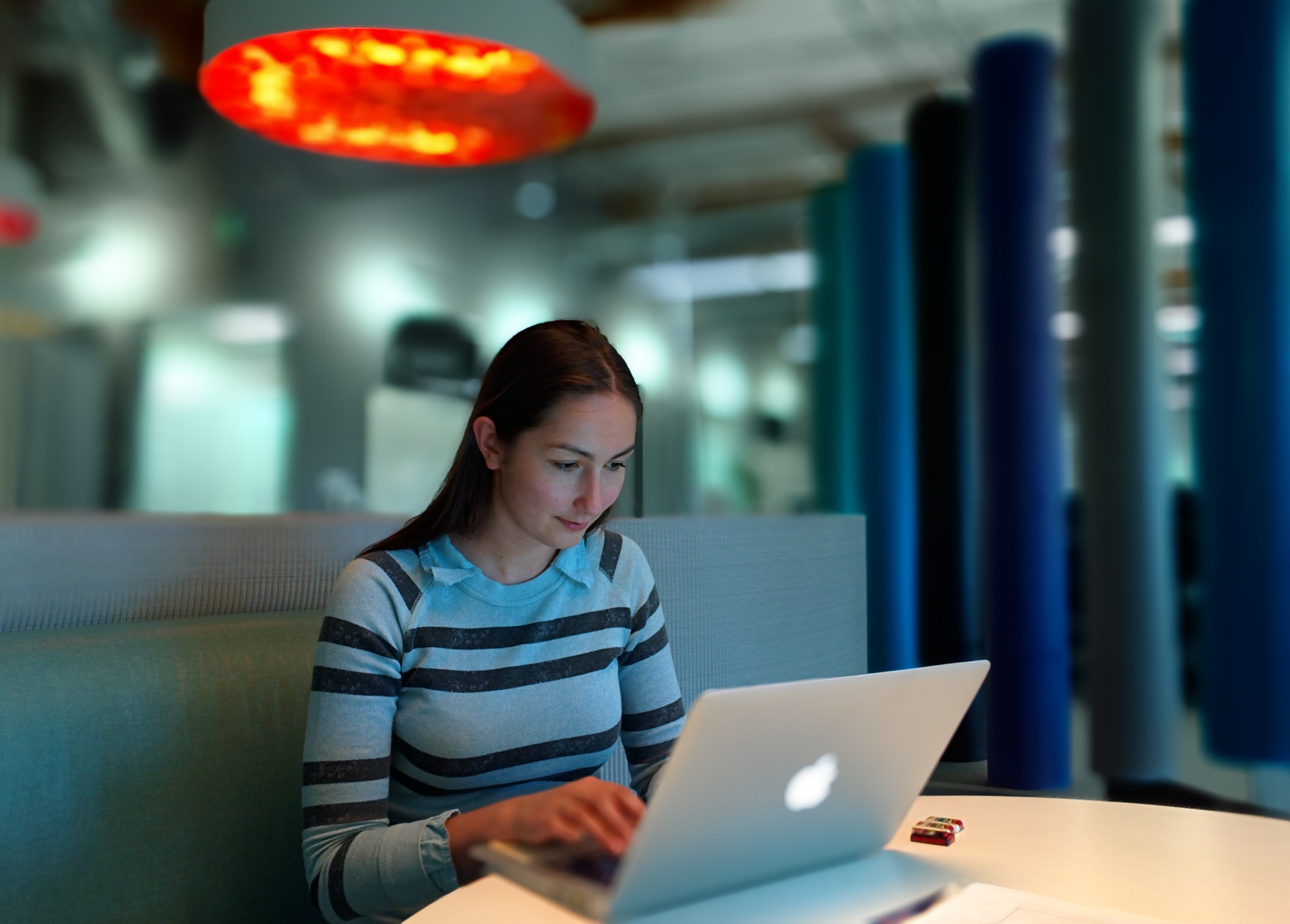}
    \caption{Our Rendering}
  \end{subfigure}
  \caption{
  Additional output for the stereo defocus task of \cite{Barron2015A}, in the same format as Figure~\ref{fig:stereo_defocus_supp1}.
  \label{fig:stereo_defocus_supp2}
  }
\end{figure*}

\newcommand{\superwidth}{1.12in}

\begin{figure*}[p]
\centering
  \begin{subfigure}[!]{\superwidth}
    \includegraphics[width=\superwidth]{figures/super_depth/1_image.jpg}
    \caption{Input Image}
  \end{subfigure}
  \begin{subfigure}[!]{\superwidth}
    \includegraphics[width=\superwidth]{figures/super_depth/1_true.jpg}
    \caption{Ground Truth}
  \end{subfigure}
  \begin{subfigure}[!]{\superwidth}
    \includegraphics[width=\superwidth]{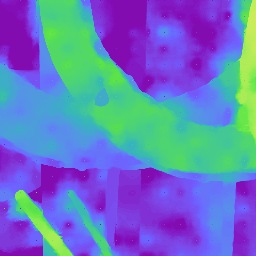}
    \caption{Liu \etal \cite{Liu2013}}
  \end{subfigure}
  \begin{subfigure}[!]{\superwidth}
    \includegraphics[width=\superwidth]{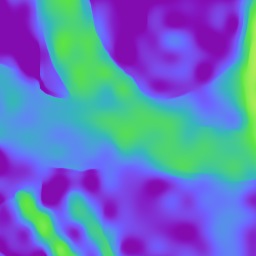}
    \caption{Shen \etal \cite{Shen2015}}
  \end{subfigure}
  \begin{subfigure}[!]{\superwidth}
    \includegraphics[width=\superwidth]{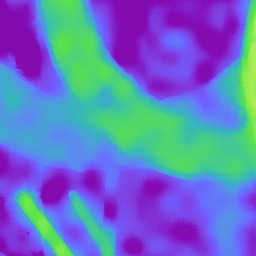}
    \caption{Chan \etal \cite{chan2008}}
  \end{subfigure}
  \begin{subfigure}[!]{\superwidth}
    \includegraphics[width=\superwidth]{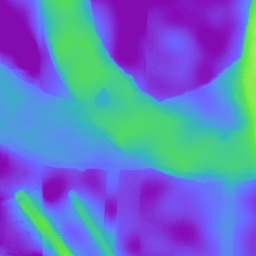}
    \caption{GF \cite{He2010,ferstl2013b}}
  \end{subfigure}
  \begin{subfigure}[!]{\superwidth}
    \includegraphics[width=\superwidth]{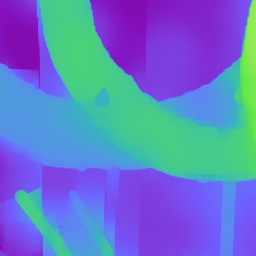}
    \caption{Min \etal \cite{Min2014}}
  \end{subfigure}
  \begin{subfigure}[!]{\superwidth}
    \includegraphics[width=\superwidth]{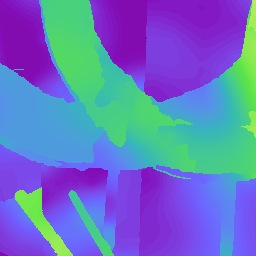}
    \caption{$\dagger$ Lu \cite{Lu2015}}
  \end{subfigure}
  \begin{subfigure}[!]{\superwidth}
    \includegraphics[width=\superwidth]{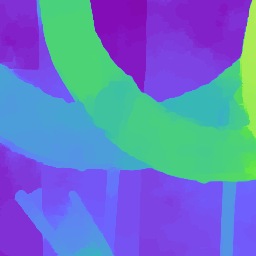}
    \caption{Park \etal \cite{Park2011}}
  \end{subfigure}
  \begin{subfigure}[!]{\superwidth}
    \includegraphics[width=\superwidth]{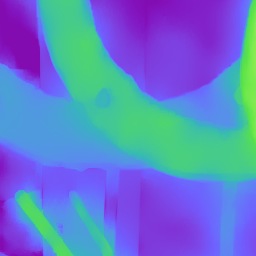}
    \caption{DT \cite{GastalOliveira2011DomainTransform}}
  \end{subfigure}
  \begin{subfigure}[!]{\superwidth}
    \includegraphics[width=\superwidth]{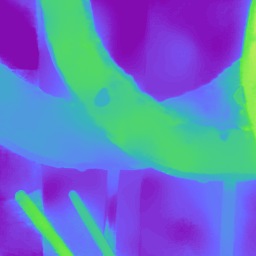}
    \caption{Ma \etal \cite{Ma2013}}
  \end{subfigure}
  \begin{subfigure}[!]{\superwidth}
    \includegraphics[width=\superwidth]{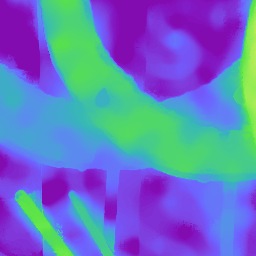}
    \caption{Zhang \etal \cite{Zhang2014}}
  \end{subfigure}
  \begin{subfigure}[!]{\superwidth}
    \includegraphics[width=\superwidth]{figures/super_depth/1_my_fastguide.jpg}
    \caption{FGF \cite{He2015}}
  \end{subfigure}
  \begin{subfigure}[!]{\superwidth}
    \includegraphics[width=\superwidth]{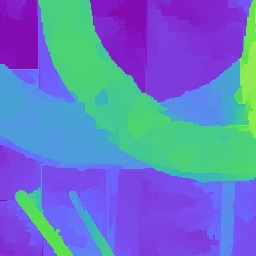}
    \caption{Yang 2015\, \cite{Yang2015}}
  \end{subfigure}
  \begin{subfigure}[!]{\superwidth}
    \includegraphics[width=\superwidth]{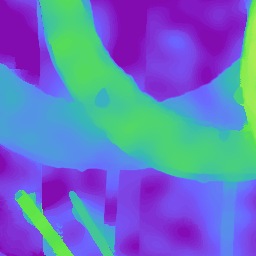}
    \caption{Yang 2007\, \cite{Yang2007}}
  \end{subfigure}
  \begin{subfigure}[!]{\superwidth}
    \includegraphics[width=\superwidth]{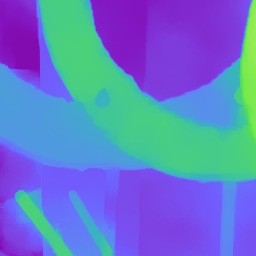}
    \caption{WLS \cite{FFLS2008}}
  \end{subfigure}
  \begin{subfigure}[!]{\superwidth}
    \includegraphics[width=\superwidth]{figures/super_depth/1_bilateral.jpg}
    \caption{JB \cite{Adams2010,Kopf2007}}
  \end{subfigure}
  \begin{subfigure}[!]{\superwidth}
    \includegraphics[width=\superwidth]{figures/super_depth/1_ferstl.jpg}
    \caption{Ferstl \etal \cite{ferstl2013b}}
  \end{subfigure}
  \begin{subfigure}[!]{\superwidth}
    \includegraphics[width=\superwidth]{figures/super_depth/1_li.jpg}
    \caption{$\dagger$ Li\etal \cite{Li2013}}
  \end{subfigure}
  \begin{subfigure}[!]{\superwidth}
    \includegraphics[width=\superwidth]{figures/super_depth/1_bsqs.jpg}
    \caption{BS (Ours)}
  \end{subfigure}
  \caption{Results for the depth superresolution task.
  Algorithms are sorted according to their average error on this benchmark, from upper left to lower right. Algorithms which use external training data are indicated with a dagger.
  \label{fig:super1}
  }
\end{figure*}

\begin{figure*}[p]
\centering
  \begin{subfigure}[!]{\superwidth}
    \includegraphics[width=\superwidth]{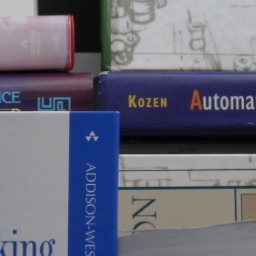}
    \caption{Input Image}
  \end{subfigure}
  \begin{subfigure}[!]{\superwidth}
    \includegraphics[width=\superwidth]{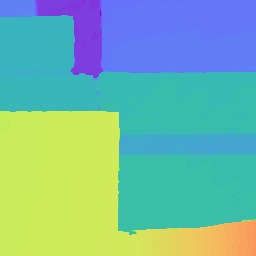}
    \caption{Ground Truth}
  \end{subfigure}
  \begin{subfigure}[!]{\superwidth}
    \includegraphics[width=\superwidth]{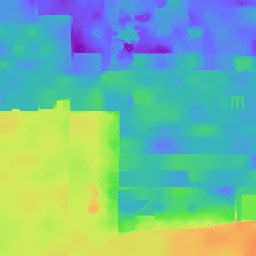}
    \caption{Liu \etal \cite{Liu2013}}
  \end{subfigure}
  \begin{subfigure}[!]{\superwidth}
    \includegraphics[width=\superwidth]{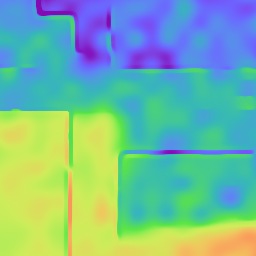}
    \caption{Shen \etal \cite{Shen2015}}
  \end{subfigure}
  \begin{subfigure}[!]{\superwidth}
    \includegraphics[width=\superwidth]{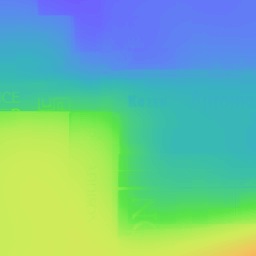}
    \caption{Chan \etal \cite{chan2008}}
  \end{subfigure}
  \begin{subfigure}[!]{\superwidth}
    \includegraphics[width=\superwidth]{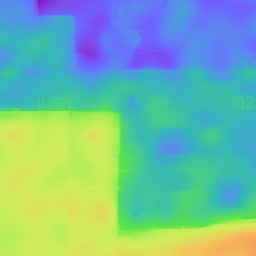}
    \caption{GF \cite{He2010,ferstl2013b}}
  \end{subfigure}
  \begin{subfigure}[!]{\superwidth}
    \includegraphics[width=\superwidth]{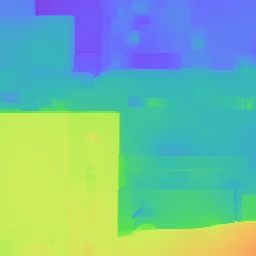}
    \caption{Min \etal \cite{Min2014}}
  \end{subfigure}
  \begin{subfigure}[!]{\superwidth}
    \includegraphics[width=\superwidth]{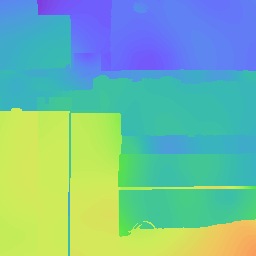}
    \caption{$\dagger$ Lu \cite{Lu2015}}
  \end{subfigure}
  \begin{subfigure}[!]{\superwidth}
    \includegraphics[width=\superwidth]{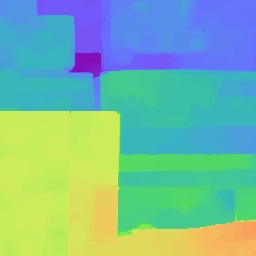}
    \caption{Park \etal \cite{Park2011}}
  \end{subfigure}
  \begin{subfigure}[!]{\superwidth}
    \includegraphics[width=\superwidth]{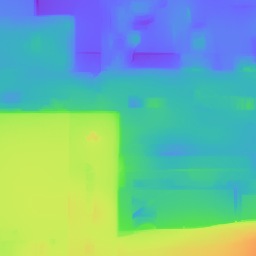}
    \caption{DT \cite{GastalOliveira2011DomainTransform}}
  \end{subfigure}
  \begin{subfigure}[!]{\superwidth}
    \includegraphics[width=\superwidth]{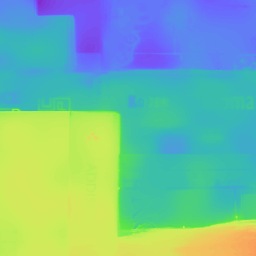}
    \caption{Ma \etal \cite{Ma2013}}
  \end{subfigure}
  \begin{subfigure}[!]{\superwidth}
    \includegraphics[width=\superwidth]{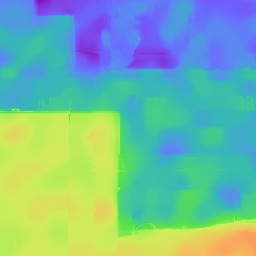}
    \caption{Zhang \etal \cite{Zhang2014}}
  \end{subfigure}
  \begin{subfigure}[!]{\superwidth}
    \includegraphics[width=\superwidth]{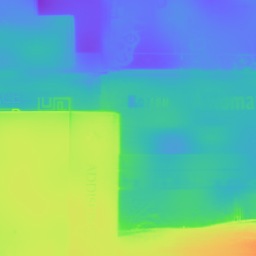}
    \caption{FGF \cite{He2015}}
  \end{subfigure}
  \begin{subfigure}[!]{\superwidth}
    \includegraphics[width=\superwidth]{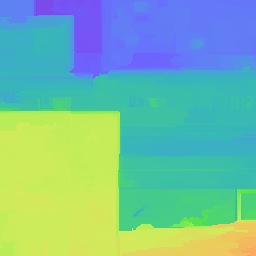}
    \caption{Yang 2015\, \cite{Yang2015}}
  \end{subfigure}
  \begin{subfigure}[!]{\superwidth}
    \includegraphics[width=\superwidth]{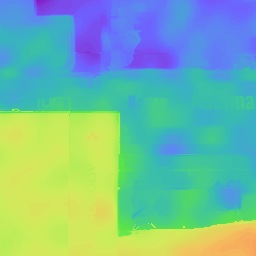}
    \caption{Yang 2007\, \cite{Yang2007}}
  \end{subfigure}
  \begin{subfigure}[!]{\superwidth}
    \includegraphics[width=\superwidth]{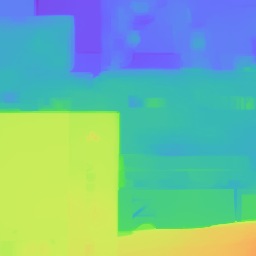}
    \caption{WLS \cite{FFLS2008}}
  \end{subfigure}
  \begin{subfigure}[!]{\superwidth}
    \includegraphics[width=\superwidth]{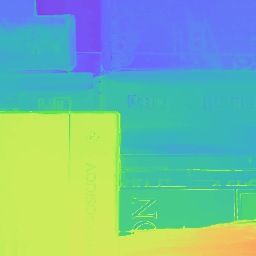}
    \caption{JB \cite{Adams2010,Kopf2007}}
  \end{subfigure}
  \begin{subfigure}[!]{\superwidth}
    \includegraphics[width=\superwidth]{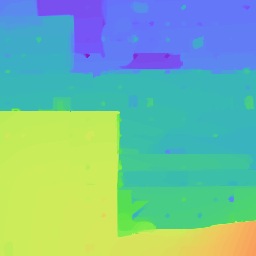}
    \caption{Ferstl \etal \cite{ferstl2013b}}
  \end{subfigure}
  \begin{subfigure}[!]{\superwidth}
    \includegraphics[width=\superwidth]{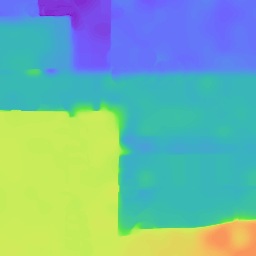}
    \caption{$\dagger$ Li\etal \cite{Li2013}}
  \end{subfigure}
  \begin{subfigure}[!]{\superwidth}
    \includegraphics[width=\superwidth]{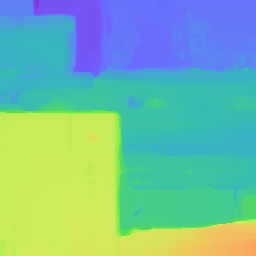}
    \caption{BS (Ours)}
  \end{subfigure}
  \caption{More results for the depth superresolution task.
  \label{fig:super2}
  }
\end{figure*}

\begin{figure*}[p]
\centering
  \begin{subfigure}[!]{\superwidth}
    \includegraphics[width=\superwidth]{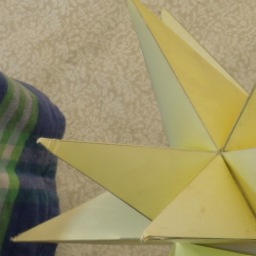}
    \caption{Input Image}
  \end{subfigure}
  \begin{subfigure}[!]{\superwidth}
    \includegraphics[width=\superwidth]{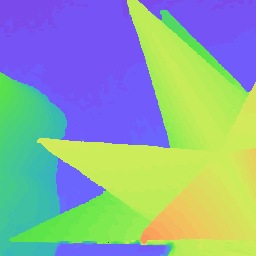}
    \caption{Ground Truth}
  \end{subfigure}
  \begin{subfigure}[!]{\superwidth}
    \includegraphics[width=\superwidth]{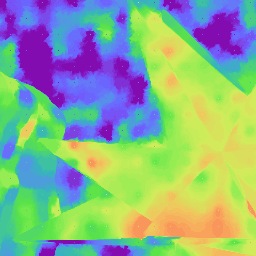}
    \caption{Liu \etal \cite{Liu2013}}
  \end{subfigure}
  \begin{subfigure}[!]{\superwidth}
    \includegraphics[width=\superwidth]{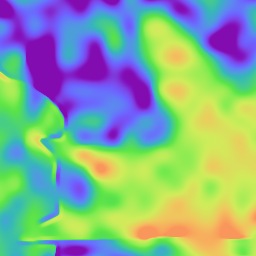}
    \caption{Shen \etal \cite{Shen2015}}
  \end{subfigure}
  \begin{subfigure}[!]{\superwidth}
    \includegraphics[width=\superwidth]{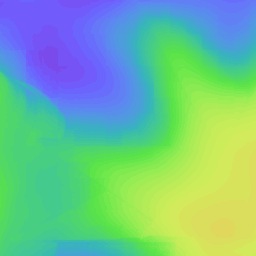}
    \caption{Chan \etal \cite{chan2008}}
  \end{subfigure}
  \begin{subfigure}[!]{\superwidth}
    \includegraphics[width=\superwidth]{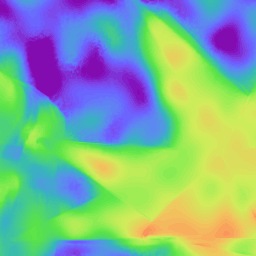}
    \caption{GF \cite{He2010,ferstl2013b}}
  \end{subfigure}
  \begin{subfigure}[!]{\superwidth}
    \includegraphics[width=\superwidth]{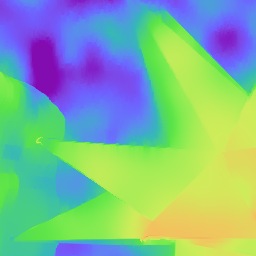}
    \caption{Min \etal \cite{Min2014}}
  \end{subfigure}
  \begin{subfigure}[!]{\superwidth}
    \includegraphics[width=\superwidth]{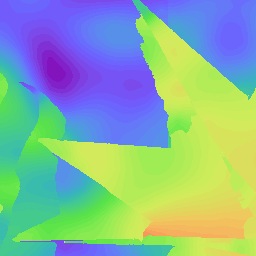}
    \caption{$\dagger$ Lu \cite{Lu2015}}
  \end{subfigure}
  \begin{subfigure}[!]{\superwidth}
    \includegraphics[width=\superwidth]{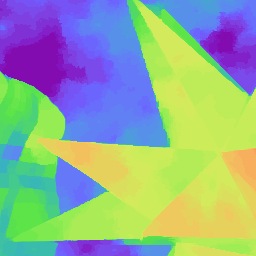}
    \caption{Park \etal \cite{Park2011}}
  \end{subfigure}
  \begin{subfigure}[!]{\superwidth}
    \includegraphics[width=\superwidth]{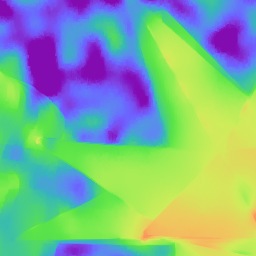}
    \caption{DT \cite{GastalOliveira2011DomainTransform}}
  \end{subfigure}
  \begin{subfigure}[!]{\superwidth}
    \includegraphics[width=\superwidth]{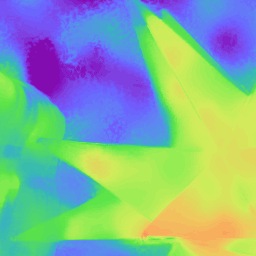}
    \caption{Ma \etal \cite{Ma2013}}
  \end{subfigure}
  \begin{subfigure}[!]{\superwidth}
    \includegraphics[width=\superwidth]{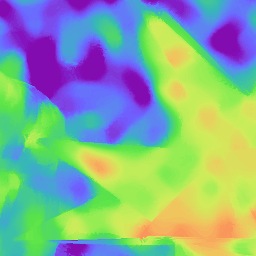}
    \caption{Zhang \etal \cite{Zhang2014}}
  \end{subfigure}
  \begin{subfigure}[!]{\superwidth}
    \includegraphics[width=\superwidth]{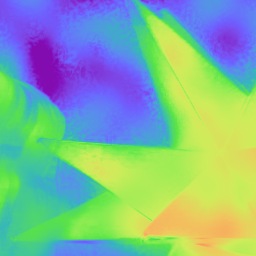}
    \caption{FGF \cite{He2015}}
  \end{subfigure}
  \begin{subfigure}[!]{\superwidth}
    \includegraphics[width=\superwidth]{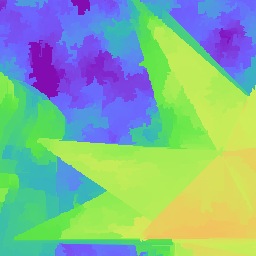}
    \caption{Yang 2015\, \cite{Yang2015}}
  \end{subfigure}
  \begin{subfigure}[!]{\superwidth}
    \includegraphics[width=\superwidth]{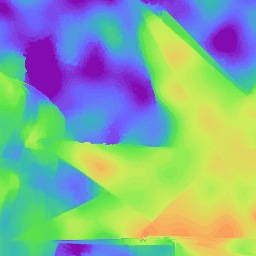}
    \caption{Yang 2007\, \cite{Yang2007}}
  \end{subfigure}
  \begin{subfigure}[!]{\superwidth}
    \includegraphics[width=\superwidth]{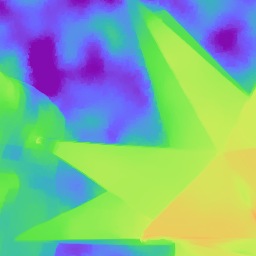}
    \caption{WLS \cite{FFLS2008}}
  \end{subfigure}
  \begin{subfigure}[!]{\superwidth}
    \includegraphics[width=\superwidth]{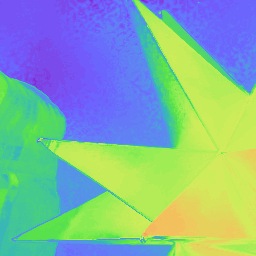}
    \caption{JB \cite{Adams2010,Kopf2007}}
  \end{subfigure}
  \begin{subfigure}[!]{\superwidth}
    \includegraphics[width=\superwidth]{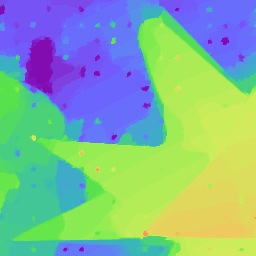}
    \caption{Ferstl \etal \cite{ferstl2013b}}
  \end{subfigure}
  \begin{subfigure}[!]{\superwidth}
    \includegraphics[width=\superwidth]{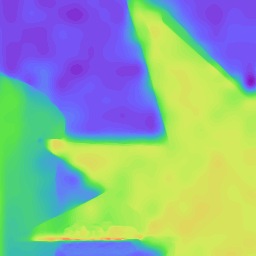}
    \caption{$\dagger$ Li\etal \cite{Li2013}}
  \end{subfigure}
  \begin{subfigure}[!]{\superwidth}
    \includegraphics[width=\superwidth]{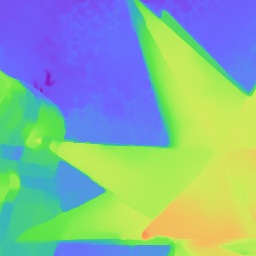}
    \caption{BS (Ours)}
  \end{subfigure}
  \caption{More results for the depth superresolution task.
  \label{fig:super3}
  }
\end{figure*}

\newcommand{\colorwidth}{1.35in}

\begin{figure*}[p]
\centering
  \begin{subfigure}[!]{\colorwidth}
  \begin{tabular}{c}
    \includegraphics[width=\colorwidth]{figures/colorize/example_m.jpg} \\
    \includegraphics[width=\colorwidth]{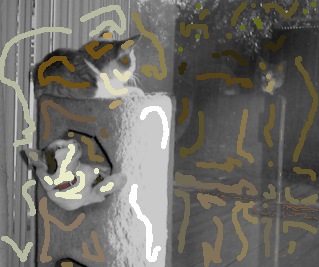} \\
    \includegraphics[width=\colorwidth]{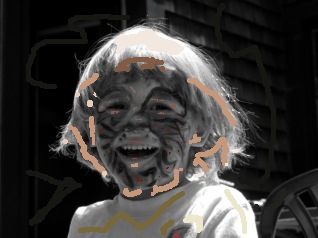} \\
    \includegraphics[width=\colorwidth]{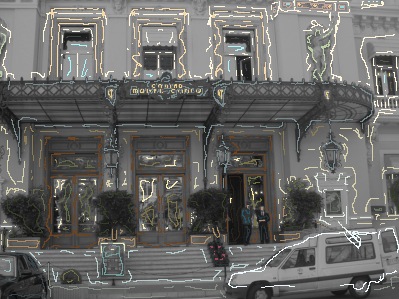} \\
    \includegraphics[width=\colorwidth]{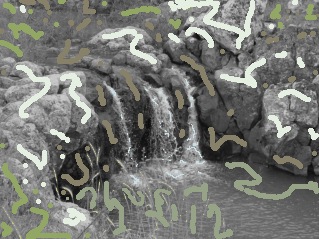} \\
    \includegraphics[width=\colorwidth]{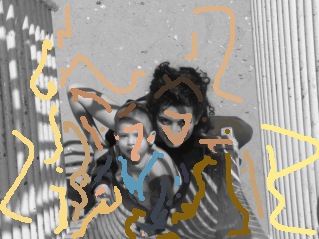}
  \end{tabular}
  \caption{Input }
  \end{subfigure}
  \begin{subfigure}[!]{\colorwidth}
  \begin{tabular}{c}
    \includegraphics[width=\colorwidth]{figures/colorize/example_res.jpg} \\
    \includegraphics[width=\colorwidth]{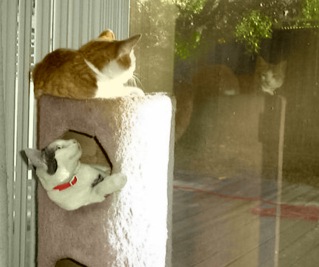} \\
    \includegraphics[width=\colorwidth]{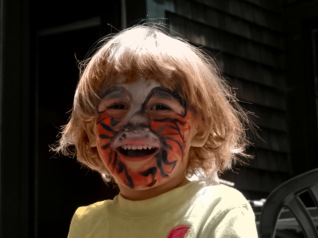} \\
    \includegraphics[width=\colorwidth]{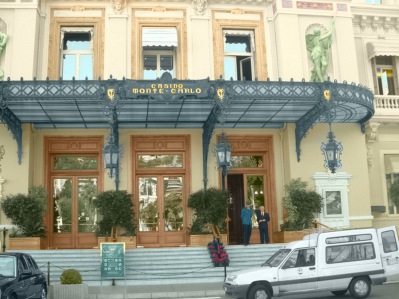} \\
    \includegraphics[width=\colorwidth]{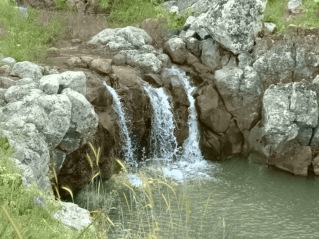} \\
    \includegraphics[width=\colorwidth]{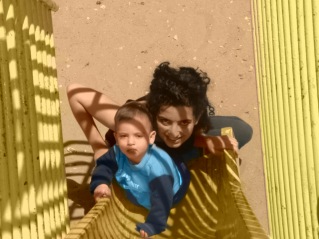}
  \end{tabular}
  \caption{Levin \etal \cite{Levin2004}}
  \end{subfigure}
  \begin{subfigure}[!]{\colorwidth}
  \begin{tabular}{c}
    \includegraphics[width=\colorwidth]{figures/colorize/example_output.jpg} \\
    \includegraphics[width=\colorwidth]{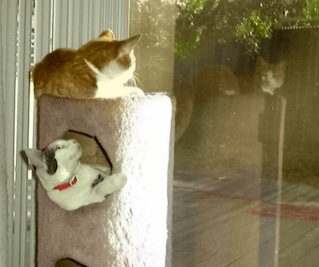} \\
    \includegraphics[width=\colorwidth]{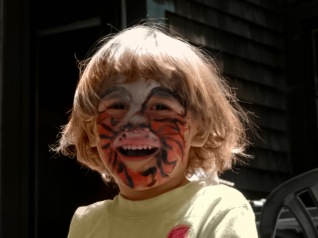} \\
    \includegraphics[width=\colorwidth]{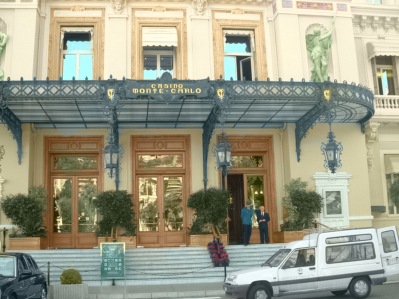} \\
    \includegraphics[width=\colorwidth]{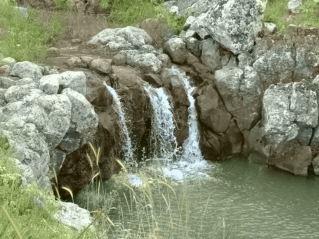} \\
    \includegraphics[width=\colorwidth]{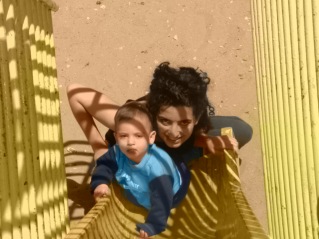}
  \end{tabular}
  \caption{Our results}
  \end{subfigure}
  \caption{
  Results for the colorization task, using the images and scribbles from \cite{Levin2004}.
  Our algorithm's output is nearly indistinguishable from that of \cite{Levin2004}, while being $95 \times$ faster.
  \label{fig:colorization_supp}
  }
\end{figure*}

\clearpage

\bibliographystyle{splncs03}
\bibliography{bs}